\documentclass[11pt]{article}

\usepackage[preprint]{acl}

\usepackage{times}
\usepackage{latexsym}
\usepackage{amsmath}

\usepackage[T1]{fontenc}

\usepackage[utf8]{inputenc}

\usepackage{microtype}

\usepackage{inconsolata}

\usepackage{graphicx}
\usepackage{booktabs}
\usepackage{array}

\usepackage[T1]{fontenc}
\usepackage{booktabs,tabularx,array,pifont,xcolor,graphicx}

\usepackage[most]{tcolorbox}
\usepackage{enumitem}
\usepackage{subcaption}
\usepackage{etoc}

\newcommand{\ourbench}{\mbox{\textsc{TrialPD}-\textsc{Bench}}}
\newcommand{\ourmethod}{\mbox{\textsc{TrialScribe}}}

\definecolor{cap}{HTML}{4A8F7D}
\usepackage{fontawesome5}
\definecolor{taskind}{HTML}{4D7FA6}\definecolor{taskcrit}{HTML}{448C7D}\definecolor{taskoe}{HTML}{8875A6}\definecolor{tasksoa}{HTML}{C38A50}
\newcommand{\taskicon}[2]{\makebox[1.3em][c]{\textcolor{#1}{#2}}\hspace{0.3em}}
\usetikzlibrary{svg.path}
\definecolor{hdrfa}{HTML}{2B5C8A}\definecolor{hdrer}{HTML}{C0622F}\definecolor{lgkimi}{HTML}{1783FF}\definecolor{lgdeepseek}{HTML}{4D6BFE}\definecolor{lgqwenA}{HTML}{6336E7}\definecolor{lgqwenB}{HTML}{6F69F7}
\newcommand{\logoopenai}{\raisebox{-0.17em}{\begin{tikzpicture}[scale=0.38,yscale=-1,even odd rule]
  \fill[black] svg {M 9.205 8.658 v -2.26 c 0 -0.19 0.072 -0.333 0.238 -0.428 l 4.543 -2.616 c 0.619 -0.357 1.356 -0.523 2.117 -0.523 2.854 0 4.662 2.212 4.662 4.566 0 0.167 0 0.357 -0.024 0.547 l -4.71 -2.759 a 0.797 0.797 0 0 0 -0.856 0 l -5.97 3.473 z m 10.609 8.8 V 12.06 c 0 -0.333 -0.143 -0.57 -0.429 -0.737 l -5.97 -3.473 1.95 -1.118 l 0.476 0 l 4.543 2.617 c 1.309 0.76 2.189 2.378 2.189 3.948 0 1.808 -1.07 3.473 -2.76 4.163 z M 7.802 12.703 l -1.95 -1.142 c -0.167 -0.095 -0.239 -0.238 -0.239 -0.428 V 5.899 c 0 -2.545 1.95 -4.472 4.591 -4.472 1 0 1.927 0.333 2.712 0.928 L 8.23 5.067 c -0.285 0.166 -0.428 0.404 -0.428 0.737 v 6.898 z M 12 15.128 l -2.795 -1.57 v -3.33 L 12 8.658 l 2.795 1.57 v 3.33 L 12 15.128 z m 1.796 7.23 c -1 0 -1.927 -0.332 -2.712 -0.927 l 4.686 -2.712 c 0.285 -0.166 0.428 -0.404 0.428 -0.737 v -6.898 l 1.974 1.142 c 0.167 0.095 0.238 0.238 0.238 0.428 v 5.233 c 0 2.545 -1.974 4.472 -4.614 4.472 z m -5.637 -5.303 l -4.544 -2.617 c -1.308 -0.761 -2.188 -2.378 -2.188 -3.948 A 4.482 4.482 0 0 1 4.21 6.327 v 5.423 c 0 0.333 0.143 0.571 0.428 0.738 l 5.947 3.449 -1.95 1.118 l -0.476 0 z m -0.262 3.9 c -2.688 0 -4.662 -2.021 -4.662 -4.519 0 -0.19 0.024 -0.38 0.047 -0.57 l 4.686 2.71 c 0.286 0.167 0.571 0.167 0.856 0 l 5.97 -3.448 v 2.26 c 0 0.19 -0.07 0.333 -0.237 0.428 l -4.543 2.616 c -0.619 0.357 -1.356 0.523 -2.117 0.523 z m 5.899 2.83 a 5.947 5.947 0 0 0 5.827 -4.756 C 22.287 18.339 24 15.84 24 13.296 c 0 -1.665 -0.713 -3.282 -1.998 -4.448 0.119 -0.5 0.19 -0.999 0.19 -1.498 0 -3.401 -2.759 -5.947 -5.946 -5.947 -0.642 0 -1.26 0.095 -1.88 0.31 A 5.962 5.962 0 0 0 10.205 0 a 5.947 5.947 0 0 0 -5.827 4.757 C 1.713 5.447 0 7.945 0 10.49 c 0 1.666 0.713 3.283 1.998 4.448 -0.119 0.5 -0.19 1 -0.19 1.499 0 3.401 2.759 5.946 5.946 5.946 0.642 0 1.26 -0.095 1.88 -0.309 a 5.96 5.96 0 0 0 4.162 1.713 z};
\end{tikzpicture}}}
\newcommand{\logoanthropic}{\raisebox{-0.17em}{\begin{tikzpicture}[scale=0.38,yscale=-1,even odd rule]
  \fill[black] svg {M 13.827 3.52 h 3.603 L 24 20 h -3.603 l -6.57 -16.48 z m -7.258 0 h 3.767 L 16.906 20 h -3.674 l -1.343 -3.461 H 5.017 l -1.344 3.46 H 0 L 6.57 3.522 z m 4.132 9.959 L 8.453 7.687 6.205 13.48 H 10.7 z};
\end{tikzpicture}}}
\newcommand{\logokimi}{\raisebox{-0.17em}{\begin{tikzpicture}[scale=0.38,yscale=-1,even odd rule]
  \fill[lgkimi] svg {M 21.846 0 a 1.923 1.923 0 1 1 0 3.846 H 20.15 l -0.227 -0.226 V 1.923 C 19.923 0.861 20.784 0 21.846 0 z};
  \fill[black] svg {M 11.065 11.199 l 7.257 -7.2 c 0.137 -0.136 0.06 -0.41 -0.116 -0.41 H 14.3 l -0.117 0.051 l -7.82 7.756 c -0.122 0.12 -0.302 0.013 -0.302 -0.179 V 3.82 c 0 -0.127 -0.083 -0.23 -0.185 -0.23 H 3.186 c -0.103 0 -0.186 0.103 -0.186 0.23 V 19.77 c 0 0.128 0.083 0.23 0.186 0.23 h 2.69 c 0.103 0 0.186 -0.102 0.186 -0.23 v -3.25 c 0 -0.069 0.025 -0.135 0.069 -0.178 l 2.424 -2.406 l 0.205 -0.023 l 6.484 4.772 a 7.677 7.677 0 0 0 3.453 1.283 c 0.108 0.012 0.2 -0.095 0.2 -0.23 v -3.06 c 0 -0.117 -0.07 -0.212 -0.164 -0.227 a 5.028 5.028 0 0 1 -2.027 -0.807 l -5.613 -4.064 c -0.117 -0.078 -0.132 -0.279 -0.028 -0.381 z};
\end{tikzpicture}}}
\newcommand{\logodeepseek}{\raisebox{-0.17em}{\begin{tikzpicture}[scale=0.38,yscale=-1,even odd rule]
  \fill[lgdeepseek] svg {M 23.748 4.482 c -0.254 -0.124 -0.364 0.113 -0.512 0.234 -0.051 0.039 -0.094 0.09 -0.137 0.136 -0.372 0.397 -0.806 0.657 -1.373 0.626 -0.829 -0.046 -1.537 0.214 -2.163 0.848 -0.133 -0.782 -0.575 -1.248 -1.247 -1.548 -0.352 -0.156 -0.708 -0.311 -0.955 -0.65 -0.172 -0.241 -0.219 -0.51 -0.305 -0.774 -0.055 -0.16 -0.11 -0.323 -0.293 -0.35 -0.2 -0.031 -0.278 0.136 -0.356 0.276 -0.313 0.572 -0.434 1.202 -0.422 1.84 0.027 1.436 0.633 2.58 1.838 3.393 0.137 0.093 0.172 0.187 0.129 0.323 -0.082 0.28 -0.18 0.552 -0.266 0.833 -0.055 0.179 -0.137 0.217 -0.329 0.14 a 5.526 5.526 0 0 1 -1.736 -1.18 c -0.857 -0.828 -1.631 -1.742 -2.597 -2.458 a 11.365 11.365 0 0 0 -0.689 -0.471 c -0.985 -0.957 0.13 -1.743 0.388 -1.836 0.27 -0.098 0.093 -0.432 -0.779 -0.428 -0.872 0.004 -1.67 0.295 -2.687 0.684 a 3.055 3.055 0 0 1 -0.465 0.137 9.597 9.597 0 0 0 -2.883 -0.102 c -1.885 0.21 -3.39 1.102 -4.497 2.623 C 0.082 8.606 -0.231 10.684 0.152 12.85 c 0.403 2.284 1.569 4.175 3.36 5.653 1.858 1.533 3.997 2.284 6.438 2.14 1.482 -0.085 3.133 -0.284 4.994 -1.86 0.47 0.234 0.962 0.327 1.78 0.397 0.63 0.059 1.236 -0.03 1.705 -0.128 0.735 -0.156 0.684 -0.837 0.419 -0.961 -2.155 -1.004 -1.682 -0.595 -2.113 -0.926 1.096 -1.296 2.746 -2.642 3.392 -7.003 0.05 -0.347 0.007 -0.565 0 -0.845 -0.004 -0.17 0.035 -0.237 0.23 -0.256 a 4.173 4.173 0 0 0 1.545 -0.475 c 1.396 -0.763 1.96 -2.015 2.093 -3.517 0.02 -0.23 -0.004 -0.467 -0.247 -0.588 z M 11.581 18 c -2.089 -1.642 -3.102 -2.183 -3.52 -2.16 -0.392 0.024 -0.321 0.471 -0.235 0.763 0.09 0.288 0.207 0.486 0.371 0.739 0.114 0.167 0.192 0.416 -0.113 0.603 -0.673 0.416 -1.842 -0.14 -1.897 -0.167 -1.361 -0.802 -2.5 -1.86 -3.301 -3.307 -0.774 -1.393 -1.224 -2.887 -1.298 -4.482 -0.02 -0.386 0.093 -0.522 0.477 -0.592 a 4.696 4.696 0 0 1 1.529 -0.039 c 2.132 0.312 3.946 1.265 5.468 2.774 0.868 0.86 1.525 1.887 2.202 2.891 0.72 1.066 1.494 2.082 2.48 2.914 0.348 0.292 0.625 0.514 0.891 0.677 -0.802 0.09 -2.14 0.11 -3.054 -0.614 z m 1 -6.44 l 0.415 -0.287 0.2 0.288 -0.31 0.307 -0.304 -0.308 z m 3.11 1.596 c -0.2 0.081 -0.399 0.151 -0.59 0.16 a 1.245 1.245 0 0 1 -0.798 -0.254 c -0.274 -0.23 -0.47 -0.358 -0.552 -0.758 a 1.73 1.73 0 0 1 0.016 -0.588 c 0.07 -0.327 -0.008 -0.537 -0.239 -0.727 -0.187 -0.156 -0.426 -0.199 -0.688 -0.199 a 0.559 0.559 0 0 1 -0.254 -0.078 c -0.11 -0.054 -0.2 -0.19 -0.114 -0.358 0.028 -0.054 0.16 -0.186 0.192 -0.21 0.356 -0.202 0.767 -0.136 1.146 0.016 0.352 0.144 0.618 0.408 1.001 0.782 0.391 0.451 0.462 0.576 0.685 0.914 0.176 0.265 0.336 0.537 0.445 0.848 0.067 0.195 -0.019 0.354 -0.25 0.452 z};
\end{tikzpicture}}}
\newcommand{\logoqwen}{\raisebox{-0.17em}{\begin{tikzpicture}[scale=0.38,yscale=-1,even odd rule]
  \shade[left color=lgqwenA!84,right color=lgqwenB!84] svg {M 12.604 1.34 c 0.393 0.69 0.784 1.382 1.174 2.075 l 0.157 0.091 h 5.552 c 0.174 0 0.322 0.11 0.446 0.327 l 1.454 2.57 c 0.19 0.337 0.24 0.478 0.024 0.837 -0.26 0.43 -0.513 0.864 -0.76 1.3 l -0.367 0.658 c -0.106 0.196 -0.223 0.28 -0.04 0.512 l 2.652 4.637 c 0.172 0.301 0.111 0.494 -0.043 0.77 -0.437 0.785 -0.882 1.564 -1.335 2.34 -0.159 0.272 -0.352 0.375 -0.68 0.37 -0.777 -0.016 -1.552 -0.01 -2.327 0.016 l -0.081 0.05 -2.705 4.74 c -0.169 0.293 -0.38 0.363 -0.725 0.364 -0.997 0.003 -2.002 0.004 -3.017 0.002 a 0.537 0.537 0 0 1 -0.465 -0.271 l -1.335 -2.323 l -0.083 -0.049 H 4.982 c -0.285 0.03 -0.553 -0.001 -0.805 -0.092 l -1.603 -2.77 a 0.543 0.543 0 0 1 -0.002 -0.54 l 1.207 -2.12 l 0 -0.197 -1.875 -3.272 l -0.79 -1.395 c -0.16 -0.31 -0.173 -0.496 0.095 -0.965 0.465 -0.813 0.927 -1.625 1.387 -2.436 0.132 -0.234 0.304 -0.334 0.584 -0.335 l 2.589 -0.001 0.107 -0.063 l 2.806 -4.895 l 0.422 -0.246 c 0.524 -0.001 1.053 0 1.583 -0.006 L 11.704 1 c 0.341 -0.003 0.724 0.032 0.9 0.34 z m -3.432 0.403 l -0.052 0.03 L 6.254 6.788 l -0.135 0.078 H 3.253 c -0.056 0 -0.07 0.025 -0.041 0.074 l 5.81 10.156 c 0.025 0.042 0.013 0.062 -0.034 0.063 l -2.795 0.015 l -0.2 0.116 l -1.32 2.31 c -0.044 0.078 -0.021 0.118 0.068 0.118 l 5.716 0.008 c 0.046 0 0.08 0.02 0.104 0.061 l 1.403 2.454 c 0.046 0.081 0.092 0.082 0.139 0 l 5.006 -8.76 0.783 -1.382 l 0.096 0 l 1.424 2.53 l 0.107 0.062 l 2.763 -0.02 l 0.035 -0.02 0 -0.04 l -2.9 -5.086 l 0 -0.113 l 0.293 -0.507 1.12 -1.977 c 0.024 -0.041 0.012 -0.062 -0.035 -0.062 H 9.2 c -0.059 0 -0.073 -0.026 -0.043 -0.077 l 1.434 -2.505 l 0 -0.114 L 9.225 1.774 l -0.053 -0.031 z m 6.29 8.02 c 0.046 0 0.058 0.02 0.034 0.06 l -0.832 1.465 -2.613 4.585 l -0.05 0.029 -0.05 -0.029 L 8.498 9.841 c -0.02 -0.034 -0.01 -0.052 0.028 -0.054 l 0.216 -0.012 6.722 -0.012 z};
\end{tikzpicture}}}

\newtcolorbox{capabilitybox}[1][]{colback=cap!5!white,colframe=cap,boxsep=-4pt,left=10pt,right=10pt,top=10pt,bottom=10pt,arc=2mm,boxrule=0.8pt,#1}

\newcommand{\PDyes}{\textcolor{green!55!black}{\ding{51}}}
\newcommand{\PDno}{\textcolor{red!75!black}{\ding{55}}}
\newcommand{\PDpart}{\textcolor{orange}{\ensuremath{\triangle}}}
\newcommand{\PDnone}{\PDno}

\title{The Last Mile of Coding Agents for Long-Document Tasks: An Empirical Study on Clinical Trial Protocol Digitization}

\author{
  \textbf{Jiacheng Lin\textsuperscript{1}},
  \textbf{Geoff Low\textsuperscript{2}},
  \textbf{Nishit Agarwal\textsuperscript{2}},
  \textbf{Rahul Jain\textsuperscript{2}},\\
  \textbf{Afrah Shafquat\textsuperscript{2}},
  \textbf{Mandis Beigi\textsuperscript{2}},
  \textbf{Jimeng Sun\textsuperscript{1}}
\\
  \textsuperscript{1}University of Illinois Urbana-Champaign,
  \textsuperscript{2}Medidata Solutions
\\
  \small{\textbf{Project page:} \url{https://linjc16.github.io/trialpd-bench/}}
}

\begin{document}
\maketitle
\etocdepthtag.toc{mtmain}
\begin{abstract}
Clinical trial protocols serve as the source of truth for study design, but are mostly written and used as unstructured PDF documents. Digitizing protocols turns static content into structured, machine-readable data that clinical software can analyze, validate, and execute.
Coding agents, which select tools, execute code, and revise their approach as they work, are increasingly applied to such long-document tasks, yet it remains unclear how reliably they turn complete documents into accurate structured outputs. We study this question using protocol digitization as a testbed. We introduce \ourbench{}, an expert-annotated benchmark of 150 publicly available protocol PDFs spanning four task categories. Beyond final-answer performance, our evaluation measures evidence recall from agent traces, agent behavior, and cost. Across nine proprietary and open-weight models in five coding-agent harnesses, \textbf{even the strongest agents do not fully solve protocol digitization}. Our analysis further shows that \textbf{agents usually locate the required evidence but fail in the last mile of transcribing and structuring it}. We further develop \ourmethod{}, a domain-specific harness that equips coding agents with task skills and a verification loop. \ourmethod{} improves a locally served open-weight model by 6.2-59.4 points across eight metrics, \textbf{enabling it to match or surpass proprietary models such as GPT-5.6-Sol}.
\end{abstract}


\section{Introduction}

\begin{figure*}[t]
    \centering
    \includegraphics[width=\textwidth]{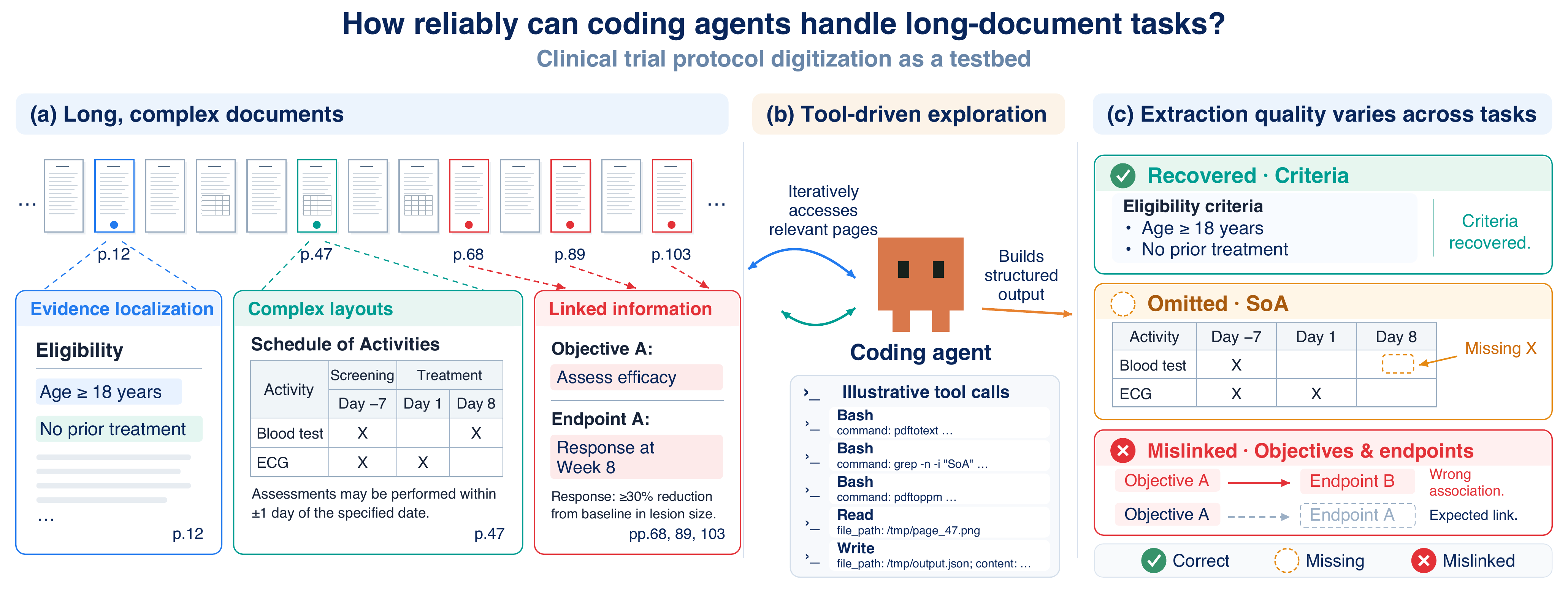}\\[3pt]
    \includegraphics[width=0.4227\textwidth]{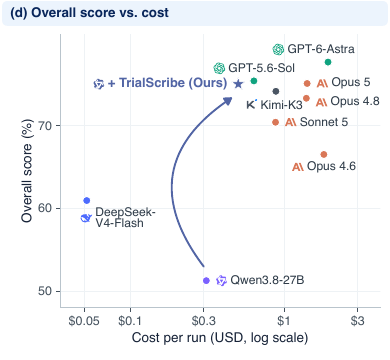}\hfill
    \includegraphics[width=0.5771\textwidth]{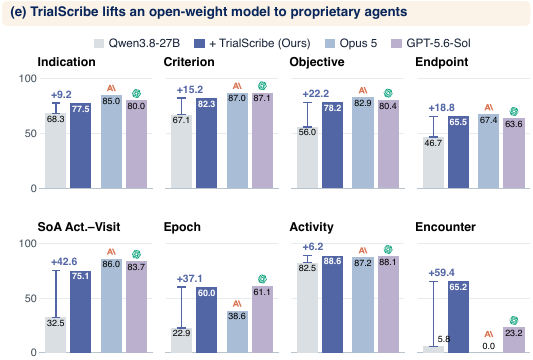}
    \caption{Overview of this study. \textbf{(a--c)} Coding agents must localize evidence, interpret complex layouts, and link information across long protocols; output quality varies across tasks. \textbf{(d)} Overall score on \ourbench{} against average cost per run. \textbf{(e)} \ourmethod{} with Qwen3.8-27B versus the base model, Opus 5, and GPT-5.6-Sol.}
    \label{fig:overview}
    \vspace{-1em}
\end{figure*}
Recent advances in coding agents, exemplified by Claude Code \citep{anthropic_claude_code} and Codex \citep{openai2025codex}, have extended their use beyond software engineering to a broader range of complex tasks \citep{wang2026pptbench,lyttelton2026agents,wang2026naturebench,lyu2026mls}. These agents can select tools, generate and execute code, and revise their approach based on intermediate results, providing flexibility in how they organize and complete a task \citep{wang2024executable, ning2026code}. Such capabilities make coding agents promising for long-document tasks, where relevant information may be distributed across many pages and different document structures may require different processing strategies. However, it remains unclear whether coding agents can reliably transform complex documents into accurate, complete, and structured outputs.

Clinical trial protocol digitization, which transforms protocol information into structured and machine-readable representations \citep{huser2015standardizing, babaeipour2026ai}, provides a high-stakes, real-world setting for investigating the capabilities and reliability of coding agents on long-document tasks. First, it is a representative long-document task. Clinical trial protocols specify the core design and operational requirements of a study, but are predominantly maintained as lengthy, unstructured PDF documents \citep{chan2025spirit,babaeipour2026ai}. These documents contain heterogeneous layouts, complex tables, and information distributed across multiple sections, requiring systems to navigate long documents, integrate scattered evidence, and reconstruct structured relationships.  


Second, it has direct practical value. When protocol requirements remain embedded in documents, they require repeated manual interpretation and translation into downstream systems, resulting in limited interoperability, reduced efficiency, and increased risk of inconsistencies and transcription errors \citep{babaeipour2026ai, boland2012elixr}. Structured protocol representation is essential for making study requirements computable, interoperable, and reusable across the clinical trial life cycle \citep{huser2015standardizing, richardson2025clinical, babaeipour2026ai}. Because these representations inform downstream trial operations, even small errors, such as altered eligibility thresholds or misplaced study activities, can change the intended study requirements.

\begin{figure*}[t]
    \centering
    \includegraphics[width=\textwidth]{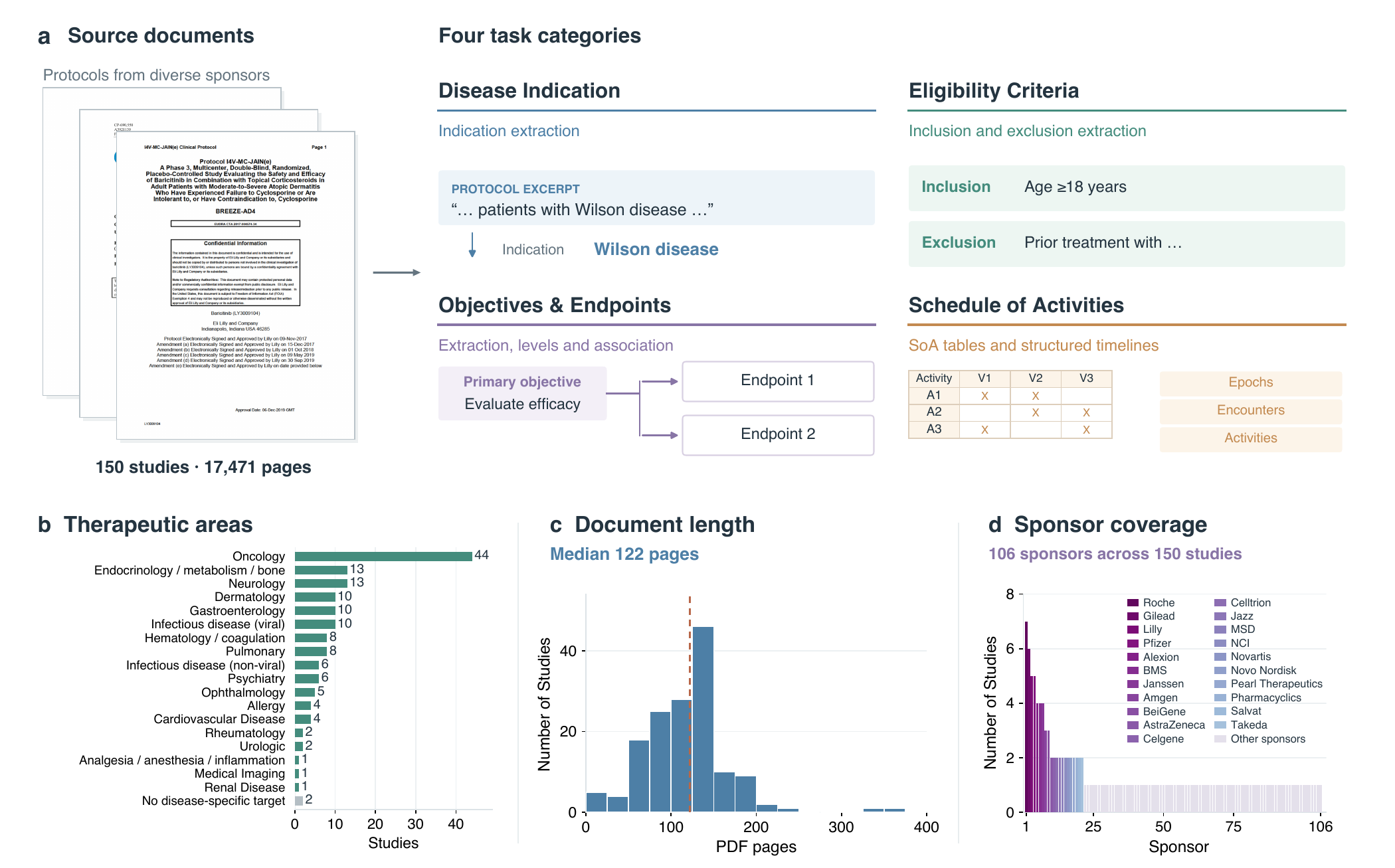}
    \caption{Overview of \ourbench{}. \textbf{(a)} Source documents and the four task categories. The corpus comprises 150 unique studies, using one representative document per NCT identifier. \textbf{(b)} Coverage of therapeutic areas. \textbf{(c)} Document-length distribution, with a median of 122 pages marked by the dashed line. \textbf{(d)} Study counts across 106 sponsors.}
    \label{fig:benchmark-overview}
    \vspace{-1em}
\end{figure*}

In this paper, we study how reliably coding agents handle long-document tasks, using clinical trial protocol digitization as a testbed (Figure~\ref{fig:overview}). Beyond final-answer performance, our evaluation protocol measures evidence recall from agent traces, agent behavior including vision use, tool errors, and interaction turns, and cost. Our contributions are as follows:

\begin{itemize}[leftmargin=*,topsep=3pt,itemsep=2pt,parsep=0pt]
\item We construct \ourbench{}, an expert-annotated benchmark of 150 complete protocol PDFs (median 122 pages) spanning four task categories: disease indication, eligibility criteria, objectives and endpoints, and schedule of activities. Each task requires locating evidence scattered across the full document and producing structured output that follows a USDM-derived schema.
\item We evaluate nine proprietary and open-weight models in five coding-agent harnesses. Even the strongest agents leave substantial errors, reaching at most 67.4\% endpoint F1 and 23.2\% encounter recall. Agents usually reach the required evidence pages but fail in the last mile of reading and structuring it. Agents with similar accuracy can also differ several-fold in cost.
\item We develop \ourmethod{}, a domain-specific harness that equips a locally served model with task skills and a verification loop. \ourmethod{} improves Qwen3.8-27B by 6.2-59.4 points across eight metrics, enabling it to match or surpass proprietary models such as GPT-5.6-Sol.
\end{itemize}

\section{\ourbench{}: Evaluating Coding Agents on Protocol Digitization}
\label{sec:benchmark}

\subsection{Task Definition}
We formulate clinical trial protocol digitization as extracting information from a protocol PDF into structured JSON output. The task-specific output schemas are derived from Unified Study Definitions Model (USDM) v4.0 \citep{cdisc_usdm}. CDISC's (Clinical Data Interchange Standards Consortium) USDM is an established standard to represent protocol documents into a structured machine-readable data. The output schemas cover selected portions relevant to four extraction tasks: disease indication, eligibility criteria, objectives and endpoints, and schedule of activities. Given a protocol document $D$ and the output schema $S_t$ for task $t$, an extraction system $f$ produces $\hat{Y}_t = f(D, S_t)$, where $\hat{Y}_t$ denotes the predicted JSON output. The objective is to recover the relevant information in $D$ accurately and completely while conforming to the fields and structure specified by $S_t$. 

\subsection{Benchmark Construction}
We construct \ourbench{} from clinical trial source documents associated with 150 unique NCT (National Clinical Trial) identifiers. (1) Disease indication extraction identifies the disease or condition investigated in the study. (2) Eligibility criteria extraction recovers participant inclusion and exclusion criteria. (3) Objectives and endpoints extraction identifies study objectives and their associated endpoints, together with their respective levels. (4) Schedule of activities (SoA) includes SoA table extraction and structured timeline extraction: the former locates the SoA tables in the full protocol and recovers their activity-visit structure, while the latter extracts study epochs, encounters, and activities. Reference outputs were produced by five experts in the clinical trial domain (Appendix~\ref{app:annotation}). We hold out the protocols used to develop \ourmethod{} (\S\ref{sec:trialscribe}) as a development split and report all results on the remaining test split.

\begin{table*}[t]
\centering
\footnotesize
\setlength{\tabcolsep}{4pt}
\renewcommand{\arraystretch}{1.0}
\begin{tabularx}{\textwidth}{@{}lrrrr>{\raggedright\arraybackslash}X>{\raggedright\arraybackslash}X@{}}
\toprule
& & \multicolumn{3}{c}{Evidence pages} & & \\
\cmidrule(lr){3-5}
Task & Samples & Mean & Median & p90 & Reference annotations & Main metrics \\
\midrule
\taskicon{taskind}{\faStethoscope}Disease indication & 40 & 1.00 & 1 & 1 & 41 indication descriptions; 1-2 per sample & Indication F1 \\
\addlinespace[3pt]
\taskicon{taskcrit}{\faUserCheck}Eligibility criteria & 134 & 4.32 & 4 & 7 & 3,974 criteria: 1,337 inclusion and 2,637 exclusion & Eligibility criteria F1 \\
\addlinespace[3pt]
\taskicon{taskoe}{\faBullseye}Objectives and endpoints & 89 & 3.17 & 3 & 5 & 665 objectives and 1,484 endpoints; 60 objectives without endpoints & Objective, endpoint, and objective-linked endpoint F1 \\
\addlinespace[3pt]
\addlinespace[3pt]
\taskicon{tasksoa}{\faTable}SoA tables & 37 & 1.89 & 1 & 4 & 70 tables; 10 protocols with multiple SoA pages & Activity-Visit and sub/sup F1 \\
\addlinespace[3pt]
\taskicon{tasksoa}{\faCalendar*}Structured timelines & 3 & 6.33 & 8 & 8 & 14 epochs; 69 encounters; 115 activities & Epoch accuracy; Activity and Encounter recall \\
\bottomrule
\end{tabularx}
\caption{Task-level sample sizes and annotation statistics on the test split, excluding the development protocols of Appendix~\ref{app:dev-protocols}. For every task, the input is the full protocol PDF. Evidence pages is the number of PDF pages per sample that contain the evidence needed for the reference annotations; p90 is the 90th percentile.}
\label{tab:task-statistics}
\vspace{-1em}
\end{table*}

\subsection{Overview and Statistics}
Figure~\ref{fig:benchmark-overview} summarizes the task categories and composition of the corpus, which comprises clinical trial source documents from ClinicalTrials.gov \citep{clinicaltrials_gov}, covering 18 therapeutic areas and 106 sponsors. Across these sponsors, the documents vary in section organization and table structure. For example, SoA tables may appear near the beginning of a protocol or in appendices, with different arrangements of study periods, visit windows, multi-level headers, and footnotes. This structural variation requires extraction systems to interpret information beyond fixed page positions or table layouts. In this dataset, protocol documents have a median length of 122 pages, with 98 documents (65.3\%) containing at least 100 pages, and depending on the task, up to 31.5\% of PDFs have task-relevant content that is missing from or garbled in the text layer and must be read from rendered pages (Appendix~\ref{app:pdf-text-accessibility}). Table~\ref{tab:task-statistics} reports the sample size and reference annotations of each task.

\subsection{Evaluation Metrics}
We evaluate coding agents along four dimensions: \textbf{(1) Final-answer Performance}, using each task’s scoring metrics to assess extraction quality; \textbf{(2) Evidence Recall}, using page recall to measure coverage of required evidence pages in observable agent traces; \textbf{(3) Agent Behavior}, measured by vision use rate on cases requiring visual recovery, average tool execution errors, and average agent response turns; and \textbf{(4) Resource Usage}, measured by average cost per run in USD. Detailed task specifications and evaluation metrics are provided in Appendix~\ref{app:tasks-and-metrics}.

\subsection{Experimental Setup}
We use Harbor~\citep{harbor_framework} to run coding agents in containerized task environments. Each environment provides the source document, task instructions, and output requirements. Agents write their extracted results to a designated output directory, which is then used for evaluation against reference annotations withheld during task execution.

We evaluate nine models, each paired with a coding-agent harness: GPT-5.6-Sol and GPT-6-Astra with Codex~\citep{openai2025codex}; Opus 4.6, Opus 4.8, Opus 5, and Sonnet 5 with Claude Code~\citep{anthropic_claude_code}; Kimi-K3 with Kimi Code~\citep{moonshot_kimi_code}; DeepSeek-V4-Flash with DeepSeek Harness~\citep{deepseek_harness}; and the open-weight Qwen3.8-27B with Pi harness~\citep{pi_coding_agent}.


\begin{table*}[t]
\centering
\footnotesize
\setlength{\tabcolsep}{2.5pt}
\begin{tabular}{@{}l@{\hspace{6pt}}w{c}{2.0em}w{c}{2.0em}@{\hspace{5.5pt}}w{c}{2.0em}w{c}{2.0em}@{\hspace{5.5pt}}cccc@{\hspace{5.5pt}}ccc@{\hspace{5.5pt}}cccc@{}}
\toprule
 & \multicolumn{2}{@{}c@{\hspace{5.5pt}}}{Indication} & \multicolumn{2}{@{}c@{\hspace{5.5pt}}}{Criteria} & \multicolumn{4}{@{}c@{\hspace{5.5pt}}}{Objectives \& endpoints} & \multicolumn{3}{@{}c@{\hspace{5.5pt}}}{SoA Tables} & \multicolumn{4}{@{}c@{}}{Timeline} \\
\cmidrule(l{4pt}r{9.5pt}){2-3}\cmidrule(l{4pt}r{9.5pt}){4-5}\cmidrule(l{4pt}r{9.5pt}){6-9}\cmidrule(l{4pt}r{9.5pt}){10-12}\cmidrule(l{4pt}r{4pt}){13-16}
Model & \textcolor{hdrfa}{F1} & \textcolor{hdrer}{ER} & \textcolor{hdrfa}{F1} & \textcolor{hdrer}{ER} & \textcolor{hdrfa}{Obj} & \textcolor{hdrfa}{End} & \textcolor{hdrfa}{Link} & \textcolor{hdrer}{ER} & \textcolor{hdrfa}{Sched.} & \textcolor{hdrfa}{Sub/sup} & \textcolor{hdrer}{ER} & \textcolor{hdrfa}{Epoch} & \textcolor{hdrfa}{Enc.} & \textcolor{hdrfa}{Act.} & \textcolor{hdrer}{ER} \\
\midrule
\makebox[1.35em][l]{\logoopenai}GPT-6-Astra & 82.5 & 95.0 & \textbf{87.8} & 91.6 & 82.2 & 65.0 & 58.1 & 89.6 & \textbf{86.8} & \textbf{84.5} & 84.8 & \textbf{73.3} & 20.3 & \textbf{91.1} & 98.3 \\
\makebox[1.35em][l]{\logoopenai}GPT-5.6-Sol & 80.0 & 92.5 & 87.1 & 94.9 & 80.4 & 63.6 & 54.7 & 88.7 & 83.7 & 81.0 & 93.6 & 61.1 & \textbf{23.2} & 88.1 & 100.0 \\[2.5pt]
\makebox[1.35em][l]{\logoanthropic}Opus 5 & \textbf{85.0} & 92.5 & 87.0 & 97.3 & 82.9 & \textbf{67.4} & \textbf{58.7} & 90.1 & 86.0 & 83.9 & 89.6 & 38.6 & 0.0 & 87.2 & 98.4 \\
\makebox[1.35em][l]{\logoanthropic}Opus 4.8 & 72.5 & 97.5 & 87.2 & 96.4 & 82.7 & 66.9 & 58.5 & 90.8 & 81.4 & 78.0 & 88.3 & 64.4 & 0.0 & 82.5 & 100.0 \\
\makebox[1.35em][l]{\logoanthropic}Opus 4.6 & 67.5 & 100.0 & 86.8 & 97.5 & 80.6 & 60.6 & 49.6 & 94.1 & 54.7 & 53.4 & 96.0 & 61.1 & 0.0 & 85.3 & 100.0 \\
\makebox[1.35em][l]{\logoanthropic}Sonnet 5 & 75.0 & 100.0 & 86.4 & 93.4 & 82.8 & 63.9 & 54.2 & 87.5 & 74.7 & 71.8 & 95.7 & 50.6 & 8.7 & 59.2 & 100.0 \\[2.5pt]
\makebox[1.35em][l]{\logokimi}Kimi-K3 & 80.0 & 87.5 & 86.9 & 89.5 & \textbf{84.9} & 64.6 & 56.4 & 72.4 & 83.9 & 82.0 & 87.9 & 61.1 & 21.7 & 50.2 & 100.0 \\[2.5pt]
\makebox[1.35em][l]{\logodeepseek}DeepSeek-V4-Flash & 75.0 & 85.0 & 80.2 & 79.7 & 63.3 & 58.5 & 36.8 & 72.7 & 47.5 & 54.4 & 77.9 & 48.5 & 2.9 & 72.2 & 100.0 \\[2.5pt]
\makebox[1.35em][l]{\logoqwen}Qwen3.8-27B & 68.3 & 85.0 & 67.1 & 89.2 & 56.0 & 46.7 & 32.1 & 75.1 & 32.5 & 42.5 & 80.2 & 22.9 & 5.8 & 82.5 & 98.6 \\
\bottomrule
\end{tabular}
\caption{Main results (\%) of coding agents on \ourbench{}. Each task reports its \textcolor{hdrfa}{final-answer metrics} followed by \textcolor{hdrer}{evidence recall (ER)}, the recall of the required evidence pages visited in the agent trace. Obj/End/Link: objective, endpoint, and objective-linked endpoint F1; Sched.\ and Sub/sup: Activity-Visit F1 (scheduling marks) and Sub/sup F1; Epoch: epoch type positional accuracy; Enc.: encounter recall; Act.: activity recall.}
\label{tab:main-results}
\end{table*}

\begin{table*}[t]
\centering
\footnotesize
\setlength{\tabcolsep}{4pt}
\begin{tabular*}{\textwidth}{@{\extracolsep{\fill}}l@{\hspace{14pt}}ccccc@{\hspace{10pt}}ccccc@{}}
\toprule
 & \multicolumn{5}{@{}c@{\hspace{10pt}}}{Tool errors per run} & \multicolumn{5}{@{}c@{}}{Turns per run} \\
\cmidrule(l{4pt}r{14pt}){2-6}\cmidrule(l{4pt}r{4pt}){7-11}
Model & Ind. & Crit. & O\&E & SoA & TL & Ind. & Crit. & O\&E & SoA & TL \\
\midrule
\makebox[1.35em][l]{\logoopenai}GPT-6-Astra & 0.17 & 0.69 & 0.78 & 1.05 & 1.33 & 4.9 & 10.6 & 11.5 & 16.0 & 27.3 \\
\makebox[1.35em][l]{\logoopenai}GPT-5.6-Sol & 0.68 & 1.71 & 1.90 & 0.62 & 2.33 & 7.1 & 15.9 & 15.3 & 19.5 & 20.3 \\[2.5pt]
\makebox[1.35em][l]{\logoanthropic}Opus 5 & 1.18 & 0.27 & 1.37 & 1.92 & 0.67 & 7.6 & 14.2 & 15.6 & 44.7 & 18.7 \\
\makebox[1.35em][l]{\logoanthropic}Opus 4.8 & 0.28 & 0.21 & 0.65 & 1.95 & 0.67 & 4.7 & 12.2 & 14.2 & 43.2 & 15.0 \\
\makebox[1.35em][l]{\logoanthropic}Opus 4.6 & 0.00 & 0.62 & 0.94 & 0.27 & 0.00 & 4.1 & 12.1 & 7.1 & 19.7 & 6.7 \\
\makebox[1.35em][l]{\logoanthropic}Sonnet 5 & 0.15 & 0.60 & 1.52 & 2.19 & 2.67 & 5.4 & 17.4 & 21.1 & 69.4 & 29.3 \\[2.5pt]
\makebox[1.35em][l]{\logokimi}Kimi-K3 & 0.25 & 0.28 & 1.03 & 1.16 & 2.67 & 7.0 & 15.1 & 18.4 & 33.6 & 16.0 \\[2.5pt]
\makebox[1.35em][l]{\logodeepseek}DeepSeek-V4-Flash & 1.25 & 1.33 & 2.64 & 5.69 & 3.00 & 10.4 & 26.8 & 25.2 & 64.2 & 34.7 \\[2.5pt]
\makebox[1.35em][l]{\logoqwen}Qwen3.8-27B & 0.95 & 1.81 & 4.96 & 4.73 & 1.33 & 12.8 & 25.3 & 43.4 & 48.2 & 25.7 \\
\bottomrule
\end{tabular*}
\caption{Agent behavior per task: average number of tool execution errors and agent response turns per run. Ind.: indication; Crit.: eligibility criteria; O\&E: objectives and endpoints; SoA: SoA tables; TL: structured timeline.}
\label{tab:behavior-errors-turns}
\end{table*}

\begin{table}[t]
\centering
\footnotesize
\setlength{\tabcolsep}{4pt}
\begin{tabular}{@{}lcccc@{}}
\toprule
 & \multicolumn{4}{c}{Vision use (\%)} \\
\cmidrule(l{4pt}){2-5}
Model & Ind. & Crit. & O\&E & SoA \\
\midrule
\makebox[1.35em][l]{\logoopenai}GPT-6-Astra & 100.0 & 100.0 & 100.0 & 100.0 \\
\makebox[1.35em][l]{\logoopenai}GPT-5.6-Sol & 71.4 & {97.5} & {92.9} & 100.0 \\[2pt]
\makebox[1.35em][l]{\logoanthropic}Opus 5 & 85.7 & 100.0 & {67.9} & 100.0 \\
\makebox[1.35em][l]{\logoanthropic}Opus 4.8 & 85.7 & 100.0 & {85.7} & 100.0 \\
\makebox[1.35em][l]{\logoanthropic}Opus 4.6 & 100.0 & 100.0 & 85.7 & 100.0 \\
\makebox[1.35em][l]{\logoanthropic}Sonnet 5 & 100.0 & 100.0 & {82.1} & 100.0 \\[2pt]
\makebox[1.35em][l]{\logokimi}Kimi-K3 & 71.4 & {97.5} & {82.1} & 100.0 \\[2pt]
\makebox[1.35em][l]{\logodeepseek}DeepSeek-V4-Flash & {42.9} & {72.5} & {64.3} & {75.0} \\[2pt]
\makebox[1.35em][l]{\logoqwen}Qwen3.8-27B & {71.4} & {70.0} & {46.4} & 62.5 \\
\bottomrule
\end{tabular}
\caption{Vision use (\%): the share of vision-required cases in which the agent obtains an image or OCR of a relevant hard page. Structured timeline has no vision-required protocol. Task abbreviations as in Table~\ref{tab:behavior-errors-turns}.}
\label{tab:behavior-vision}
\vspace{-1em}
\end{table}

\begin{table}[t]
\centering
\footnotesize
\setlength{\tabcolsep}{4pt}
\begin{tabular}{@{}lccccc@{}}
\toprule
 & \multicolumn{5}{c}{Cost per run (USD)} \\
\cmidrule(l{4pt}){2-6}
Model & Ind. & Crit. & O\&E & SoA & TL \\
\midrule
\makebox[1.35em][l]{\logoopenai}GPT-6-Astra & 0.32 & 1.35 & 1.07 & 2.13 & 5.59 \\
\makebox[1.35em][l]{\logoopenai}GPT-5.6-Sol & 0.14 & 0.50 & 0.48 & 0.85 & 1.35 \\[2pt]
\makebox[1.35em][l]{\logoanthropic}Opus 5 & 0.21 & 0.73 & 0.75 & 2.93 & 3.08 \\
\makebox[1.35em][l]{\logoanthropic}Opus 4.8 & 0.12 & 0.71 & 0.89 & 4.12 & 1.61 \\
\makebox[1.35em][l]{\logoanthropic}Opus 4.6 & 0.13 & 0.83 & 1.16 & 4.49 & 3.06 \\
\makebox[1.35em][l]{\logoanthropic}Sonnet 5 & 0.08 & 0.40 & 0.52 & 2.44 & 1.30 \\[2pt]
\makebox[1.35em][l]{\logokimi}Kimi-K3 & 0.11 & 0.44 & 0.63 & 2.55 & 0.92 \\[2pt]
\makebox[1.35em][l]{\logodeepseek}DeepSeek-V4-Flash & 0.01 & 0.04 & 0.04 & 0.08 & 0.11 \\[2pt]
\makebox[1.35em][l]{\logoqwen}Qwen3.8-27B & 0.04 & 0.20 & 0.46 & 0.46 & 0.25 \\
\bottomrule
\end{tabular}
\caption{Average cost per run in USD. Task abbreviations as in Table~\ref{tab:behavior-errors-turns}.}
\label{tab:behavior-cost}
\vspace{-1em}
\end{table}



\subsection{Main Results}
Table~\ref{tab:main-results} reports final-answer performance and evidence recall for all coding agents. Tables~\ref{tab:behavior-errors-turns} and~\ref{tab:behavior-vision} report agent behavior, and Table~\ref{tab:behavior-cost} reports the average cost per run.

\textbf{Finding 1: Task difficulty varies widely, and even the strongest coding agents do not fully solve protocol digitization.}
Frontier agents largely handle tasks that mainly require locating and transcribing contiguous text or tables. The best agent reaches 87.8\% F1 on eligibility criteria, 86.8\% Activity-Visit F1 on SoA tables, 85.0\% indication F1, and 84.9\% objective F1. Components that require linking information across sections or mapping it onto the study design are considerably harder. The best endpoint F1 is 67.4\% and the best objective-linked endpoint F1 is 58.7\%, both well below objective F1. On structured timelines, agents recover activities with up to 91.1\% recall, yet epoch accuracy peaks at 73.3\% and encounter recall at 23.2\%, and the three Opus agents recover no reference encounter. The gaps between models also depend on the task. On eligibility criteria, seven agents lie within 1.4 points of each other (86.4--87.8\%), whereas Activity-Visit F1 on SoA tables ranges from 32.5\% to 86.8\%. Apart from timeline activity recall, no final-answer metric exceeds 88\%. Because extracted study definitions feed directly into downstream trial systems, the remaining errors are consequential. Directly deploying current frontier coding agents therefore does not yet provide complete and reliable protocol digitization, and the largest gaps lie in the relational components of a protocol rather than in its surface content.

\textbf{Finding 2: Locating the evidence is not the main bottleneck, and high evidence recall does not translate into accurate extraction.}
Agents generally reach the pages that contain the required evidence. Except for Kimi-K3 on objectives and endpoints (72.4\%), the seven frontier agents visit at least 84.8\% of the required evidence pages on every task. Final-answer performance, however, does not follow evidence recall. The clearest case is structured timelines, where every agent visits at least 98.3\% of the evidence pages, yet encounter recall stays at or below 23.2\%. Across models, visiting more evidence pages does not yield better extraction. Opus 4.6 has the highest SoA evidence recall of any agent (96.0\%) but reaches only 54.7\% Activity-Visit F1, whereas GPT-6-Astra has the lowest among frontier agents (84.8\%) and achieves the best score (86.8\%). Similarly, Opus 4.6 and Sonnet 5 visit every indication evidence page yet reach 67.5\% and 75.0\% F1, below Opus 5 at 85.0\% with 92.5\% evidence recall. Across the seven frontier agents, the ranking by evidence recall does not track the ranking by final-answer performance. The main bottleneck therefore lies after localization: agents must read the evidence faithfully and reconstruct its structure, which we analyze in \S\ref{sec:failure-analysis}.

\textbf{Finding 3: Tool errors and interaction turns scale with task complexity.}
Agent behavior varies systematically across tasks (Table~\ref{tab:behavior-errors-turns}). Averaged over the nine agents, indication extraction takes 7.1 turns and incurs 0.55 tool errors per run, whereas SoA table extraction takes 39.8 turns and incurs 2.18 errors, a 5.6-fold and 4.0-fold increase, respectively. The average number of turns rises from indication (7.1) to eligibility criteria (16.6), objectives and endpoints (19.1), structured timelines (21.5), and SoA tables (39.8). This order follows the structural complexity of the output. Weaker models amplify this effect. DeepSeek-V4-Flash and Qwen3.8-27B incur 2.64--5.69 tool errors per run on objectives and endpoints and on SoA tables, compared with at most 2.19 for the frontier agents. They also obtain visual evidence less often when it is required ({42.9--75.0\%}, Table~\ref{tab:behavior-vision}), whereas every frontier agent does so in at least {67.9\%} of such cases and in all SoA cases. Longer trajectories do not produce better answers. On SoA tables, Sonnet 5 takes 69.4 turns to reach 74.7\% Activity-Visit F1, while GPT-6-Astra reaches 86.8\% in 16.0 turns. Turns and errors therefore indicate task difficulty and execution reliability rather than extraction quality.

\textbf{Finding 4: Similar final-answer performance can come at very different costs.}
Because agents differ in both trajectory length and token price, comparable accuracy can carry very different costs (Table~\ref{tab:behavior-cost}). On eligibility criteria, the seven frontier agents lie within 1.4 F1 points of each other, while the cost per run ranges from \$0.40 (Sonnet 5) to \$1.35 (GPT-6-Astra), a 3.4-fold difference. Kimi-K3 (86.9\%, \$0.44) and GPT-5.6-Sol (87.1\%, \$0.50) are both within one point of the best agent at about a third of its cost. On SoA tables, GPT-5.6-Sol reaches 83.7\% Activity-Visit F1 at \$0.85 per run, within 3.1 points of GPT-6-Astra at 40\% of its cost (\$2.13). In contrast, Opus 4.6 is the most expensive agent on this task (\$4.49) yet scores only 54.7\%. On objectives and endpoints, Kimi-K3 attains the highest objective F1 (84.9\%) at \$0.63 per run, below GPT-6-Astra (\$1.07) and Opus 4.6 (\$1.16). DeepSeek-V4-Flash marks the low-cost extreme: it reaches 80.2\% criteria F1 at \$0.04 per run, 34 times cheaper than GPT-6-Astra. Cost is thus not a proxy for quality. For tasks on which frontier agents perform similarly, substantially cheaper configurations can be chosen with little loss in performance.

\begin{capabilitybox}
\textbf{Takeaways.}
\begin{itemize}[leftmargin=*,topsep=3pt,itemsep=3pt,parsep=0pt]
\item Frontier coding agents reliably locate protocol evidence and recover much of its surface content, but still fall short on relational structure such as endpoint linkage and study timelines.
\item Turns and tool errors vary with the task and peak on SoA tables, but within a task, longer trajectories are not associated with higher accuracy.
\item When agents perform similarly, cheaper configurations can deliver comparable accuracy.
\end{itemize}
\end{capabilitybox}

\section{Further Analysis}
\label{sec:analysis}
This section examines how agent performance changes across model releases and where agents fail. Appendix~\ref{app:additional-results} reports further analyses, including comparisons with OCR and document-understanding models and with a RAG pipeline, eligibility criteria by category, and results by PDF text accessibility.

\subsection{Performance across Model Releases}
\label{sec:release-trend}

\begin{figure}[t]
    \centering
    \includegraphics[width=\columnwidth]{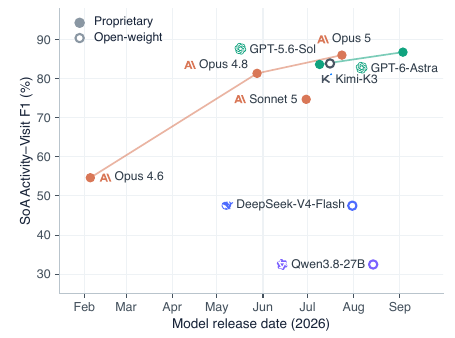}
    \caption{SoA Activity-Visit F1 on complete protocols (Table~\ref{tab:main-results}) against model release date. Lines connect models of the same family; hollow markers denote open-weight models.}
    \label{fig:soa-release}
    \vspace{-1em}
\end{figure}

We analyze how SoA table extraction performance changes across model releases, plotting document-level Activity-Visit F1 against each model's release date (Figure~\ref{fig:soa-release}). Performance improves quickly within model families: Opus rises from 54.7\% for Opus 4.6 (February 2026) to 86.0\% for Opus 5 (July 2026). Since July, frontier agents have clustered between 83.7\% and 86.8\%, still short of full accuracy. Open-weight models released in the same period diverge: Kimi-K3 (83.9\%) matches proprietary agents, whereas DeepSeek-V4-Flash (47.5\%) and Qwen3.8-27B (32.5\%) fall far behind.

\subsection{Failure Analysis}
\label{sec:failure-analysis}

\begin{figure*}[t]
    \centering
    \includegraphics[width=\textwidth]{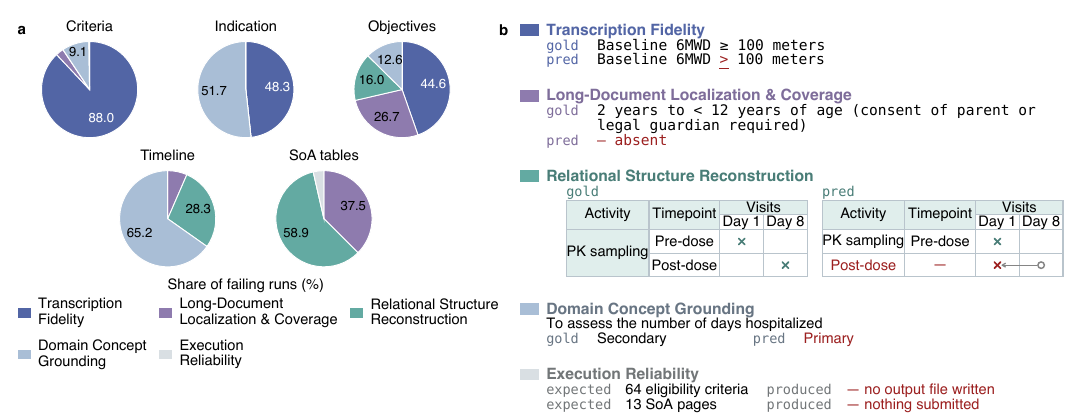}
    \caption{Failure analysis across protocol digitization tasks. \textbf{(a)} Each pie shows the distribution of failure categories among failing runs within one task; labels indicate shares of at least 8\%. \textbf{(b)} Illustrative examples of the five main failure categories, with incorrect or missing predictions highlighted in red.}
    \label{fig:failure-profile}
    \vspace{-1em}
\end{figure*}

To understand where coding agents fail, we classify each failing agent run, defined as a run with at least one detected defect. These failures fall mainly into five categories: transcription fidelity, long-document localization and coverage, relational structure reconstruction, domain concept grounding, and execution reliability. As Figure~\ref{fig:failure-profile}a shows, the dominant failure category differs markedly across tasks.

For eligibility criteria, transcription fidelity accounts for 88.0\% of failing runs: agents locate and read the eligibility section but do not reproduce it exactly. These deviations can change the clinical meaning. Among near-verbatim transcriptions, 167 alter a comparison operator and 308 alter a number, for example rendering ``$\geq$ 100 meters'' as ``$>$ 100 meters'' (Figure~\ref{fig:failure-profile}b). For indication extraction, domain concept grounding is the largest category (51.7\%), reflecting failures to identify the condition under study. DeepSeek-V4-Flash, for example, returns ``Non-Valvular Atrial Fibrillation'' for a study whose indication is recurrent embolic stroke. For objectives and endpoints, localization and coverage account for a substantial share (26.7\%), because endpoints are distributed across the synopsis, main text, and tables. For SoA tables, relational structure accounts for 58.9\% of failing runs, most often because agents reconstruct the table but place scheduling marks in the wrong cells; a single misresolved merged cell can shift every subsequent mark on a page. Most of the remaining failing runs (37.5\%) are localization failures, in which agents miss the annotated SoA pages.

Structured timeline extraction is dominated by domain concept grounding (65.2\%): agents often identify the right visits but do not describe them with the expected encounter categories. Although the task instruction explicitly asks agents to follow USDM v4.0, they do not apply its controlled terminology, suggesting that agents recognize the standard but lack knowledge of its detailed specifications. Execution reliability, by contrast, accounts for at most 3.6\% of failing runs in any task: the nine agents almost always produce usable output, and their errors lie in its content. 

\section{\ourmethod{}: A Domain-Specific Harness for Protocol Digitization}
\label{sec:trialscribe}

Our analysis points to two practical obstacles. First, protocols are often confidential, which favors compact open-weight models (e.g., Qwen3.8-27B) that run locally in private environments, yet these models fall behind proprietary agents (\S\ref{sec:release-trend}). Second, most failures arise after the evidence is found, from unfaithful transcription, misplaced structure, and unfamiliar domain conventions (\S\ref{sec:failure-analysis}). We therefore introduce \ourmethod{}, a domain-specific harness layer that equips an existing coding agent with task skills, which supply domain knowledge and output conventions, and a verification loop, which audits the output against the source. We examine whether \ourmethod{} allows a locally served open-weight model to match proprietary coding agents.

\subsection{Design}
\textbf{Task skills.} For each task, \ourmethod{} provides a skill, a SKILL.md file placed in the agent's environment and registered with the harness. The agent reads it at the start of the task and then decides when and how to apply its guidance and tools as it works. Each skill encodes the domain knowledge and working procedures that the task requires: how protocols typically organize and cross-reference the relevant information, which conventions determine a correct output, and which checks confirm that the output is complete and faithful to the source.

\textbf{Verification loop.} After the agent writes a draft, a fresh agent session audits it against the protocol with read-only access to both the protocol and the candidate output. The auditor reports concrete errors together with their source pages and suggested corrections, and a gold-free gate checks the report for schema validity and internal consistency. If errors remain, the agent applies minimal source-supported fixes, and the revised output is audited again, for at most two rounds. For SoA table extraction, the agent instead verifies its output within its own session (Appendix~\ref{app:verification}).

\subsection{Results}
We instantiate \ourmethod{} with Qwen3.8-27B and Pi harness, served locally on two NVIDIA A100 GPUs.

\textbf{\ourmethod{} narrows the gap between a locally served agent and proprietary coding agents.} As Figure~\ref{fig:overview}(e) shows, \ourmethod{} improves Qwen3.8-27B on every evaluated task, e.g., 15.2\% on eligibility criteria F1, 22.2\% on objective F1, 18.8\% on endpoint F1, and by 37.1\% on epoch accuracy, 6.2\% on activity recall, and 59.4\% on encounter recall. With \ourmethod{}, the local model surpasses GPT-5.6-Sol on endpoint extraction (65.5\% vs.\ 63.6\%) and comes within 1.9\% of Opus-5 on endpoints. On timeline extraction, it exceeds both proprietary agents on activity recall (88.6\% vs.\ 88.1\% and 87.2\%) and encounter recall (65.2\% vs.\ 23.2\% and 0.0\%), and trails GPT-5.6-Sol on epoch accuracy (60.0\% vs.\ 61.1\%).

\textbf{Verification helps most where transcription fidelity dominates.} Comparing the draft and final outputs of the same runs, we find that verification raises eligibility criteria F1 from 72.2\% to 82.3\%, objective F1 from 73.2\% to 78.2\%, and endpoint F1 from 62.0\% to 65.5\%, but leaves indication F1 unchanged at 77.5\%. This pattern mirrors the failure profiles in Figure~\ref{fig:failure-profile}: eligibility criteria fail mainly through transcription errors that a source-grounded audit can detect.

\textbf{Structural and standard-specific gaps remain, but they are likely to narrow as open-weight models improve.} {On SoA tables, \ourmethod{} reaches 75.1\% Activity-Visit F1 and 78.6\% Sub/sup F1, but still trails GPT-5.6-Sol and Opus 5 in Activity-Visit F1 (75.1\% vs.\ 83.7\% and 86.0\%),} suggesting that reconstructing complex table structure remains limited by the underlying model. Nevertheless, open-weight models have improved rapidly~\citep{edwards2026openmodels}, and because \ourmethod{} does not rely on a particular model, it can be applied directly to stronger ones, which we expect to close the remaining gaps.

\section{Related Work}
Prior work on protocol digitization structures eligibility criteria for cohort queries \citep{kang2017eliie,yuan2019criteria2query,park2024criteria2query} or interprets schedule-of-activities tables \citep{dhuliawala2018happens,kumar2024selective}, mostly on registry text; \citet{babaeipour2026ai} cover all four categories with a RAG system on 23 unreleased protocols. Document parsers \citep{cui2025paddleocrvlboostingmultilingualdocument,poznanski2025olmocr} and benchmarks \citep{ouyang2025omnidocbench,zhang2026extractbench} evaluate layout recovery and schema-guided extraction, while coding agents \citep{wang2024executable,yang2024swe,soni2026coding} and agent skills \citep{li2026skillsbench} increasingly reach beyond software engineering. See Appendix~\ref{app:related} for more details.

\section{Conclusion}
We introduce \ourbench{} and find that even the strongest coding agents usually locate the required evidence but fail in the last mile of transcribing and structuring it. We further develop \ourmethod{}, which enables a locally served open-weight model to match or surpass proprietary agents such as GPT-5.6-Sol.

\section*{Limitations}
We use clinical trial protocol digitization as a testbed for studying coding agents on long-document tasks, so our findings are bounded by this setting. At the same time, the properties that make protocols challenging, namely their length, heterogeneous layouts, complex tables, and information distributed across sections, are shared by many other long documents, such as regulatory filings, contracts, and technical specifications. We therefore expect some of our findings, in particular that agents usually locate the required evidence but lose accuracy when transcribing and structuring it, to be informative beyond this setting.

Moreover, structured timeline extraction remains challenging for current coding agents, and our evaluation of this task includes only three protocols. The structured timeline also underlies many downstream tasks, such as main and sub timeline extraction, which are beyond the scope of this paper.

\section*{Ethics Statement}
All protocol documents in \ourbench{} are publicly available on ClinicalTrials.gov, and we use them as released without collecting any additional data. The reference annotations were produced by the authors; no external annotators were recruited or compensated, and the work involves no human-subject research. We do not foresee significant ethical risks from this work.

\bibliography{custom}
\clearpage
\appendix
\etocdepthtag.toc{mtappendix}
\etocsettagdepth{mtmain}{none}
\etocsettagdepth{mtappendix}{subsection}
\etocsettocstyle{\section*{Appendix Contents}}{}
\tableofcontents
\section{Background}
\subsection{Coding Agents}
A coding agent combines a language model with an agent harness. The model interprets task instructions, reasons over the available context, and generates code or tool calls. The harness is the surrounding software infrastructure that manages instructions and context, provides tool interfaces, and coordinates interaction with the execution environment. Typical tools support file access, search, edit, and code execution, while the harness manages the execution loop and associated controls.

These components operate through an iterative interaction loop: the model selects an action, the harness executes the corresponding tool call and returns its result. The model then uses this feedback to determine subsequent actions. This process continues until the agent produces a final response or an execution limit is reached. The resulting behavior depends on both the underlying model and the harness that provides the operational layer for the model.

\section{Additional Related Work}
\label{app:related}

\paragraph{Clinical Trial Protocol Digitization.}
Research on protocol digitization spans semantic parsing, supervised extraction, and LLM-based methods. EliIE \citep{kang2017eliie} and Criteria2Query \citep{yuan2019criteria2query} transform eligibility text into structured representations for clinical database querying, while BERT-based approaches learn clinically relevant entities and relations \citep{chen2020joint,liu2021clinical}. Chia \citep{kury2020chia} and LCT \citep{dobbins2022leaf} provide annotated registry eligibility text for developing these capabilities. AutoCriteria \citep{datta2024autocriteria} and Criteria2Query 3.0 \citep{park2024criteria2query} subsequently introduce LLMs for criteria extraction and query generation. Beyond eligibility, \citet{dhuliawala2018happens} interpret schedule-of-activities tables through structural and semantic analysis, and \citet{kumar2024selective} study schedule-of-event table detection. \citet{babaeipour2026ai} address broader extraction from complete protocol PDFs through a RAG system. Our work complements these approaches by evaluating general-purpose coding agents across multiple protocol extraction tasks and diagnosing their remaining failures.

\paragraph{Document Understanding and Structured Extraction.}
Document understanding research addresses the recovery of text, layout, and tabular structure from heterogeneous documents. PubTables-1M \citep{smock2022pubtables} supports table detection and structure recognition, while OmniDocBench \citep{ouyang2025omnidocbench} evaluates document parsing across diverse layouts and content types. Recent models, including PaddleOCR-VL \citep{cui2025paddleocrvlboostingmultilingualdocument} and olmOCR 2 \citep{poznanski2025olmocr}, advance these parsing capabilities. At the document-processing level, DocAgent \citep{sun2025docagent} combines interactive navigation with evidence review. ExtractBench \citep{zhang2026extractbench} further evaluates schema-guided enterprise extraction, including coding agents. Our benchmark focuses on clinical protocol requirements, evaluating document recovery, the application of domain knowledge, and the extraction of study concepts and relationships, including activity-visit associations and objective-endpoint links.

\paragraph{Coding Agents and Domain Support.}
CodeAct \citep{wang2024executable} and SWE-agent \citep{yang2024swe} demonstrate the importance of executable actions and agent-computer interfaces for interactive problem solving. OpenHands-Versa \citep{soni2026coding} further investigates how coding agents equipped with general tools transfer across software engineering, research, and browsing tasks. Domain support is explored through plugins and examples in TaskWeaver \citep{qiao2023taskweaver}, and through clinical programming skills in ClinAgent \citep{yan2026clinagent}. SkillsBench \citep{li2026skillsbench} evaluates the benefits of curated skills across domains, while Self-Refine \citep{madaan2023self} and Reflexion \citep{shinn2023reflexion} investigate feedback-guided revision and adaptation. Building on these directions, we examine coding-agent reliability in clinical trial protocol digitization and introduce \ourmethod{}, which combines domain skills with a verification loop.

\section{Benchmark Details}
\label{app:benchmark}

\subsection{Comparison with Prior Work}
\label{app:prior-work}
Table~\ref{tab:protocol-comparison} compares \ourbench{} with prior datasets and systems for clinical trial protocol digitization. Existing resources mostly target a single information type, typically eligibility criteria, and often use registry or publication text rather than complete protocol documents. The system of \citet{babaeipour2026ai} covers all four task categories from protocol PDFs but evaluates 23 protocols and does not release its data. \ourbench{} covers all four categories on 150 complete protocol PDFs.

\begin{table*}[t]
\centering
\begingroup
\small
\setlength{\tabcolsep}{5pt}
\renewcommand{\arraystretch}{1.22}
\resizebox{\textwidth}{!}{%
\begin{tabular}{@{}llrccccc@{}}
\toprule
\textbf{Work} & \textbf{Input} & \textbf{Size (unit)} & \textbf{Disease} & \textbf{Eligibility} & \textbf{Objectives \&} & \textbf{Schedule of} & \textbf{Open} \\
 & & & \textbf{indication} & \textbf{criteria} & \textbf{endpoints} & \textbf{activities} & \textbf{source} \\
\midrule
Chia \citep{kury2020chia} & Registry text & 1,000 trials & \PDnone & \PDyes & \PDnone & \PDnone & \PDyes \\
LCT \citep{dobbins2022leaf} & Registry text & 1,006 trials & \PDnone & \PDyes & \PDnone & \PDnone & \PDyes \\
AutoCriteria \citep{datta2024autocriteria} & Registry text/XML & 180 trials & \PDnone & \PDyes & \PDnone & \PDnone & \PDno \\
Kumar et al. \citep{kumar2024selective} & Protocol PDFs & 91 protocols & \PDnone & \PDnone & \PDnone & \PDpart & \PDno \\
SPIRIT-CONSORT-TM \citep{jiang2025spirit} & Publication text & 200 publications & \PDpart & \PDpart & \PDpart & \PDpart & \PDyes \\
Babaeipour et al. \citep{babaeipour2026ai} & Protocol PDFs & 23 protocols & \PDyes & \PDyes & \PDyes & \PDyes & \PDno \\
\midrule
\textbf{\ourbench{} (Ours)} & \textbf{Protocol PDFs} & \textbf{150 protocols} & \PDyes & \PDyes & \PDyes & \PDyes & \PDyes \\
\bottomrule
\end{tabular}%
}

\endgroup
\caption{Comparison with prior datasets and systems for clinical trial protocol digitization. \PDyes\ indicates explicit coverage, \PDpart\ partial or related-task coverage, and \PDno\ no reported coverage.}

\label{tab:protocol-comparison}
\end{table*}

\subsection{Corpus Construction}
\label{app:corpus}
\paragraph{Source documents.} The corpus consists of clinical trial protocol documents associated with 150 unique NCT identifiers on ClinicalTrials.gov \citep{clinicaltrials_gov}. Each task uses the subset of studies for which reference annotations are available, so a study can contribute to several tasks (Table~\ref{tab:task-statistics}).

\paragraph{Document statistics.} The 150 representative documents contain 17,471 pages in total, with a median of 122 pages and a range of 2 to 368 pages. Ninety-eight documents (65.3\%) contain at least 100 pages, five contain at least 200 pages, and five contain fewer than 20 pages. Sixteen documents (10.7\%) have no extractable text layer, so their content can be recovered only from rendered pages or OCR.

\paragraph{Therapeutic areas and sponsors.} We categorize each study by its primary indication using therapeutic-area categories adapted from the FDA's Spectrum of Diseases/Conditions \citep{fda_therapeutic_areas}. Assignments follow the principal studied indication and trial intent rather than exclusions, drug mechanism, or sponsor; tumor-directed therapy, including hematologic malignancy, maps to Oncology, and prevention trials map to the disease being prevented. The 150 studies populate 18 FDA categories, and two healthy-volunteer studies without a disease-specific target are reported separately. Sponsor names are taken from registry metadata; after merging duplicate spellings of the same sponsor name, the corpus contains 106 sponsor labels.

\subsection{Annotation Process}
\label{app:annotation}
\paragraph{Annotators and workflow.} Reference annotations were produced by five experts in the clinical trial domain. Each task was annotated and reviewed separately, on its own set of protocols, so the number of instances differs across tasks (Table~\ref{tab:task-statistics}). For each protocol and task, Sonnet-4.6 running in Claude Code first produced a draft extraction from the source PDF, which served only as a starting point. An expert then checked every field of the draft against the PDF, corrected errors, added missing items, and removed unsupported ones. Verbatim fields were taken from the PDF text, or read from the rendered page when the text layer was missing or corrupted, and no draft text was kept without being checked against the source. The lead annotator then reviewed each annotated protocol against the PDF and made the final decision whenever the annotation needed to be changed. This process was used for all tasks except the epochs, encounters, and activities of structured timeline extraction, whose references are taken from the official USDM example studies published by CDISC~\citep{cdisc_usdm}.

\paragraph{Guidelines.} Each task has its own annotation guideline. Table~\ref{tab:annotation-runbook} lists the steps that annotators followed: general steps that familiarize the annotator with the study, performed for every protocol, followed by the steps of the task being annotated. The rules below define a correct annotation.
\begin{itemize}[leftmargin=*,topsep=2pt,itemsep=1pt,parsep=0pt]
\item \textbf{Indication.} Record the disease or condition under study in the protocol's own wording, keeping disease-defining qualifiers such as stage, severity, or biomarker status and omitting enrolment wording such as ``patients with''. Distinct diseases are separate entries, whereas a single disease with several qualifiers is one entry.
\item \textbf{Eligibility criteria.} Use the detailed eligibility section rather than the synopsis, and include cohort-specific criteria. Each top-level numbered criterion is one item, with its sub-items and any laboratory-threshold table kept inside it. Text is transcribed verbatim, including thresholds, units, and comparison symbols.
\item \textbf{Objectives and endpoints.} Record every objective with its level, normalized to primary, secondary, or exploratory, and nest under it the endpoints that measure it. An objective without a stated endpoint receives none.
\item \textbf{Schedule of Activities.} Record a mark for every activity-visit cell, resolving merged cells and multi-level visit headers. Footnote markers are kept as superscripts on the row, cell, or visit they belong to.
\end{itemize}

\begin{table*}[t]
\centering
\small
\setlength{\tabcolsep}{3pt}
\begin{tabular}{@{}l>{\raggedright\arraybackslash}p{0.22\textwidth}>{\raggedright\arraybackslash}p{0.34\textwidth}>{\raggedright\arraybackslash}p{0.34\textwidth}@{}}
\toprule
Step & Section & Action & Notes \\
\midrule
\multicolumn{4}{@{}l}{\textit{General (all tasks)}} \\
1 & Title page & Extract titles & At least the brief title \\
2 & Title page & Extract identifiers & Internal and NCT identifiers \\
3 & Title page & Read the therapeutic area & Preferably codeable \\
4 & Title page & Read the interventions & \\
\midrule
\multicolumn{4}{@{}l}{\textit{Indication}} \\
1 & Title page, synopsis & Identify the indication & \\
\midrule
\multicolumn{4}{@{}l}{\textit{Eligibility criteria}} \\
1 & Eligibility criteria & Read the inclusion criteria & \\
2 & Eligibility criteria & Read the exclusion criteria & \\
\midrule
\multicolumn{4}{@{}l}{\textit{Objectives and endpoints}} \\
1 & Objectives and endpoints & Read the objectives & \\
2 & Objectives and endpoints & Read the endpoints of each objective & \\
\midrule
\multicolumn{4}{@{}l}{\textit{Schedule of Activities}} \\
1 & Study evaluations & Browse the expected activities & \\
2 & Schedule of Activities & Evaluate the planned encounters & Count first, then name \\
3 & Schedule of Activities & Evaluate the planned activities & Reconcile with the study evaluations \\
4 & Schedule of Activities & Evaluate the activities at each encounter & \\
5 & Schedule of Activities & Review the footnotes & Look for key items such as timepoints \\
\bottomrule
\end{tabular}
\caption{Annotation steps. The general steps were performed for every protocol, before the steps of the task being annotated.}
\label{tab:annotation-runbook}
\end{table*}

\section{Tasks and Evaluation Metrics}
\label{app:tasks-and-metrics}
This appendix specifies the inputs, required outputs, and evaluation metrics for each task. Table~\ref{tab:task-statistics} summarizes the available task instances and reference annotations. Counts refer to task instances rather than additional unique studies: a source document can contribute to multiple tasks. Each agent is run once per task instance, and all reported scores are means over task instances unless stated otherwise.

\paragraph{Evaluation notation.}
Let $G$ and $\hat{G}$ denote the reference and predicted items for one task instance, with sizes $n$ and $\hat{n}$. For exact matching over multisets, let $c_G(k)$ and $c_{\hat{G}}(k)$ be the multiplicities of comparison key $k$. The number of matched items is
\begin{equation}
T=\sum_{k}\min\{c_G(k),c_{\hat{G}}(k)\}.
\label{eq:matched-items}
\end{equation}
For sets, each multiplicity is at most one. With nonempty reference and prediction, precision, recall, and F1 are
\begin{equation}
P=\frac{T}{\hat{n}},\qquad R=\frac{T}{n},\qquad F_1=\frac{2T}{n+\hat{n}}.
\label{eq:extraction-prf}
\end{equation}
An empty prediction against a nonempty reference receives F1 of zero. Task-specific conventions for empty references are stated below. Text comparisons first apply the task-specific normalization described below, which removes formatting differences such as case, whitespace, and markup. Matching remains exact after normalization; it does not use a language-model judge or accept arbitrary paraphrases. Unless stated otherwise, a metric $s$ is aggregated as the unweighted mean of its instance-level scores, $\bar{s}=N^{-1}\sum_{i=1}^{N}s_i$. Timeline Encounter recall instead pools counts across documents. Timeline Activity and Encounter matching use the semantic procedures specified below rather than the exact matching used for the text-extraction tasks.

\subsection{Indication Extraction}
\label{app:indication}
\paragraph{Input.}
The complete source PDF.

\paragraph{Output.}
A JSON object with a \texttt{descriptions} array listing the diseases or conditions investigated in the study. Each description uses the protocol's wording, including disease qualifiers. Distinct diseases in a multi-cohort study are separate entries, whereas qualifiers of one disease remain part of the same entry.

\paragraph{Metrics.}
The metric is the F1 score. Descriptions are normalized and matched one to one. Let $M_{\mathrm{ind}}$ be the resulting matching. The score is
\begin{equation}
F_{\mathrm{ind}}=\frac{2|M_{\mathrm{ind}}|}{|G|+|\hat{G}|}.
\end{equation}
Both sides empty receive a score of one; only one side empty receives F1 of zero.

\subsection{Eligibility Criteria Extraction}
\label{app:eligibility}
\paragraph{Input.}
The complete source PDF.

\paragraph{Output.}
A JSON object with a \texttt{criteria} array. Each criterion has a \texttt{category} (\texttt{Inclusion} or \texttt{Exclusion}) and its details in \texttt{text}. A document that states no eligibility criteria requires an empty array.

\paragraph{Metrics.}
The comparison key is $(\mathrm{category},\mathrm{text})$. Categories are compared case-insensitively after whitespace normalization. Before matching, text normalization removes HTML tags, decodes HTML entities, lowercases text, collapses whitespace, and strips trailing punctuation. Matching uses the multiset matching in Equation~\ref{eq:matched-items}, preserving repeated criteria. The F1 score is
\begin{equation}
F_{\mathrm{crit}}=
\begin{cases}
1, & n=0,\ \hat{n}=0,\\
2T/(n+\hat{n}), & \text{otherwise}.
\end{cases}
\label{eq:criterion-score}
\end{equation}
A nonempty prediction against an empty reference therefore scores zero. The same convention applies to per-category F1.

\paragraph{Category diagnostics.}
Inclusion and exclusion F1 are also computed separately. To distinguish extraction from classification errors, criteria are aligned on normalized text alone and category accuracy is computed over those matched pairs:
\begin{equation}
A_{\mathrm{cat}}=\frac{1}{|M_{\mathrm{text}}|}\sum_{(g,p)\in M_{\mathrm{text}}}\mathbf{1}[c_g=c_p].
\end{equation}
The model-level accuracy averages $A_{\mathrm{cat}}$ over the instances with at least one text-matched pair. Results are reported in Appendix~\ref{app:criteria-category}.

\subsection{Objectives and Endpoints Extraction}
\label{app:objectives}
\paragraph{Input.}
The complete source PDF.

\paragraph{Output.}
A JSON object with an \texttt{objectives} array. Each objective contains \texttt{objectiveText}, \texttt{objectiveLevel}, and a nested \texttt{endpoints} array. Each endpoint contains \texttt{endpointText} and \texttt{endpointLevel}. Levels are normalized to primary, secondary, or exploratory. Endpoints must remain associated with the objective they address; an objective without a stated endpoint receives an empty array.

\paragraph{Metrics.}
We evaluate both objective recovery and endpoint association. For an objective $o$, let $\ell_o$ and $t_o$ denote its normalized level and text. For an endpoint $e$, let $\ell_e$ and $t_e$ denote its level and text, and let $o(e)$ be its parent objective. The three comparison keys are
\begin{align}
k_{\mathrm{obj}}(o)&=(\ell_o,t_o),\\
k_{\mathrm{end}}(e)&=(\ell_e,t_e),\\
k_{\mathrm{att}}(e)&=(t_{o(e)},\ell_e,t_e).
\end{align}
After field normalization, we compute multiset F1 for each key type, giving objective F1 $F_{\mathrm{obj}}$, endpoint F1 $F_{\mathrm{end}}$, and objective-linked endpoint F1 $F_{\mathrm{att}}$, which are reported separately. Consequently, recovering an endpoint under the wrong objective fails the association metric even when its text is correct. Before comparison, text fields are normalized by stripping HTML tags, decoding entities, lowercasing text, collapsing whitespace, and removing trailing punctuation.

\subsection{Schedule of Activities Table Extraction}
\label{app:soa}
\paragraph{Track 1: page-level extraction.}
The input is a single PDF page containing an SoA table. The system transcribes the table into HTML, preserving activity labels, visit headers, cell contents, merged cells, and footnote markers. The Harbor output specification additionally requests surrounding page text and the footnote block as Markdown in the same output file. Scheduling marks such as ticks are represented as \texttt{X}, and superscript markers are retained using \texttt{<sup>}. This track contains 86 page samples from 40 NCT identifiers, with one reference table per sample. The scorer extracts 5,001 distinct activity-visit pairs across these samples, with a median of 52 and a range of 3-239 per sample.

\paragraph{Track 2: document-level extraction.}
The input is a complete protocol PDF. The system must locate all SoA pages, recover their tables, and declare their source PDF page numbers in the output, for example using \texttt{data-pages} attributes. This track contains 37 protocols and 70 reference tables; 10 protocols have multiple SoA pages, and the number of annotated SoA pages ranges from 1 to 13 per protocol. It includes only protocols for which the reference covers all SoA pages. The task therefore adds page localization to table transcription, while using the same table-content metrics as Track 1.

\paragraph{Activity--visit relations.}
Each scheduling mark defines a pair $(a,v)$ between an activity and a visit. The evaluator expands merged cells, combines hierarchical visit headers, and normalizes labels before forming the pair sets $X$ and $\hat{X}$. Normalization standardizes case, spacing, selected punctuation, and equivalent typographic forms; it does not perform semantic paraphrase matching. The Activity-Visit F1 is
\begin{equation}
F_X=\frac{2|X\cap\hat{X}|}{|X|+|\hat{X}|}.
\end{equation}
A scheduling mark assigned to the wrong visit counts as an incorrect relation. Duplicate identical pairs within an output are counted once.

\paragraph{Footnote attachment.}
SoA tables use superscript or subscript markers, such as ``Hematology$^{a}$'' or ``X$^{c}$'', to link an activity row, a single cell, or a visit column to a footnote below the table. Sub/sup F1 checks whether each marker is attached to the right place, not only whether it appears. A marker on an activity label is recorded as a pair $(a,m)$ of the activity $a$ and the marker $m$; a marker inside a cell is recorded as a triple $(a,v,m)$ that also includes the visit $v$; and a marker on a visit header is recorded as $(\ast,v,m)$, with a placeholder $\ast$ in place of the activity. With $A$ and $\hat{A}$ the resulting sets for the reference and the prediction,
\begin{equation}
F_A=\frac{2|A\cap\hat{A}|}{|A|+|\hat{A}|}.
\end{equation}
A marker that is recovered but attached to a different row, cell, or visit therefore does not count. Markers are recognized from \texttt{<sup>}/\texttt{<sub>} tags and Unicode superscript and subscript characters; a marker written as plain text next to a label is not counted. For both Activity-Visit F1 and Sub/sup F1, two empty sets score one, while a single empty side scores zero.

\paragraph{Fixed-gold-page aggregation.}
Document-level extraction uses every annotated SoA page, not only pages successfully located by a method. For protocol $i=1,\ldots,N$, let $L_i$ be its gold page set and $f_{ip}$ the Activity-Visit or Sub/sup F1 for page $p$, with $f_{ip}=0$ when that page is not recovered. The reported extraction metric is
\begin{equation}
\bar{F}=\frac{1}{N}\sum_{i=1}^{N}\frac{1}{|L_i|}\sum_{p\in L_i}f_{ip}.
\end{equation}
Thus, a protocol with no recovered gold page scores zero and remains in the denominator. 

\subsection{Structured Timeline Extraction}
\label{app:timeline}
\paragraph{Input.}
The complete source PDF.

\paragraph{Output.}
A JSON object listing the study's epochs, encounters, and activities.
\begin{itemize}[leftmargin=*,topsep=2pt,itemsep=1pt,parsep=0pt]\raggedright
\item \textbf{Epoch}: \texttt{type}, with epochs listed in study order.
\item \textbf{Encounter}: \texttt{label} and \texttt{description}, which identify the visit, and {\texttt{type} and \texttt{contactModes}.}
\item \textbf{Activity}: \texttt{label}.
\end{itemize}

\paragraph{Epoch accuracy.}
Let $g_1,\ldots,g_n$ and $p_1,\ldots,p_m$ be the ordered reference and predicted epoch types. Types are case-folded, trimmed, and whitespace-normalized. Positional accuracy is
\begin{equation}
A_{\mathrm{epoch}}=\frac{\sum_{j=1}^{\min(n,m)}\delta(g_j,p_j)}{\max(n,m)}.
\end{equation}
Here $\delta(g,p)=1$ only when both types are nonmissing and equal, and is zero otherwise. Missing types retain their positions, and there is no sequence realignment or synonym mapping, so an extra or omitted epoch shifts every later comparison. In the following example, one inserted epoch leaves only the first position matching, giving $A_{\mathrm{epoch}}=1/\max(3,4)=0.25$:
\begin{center}
\footnotesize\setlength{\tabcolsep}{3pt}
\begin{tabular}{@{}lcccc@{}}
\toprule
Position & 1 & 2 & 3 & 4\\
\midrule
Reference & Screening & Treatment & Follow-up & --\\
Prediction & Screening & Run-in & Treatment & Follow-up\\
Match & \ding{51} & \ding{55} & \ding{55} & \ding{55}\\
\bottomrule
\end{tabular}
\end{center}
Scores are averaged over all three documents, with failed extraction scored zero.

\paragraph{Activity recall.}
For study $i=1,\ldots,N$, let $G_i^{\mathrm{activity}}$ and $\hat{G}_i^{\mathrm{activity}}$ be the reference and predicted activity labels. We use Opus-5 with Claude Code as an agentic matcher: given the two label lists, it identifies the reference and predicted labels that denote the same clinical activity. From these pairs, we compute a maximum-cardinality one-to-one matching $M_i$, in which each reference and each predicted activity is used at most once. The reported recall is
\begin{equation}
R_{\mathrm{activity}}=\frac{1}{N}\sum_{i=1}^{N}\frac{|M_i|}{|G_i^{\mathrm{activity}}|}.
\end{equation}
A predicted activity with a blank label cannot be matched, and a failed extraction scores zero.

\paragraph{Encounter recall.}
For study $i$, let $G_i^{\mathrm{enc}}$ and $\hat{G}_i^{\mathrm{enc}}$ be the reference and predicted encounters. The same agentic matcher first proposes pairs referring to the same visit using encounter labels, descriptions, and explicit timing context. A deterministic maximum-cardinality one-to-one matching is then computed before any attribute is checked. Let $M_i^{\mathrm{enc}}$ denote this matching, whose elements are pairs $(g,p)$ with $g\in G_i^{\mathrm{enc}}$ and $p\in\hat{G}_i^{\mathrm{enc}}$. {A matched pair counts as a recovered encounter only when the two encounters also agree on two additional attributes, the encounter type and the contact modes. Let $C(g,p)=1$ if $g$ and $p$ agree on both attributes, and $0$ otherwise. We report}
\begin{equation}
R_{\mathrm{enc}}=\frac{\sum_{i=1}^{N}\sum_{(g,p)\in M_i^{\mathrm{enc}}}C(g,p)}{\sum_{i=1}^{N}|G_i^{\mathrm{enc}}|}.
\end{equation}
Unmatched gold visits and failed extractions contribute zero correct matches. Type uses case- and whitespace-normalized equality; contact modes use normalized unordered sets. This metric measures recovered visits with correct attributes.

\section{Experimental Details}
\label{app:experimental}

\subsection{Coding Agents}
\label{app:agents}
Table~\ref{tab:agent-config} lists the evaluated models and harnesses. In addition to the nine models reported in the main text, we also evaluate GPT-4o with Codex, whose results are reported only in the appendix. Each task instance runs in an isolated container that contains the source PDF, the task instruction, and an output directory; reference annotations are never mounted. Qwen3.8-27B is served locally with SGLang \citep{zheng2024sglang}, with a 131,072-token context window and a maximum output of 16,384 tokens.

\begin{table}[t]
\centering
\small
\setlength{\tabcolsep}{3pt}
\resizebox{\columnwidth}{!}{%
\begin{tabular}{lll}
\toprule
Model & Model identifier & Harness \\
\midrule
GPT-4o & \texttt{gpt-4o} & Codex \\
GPT-5.6-Sol & \texttt{gpt-5.6-sol} & Codex \\
GPT-6-Astra & \texttt{gpt-6-astra} & Codex \\
Opus 4.6 & \texttt{claude-opus-4-6} & Claude Code \\
Opus 4.8 & \texttt{claude-opus-4-8} & Claude Code \\
Opus 5 & \texttt{claude-opus-5} & Claude Code \\
Sonnet 5 & \texttt{claude-sonnet-5} & Claude Code \\
Kimi-K3 & \texttt{Kimi-K3} & Kimi Code \\
DeepSeek-V4-Flash & \texttt{deepseek-v4-flash-0731} & DeepSeek Harness \\
Qwen3.8-27B & \texttt{Qwen/Qwen3.8-27B} & pi \\
\bottomrule
\end{tabular}%
}
\caption{Evaluated models and coding-agent harnesses. GPT-4o is reported only in the appendix.}
\label{tab:agent-config}
\end{table}

\subsection{RAG Baseline}
\label{app:rag}
The RAG baseline follows the protocol information extraction system of \citet{babaeipour2026ai}. The generator is the same model that is evaluated as a coding agent. For retrieval, we use \texttt{text-embedding-3-large}.

\subsection{OCR and Document-Understanding Baselines}
\label{app:ocr}
We evaluate eight OCR and document-understanding baselines: Mistral OCR \citep{mistral2025ocr}, LightOnOCR-2-1B \citep{taghadouini2026lightonocr}, LandingAI \citep{landingai2026}, olmOCR-2-7B \citep{poznanski2025olmocr}, PaddleOCR \citep{cui2025paddleocr}, PaddleOCR-VL \citep{cui2025paddleocrvlboostingmultilingualdocument}, and the 2B and 4B variants of Qwen3-VL \citep{bai2025qwen3}.

The eight OCR and document-understanding baselines process the same 86 single-page SoA inputs as the page-level task. Their raw table outputs are scored with the same evaluator. Because Sub/sup F1 requires footnote markers to be encoded as superscripts or subscripts, baselines that emit plain text, such as Mistral OCR, score low on this metric even when their Activity-Visit F1 is high.

\subsection{Task Instructions}
\label{app:instructions}
Each task instruction specifies the input file, the output path and format, and the conventions that define a correct output, such as verbatim transcription, one item per numbered criterion, or nesting endpoints under the objective they belong to. For page-level SoA extraction, a separate specification file defines the HTML format; for timeline extraction, a JSON schema following USDM v4.0 defines the output. The instructions do not prescribe how the agent should process the document.

\section{Additional Analyses}
\label{app:additional-results}

\subsection{Comparison with OCR and Document-Understanding Models}
\label{app:ocr-comparison}
\begin{figure}[t]
    \centering
    \includegraphics[width=\columnwidth]{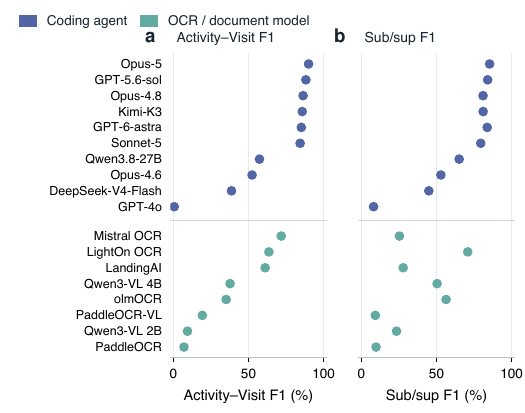}
    \caption{{Page-level SoA extraction, comparing coding agents with OCR and document-understanding baselines using \textbf{(a)} Activity-Visit F1 and \textbf{(b)} Sub/sup F1.} Activity-Visit F1 measures the recovery of scheduling associations between activities and visits, usually represented by X marks in SoA table cells. Sub/sup F1 measures the recovery of subscript and superscript footnote markers.}
    \label{fig:soa-results}
    \vspace{-1em}
\end{figure}

We compare coding agents with the eight OCR and document-understanding baselines of Appendix~\ref{app:ocr} on page-level SoA extraction {(Figure~\ref{fig:soa-results})}.

\textbf{Coding agents outperform specialist OCR and document-understanding models on SoA table extraction.}
On page-level SoA extraction (Figure~\ref{fig:soa-results}a), the six strongest agents reach 84.3-90.0\% Activity-Visit F1, exceeding the best specialist baseline, Mistral OCR, at 71.7\%. The advantage extends to footnote markers (Figure~\ref{fig:soa-results}{b}), where the best agent reaches 85.3\% Sub/sup F1, compared with 70.8\% for the strongest baseline, LightOnOCR-2-1B. Coding agents also handle the additional localization required at the document level: for five of the six strongest agents, document-level Activity-Visit F1 {(Table~\ref{tab:full-soa-doc-results})} remains within 5 points of the page-level score. General-purpose coding agents can therefore both find SoA tables in complete protocols and transcribe them more accurately than models designed specifically for document parsing.

\subsection{Comparison with a RAG Pipeline}
\label{app:rag-comparison}
Each model is deployed both as a coding agent and as the generator of the RAG pipeline (Appendix~\ref{app:rag}).

\begin{figure*}[t]
    \centering
    \includegraphics[width=\textwidth]{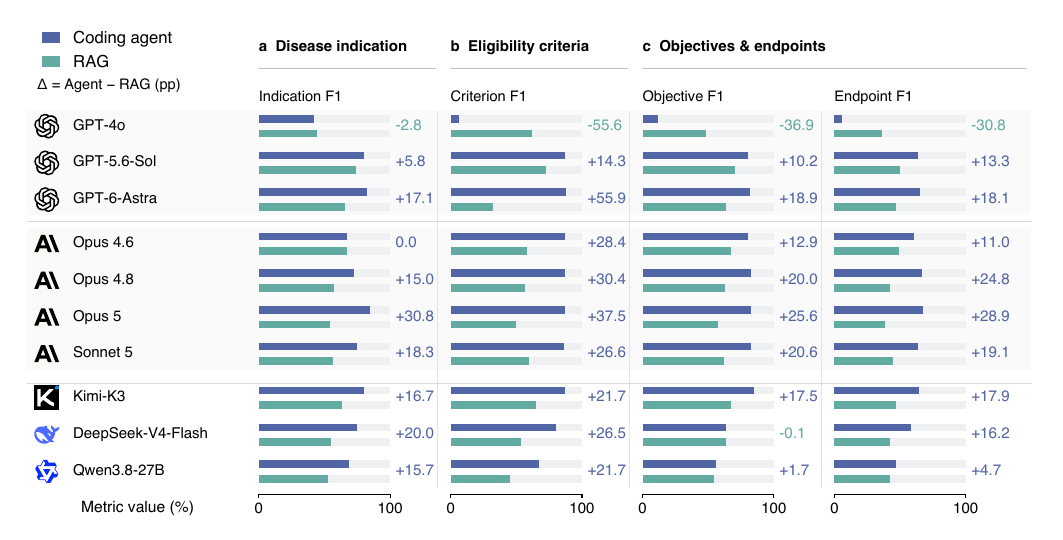}
    \caption{Comparison of coding agents and the RAG baseline across \textbf{(a)} disease indication, \textbf{(b)} eligibility criteria, and \textbf{(c)} objectives and endpoints. Bars show F1 scores (\%), and annotations report the difference between the coding agent and RAG for each model in percentage points ($\Delta = \mathrm{Agent} - \mathrm{RAG}$). Positive values favor coding agents.}
    \label{fig:agent-rag-comparison}
    \vspace{-1em}
\end{figure*}

\textbf{Coding agents substantially outperform a RAG pipeline on the protocol digitization tasks in \ourbench{}.}
Coding agents outperform RAG with the same model on disease indication (Figure~\ref{fig:agent-rag-comparison}a), eligibility criteria (Figure~\ref{fig:agent-rag-comparison}b), objectives and endpoints (Figure~\ref{fig:agent-rag-comparison}c), and SoA table extraction {(Table~\ref{tab:full-soa-doc-results})}. For nine of the ten models, agents improve eligibility criteria F1 by 14.3-55.9\% and endpoint F1 by 4.7-28.9\%; among these nine, indication F1 also improves for all but Opus 4.6, which ties. The gap is largest for SoA table extraction, where the six strongest agents (GPT-5.6-Sol, GPT-6-Astra, Opus 4.8, Opus 5, Sonnet 5, and Kimi-K3) reach 74.7-86.8\% Activity-Visit F1, whereas RAG with the same models reaches only 19.8-51.5\%. One possible explanation is that a predefined pipeline determines in advance how documents are parsed and presented to the model, which may limit its generalization and flexibility across protocols and tasks, whereas coding agents can choose and adjust their document-processing strategy for each protocol and task.

\subsection{Eligibility Criteria by Category}
\label{app:criteria-category}
Table~\ref{tab:criteria-category} reports inclusion and exclusion F1. Category accuracy is at least 99.5\% for every agent and RAG model, so criteria that are recovered are almost never filed under the wrong category, and the loss in criterion F1 comes from extraction rather than classification.

\begin{table}[t]
\centering
\small
\setlength{\tabcolsep}{4pt}
\begin{tabular}{lrrrr}
\toprule
 & \multicolumn{2}{c}{Inclusion F1} & \multicolumn{2}{c}{Exclusion F1} \\
\cmidrule(lr){2-3}\cmidrule(lr){4-5}
Model & Agent & RAG & Agent & RAG \\
\midrule
GPT-4o & 9.4 & 58.6 & 4.7 & 64.2 \\
GPT-5.6-Sol & 81.8 & 67.6 & 90.9 & 77.1 \\
GPT-6-Astra & 81.7 & 29.8 & 91.0 & 34.5 \\
Opus 4.6 & 81.9 & 53.8 & 89.5 & 62.4 \\
Opus 4.8 & 82.1 & 52.1 & 90.0 & 60.7 \\
Opus 5 & 82.2 & 46.0 & 90.5 & 52.9 \\
Sonnet 5 & 81.2 & 54.0 & 89.1 & 64.4 \\
Kimi-K3 & 81.7 & 59.1 & 90.0 & 69.8 \\
DeepSeek-V4-Flash & 74.2 & 47.5 & 85.5 & 58.3 \\
Qwen3.8-27B & 63.3 & 41.1 & 69.2 & 48.2 \\
\bottomrule
\end{tabular}
\caption{Eligibility criteria F1 (\%) by category for coding agents and RAG.}
\label{tab:criteria-category}
\end{table}

\subsection{Task-Level Results by PDF Text Accessibility}
\label{app:pdf-text-accessibility}
Tables~\ref{tab:pdf-modality-indication}--\ref{tab:pdf-modality-timeline} report coding-agent results separately for each task, with each metric grouped into all PDFs, text-sufficient PDFs, and vision-required PDFs.

We divide the PDFs of each task into these two groups by whether the content needed for that task can be read from the PDF's native text layer. Only the pages that hold this content are examined, such as the eligibility section for criteria extraction or the SoA page for table extraction. If native text extraction with Poppler and PyMuPDF recovers the content, the PDF is text-sufficient. If the content is scanned, garbled by encoding errors, or otherwise missing from the text layer, so that it can be read only from rendered pages or by OCR, the PDF is vision-required. Scans, encoding failures, mixed documents, and redactions were reviewed manually. Blank pages, scanned signatures, and other pages outside the task content do not affect the division, and model outputs and scores are not used. As a result, the same protocol can be text-sufficient for one task and vision-required for another. Vision-required describes the PDF, not the model: an agent can still read such content with OCR tools even if its language model does not accept images.

\paragraph{Task coverage.}
The all/text/vision sample counts are 40/33/7 for indication, 134/94/40 for eligibility criteria, 89/61/28 for objectives and endpoints, 86/61/25 for page-level SoA extraction, 37/29/8 for document-level SoA extraction, and 3/3/0 for structured timeline extraction. In SoA Track~1 each sample is a single PDF page, whereas in SoA Track~2 each sample is a whole protocol.

\paragraph{Scores and aggregation.}
Indication and criteria tables report F1. Objectives and endpoints include objective F1, endpoint F1, and objective-linked endpoint F1. SoA tables report Activity-Visit F1 and Sub/sup F1. Document-level extraction uses the fixed set of gold SoA pages: a missing page contributes zero, and scores are averaged first within each protocol and then across protocols. Timeline reports Epoch type positional accuracy, semantic Activity recall, and Encounter recall. For Encounter recall, a reference visit counts as recovered only when a matched prediction also has the correct type and contact modes. Encounter recall pools correct visits and gold visits across documents; other reported metrics are macro-averaged over task instances. Failed or missing outputs remain in their prescribed denominators. A dash denotes an empty cohort, not zero performance. Cohort sizes in table headers count PDFs, including for Encounter recall, whose metric denominator counts gold encounters.

\begin{table*}[t]
\centering
\small
\caption{Study indication. Agent results (\%); higher is better. Each metric reports all PDFs, text-sufficient PDFs, and vision-required PDFs.}
\label{tab:pdf-modality-indication}
\setlength{\tabcolsep}{4pt}
\resizebox{\ifdim\width>\textwidth\textwidth\else\width\fi}{!}{%
\begin{tabular}{lrrr}
\toprule
Agent & \multicolumn{3}{c}{Indication F1} \\
\cmidrule(lr){2-4}
 & \shortstack{All\\$n=40$} & \shortstack{Text\\$n=33$} & \shortstack{Vision\\$n=7$} \\
\midrule
GPT-4o & 41.7 & 44.4 & 28.6 \\
GPT-5.6-Sol & 80.0 & 84.8 & 57.1 \\
GPT-6-Astra & 82.5 & 87.9 & 57.1 \\
Opus 4.6 & 67.5 & 69.7 & 57.1 \\
Opus 4.8 & 72.5 & 72.7 & 71.4 \\
Opus 5 & 85.0 & 87.9 & 71.4 \\
Sonnet 5 & 75.0 & 78.8 & 57.1 \\
Kimi-K3 & 80.0 & 81.8 & 71.4 \\
DeepSeek-V4-Flash & 75.0 & 72.7 & 85.7 \\
Qwen3.8-27B & 68.3 & 71.7 & 52.4 \\
\bottomrule
\end{tabular}%
}
\par\vspace{2pt}
\begin{minipage}{\textwidth}\footnotesize
\textemdash{} denotes an empty cohort, not zero performance. All = all PDFs; Text = text-sufficient; Vision = vision-required. The labels describe task-relevant PDF text accessibility. Failed/missing outputs remain in the source metric denominator.
\end{minipage}
\end{table*}

\begin{table*}[t]
\centering
\small
\caption{Eligibility criteria. Agent results (\%); higher is better. Each metric reports all PDFs, text-sufficient PDFs, and vision-required PDFs.}
\label{tab:pdf-modality-criteria}
\setlength{\tabcolsep}{4pt}
\resizebox{\ifdim\width>\textwidth\textwidth\else\width\fi}{!}{%
\begin{tabular}{lrrr}
\toprule
Agent & \multicolumn{3}{c}{Criterion F1} \\
\cmidrule(lr){2-4}
 & \shortstack{All\\$n=134$} & \shortstack{Text\\$n=94$} & \shortstack{Vision\\$n=40$} \\
\midrule
GPT-4o & 6.2 & 7.7 & 2.7 \\
GPT-5.6-Sol & 87.1 & 88.2 & 84.4 \\
GPT-6-Astra & 87.8 & 89.1 & 84.6 \\
Opus 4.6 & 86.8 & 88.5 & 82.9 \\
Opus 4.8 & 87.2 & 88.9 & 83.0 \\
Opus 5 & 87.0 & 88.4 & 83.7 \\
Sonnet 5 & 86.4 & 87.6 & 83.6 \\
Kimi-K3 & 86.9 & 88.7 & 82.7 \\
DeepSeek-V4-Flash & 80.2 & 85.0 & 68.9 \\
Qwen3.8-27B & 67.1 & 77.1 & 43.8 \\
\bottomrule
\end{tabular}%
}
\par\vspace{2pt}
\begin{minipage}{\textwidth}\footnotesize
\textemdash{} denotes an empty cohort, not zero performance. All = all PDFs; Text = text-sufficient; Vision = vision-required. The labels describe task-relevant PDF text accessibility. Failed/missing outputs remain in the source metric denominator.
\end{minipage}
\end{table*}

\begin{table*}[t]
\centering
\small
\caption{Objectives and endpoints. Agent results (\%); higher is better. Each metric reports all PDFs, text-sufficient PDFs, and vision-required PDFs.}
\label{tab:pdf-modality-objectives}
\setlength{\tabcolsep}{4pt}
\resizebox{\ifdim\width>\textwidth\textwidth\else\width\fi}{!}{%
\begin{tabular}{lrrrrrrrrr}
\toprule
Agent & \multicolumn{3}{c}{Objective F1} & \multicolumn{3}{c}{Endpoint F1} & \multicolumn{3}{c}{\shortstack{Objective-linked\\endpoint F1}} \\
\cmidrule(lr){2-4} \cmidrule(lr){5-7} \cmidrule(lr){8-10}
 & \shortstack{All\\$n=89$} & \shortstack{Text\\$n=61$} & \shortstack{Vision\\$n=28$} & \shortstack{All\\$n=89$} & \shortstack{Text\\$n=61$} & \shortstack{Vision\\$n=28$} & \shortstack{All\\$n=89$} & \shortstack{Text\\$n=61$} & \shortstack{Vision\\$n=28$} \\
\midrule
GPT-4o & 11.4 & 16.6 & 0.0 & 5.6 & 8.0 & 0.4 & 2.7 & 3.9 & 0.0 \\
GPT-5.6-Sol & 80.4 & 82.1 & 76.6 & 63.6 & 64.4 & 61.7 & 54.7 & 56.6 & 50.6 \\
GPT-6-Astra & 82.2 & 84.5 & 77.0 & 65.0 & 67.7 & 59.2 & 58.1 & 61.4 & 50.8 \\
Opus 4.6 & 80.6 & 82.7 & 75.9 & 60.6 & 61.7 & 58.2 & 49.6 & 50.1 & 48.7 \\
Opus 4.8 & 82.7 & 86.2 & 75.1 & 66.9 & 67.1 & 66.4 & 58.5 & 62.2 & 50.5 \\
Opus 5 & 82.9 & 88.0 & 71.9 & 67.4 & 69.1 & 63.6 & 58.7 & 63.7 & 47.8 \\
Sonnet 5 & 82.8 & 84.2 & 79.7 & 63.9 & 64.9 & 61.6 & 54.2 & 56.9 & 48.4 \\
Kimi-K3 & 84.9 & 85.0 & 84.7 & 64.6 & 64.0 & 65.8 & 56.4 & 55.7 & 58.0 \\
DeepSeek-V4-Flash & 63.3 & 64.6 & 60.4 & 58.5 & 59.9 & 55.5 & 36.8 & 39.3 & 31.4 \\
Qwen3.8-27B & 56.0 & 67.1 & 31.8 & 46.7 & 54.8 & 29.1 & 32.1 & 39.5 & 16.0 \\
\bottomrule
\end{tabular}%
}
\par\vspace{2pt}
\begin{minipage}{\textwidth}\footnotesize
\textemdash{} denotes an empty cohort, not zero performance. All = all PDFs; Text = text-sufficient; Vision = vision-required. The labels describe task-relevant PDF text accessibility. Failed/missing outputs remain in the source metric denominator.
\end{minipage}
\end{table*}

\begin{table*}[t]
\centering
\small
\caption{SoA page extraction. Agent results (\%); higher is better. Each metric reports all PDFs, text-sufficient PDFs, and vision-required PDFs.}
\label{tab:pdf-modality-soa-track1}
\setlength{\tabcolsep}{4pt}
\resizebox{\ifdim\width>\textwidth\textwidth\else\width\fi}{!}{%
\begin{tabular}{lrrrrrr}
\toprule
Agent & \multicolumn{3}{c}{Activity--Visit F1} & \multicolumn{3}{c}{Sub/sup F1} \\
\cmidrule(lr){2-4} \cmidrule(lr){5-7}
 & \shortstack{All\\$n=86$} & \shortstack{Text\\$n=61$} & \shortstack{Vision\\$n=25$} & \shortstack{All\\$n=86$} & \shortstack{Text\\$n=61$} & \shortstack{Vision\\$n=25$} \\
\midrule
GPT-4o & 0.5 & 0.0 & 1.9 & 8.1 & 10.5 & 2.4 \\
GPT-5.6-Sol & 88.2 & 86.9 & 91.3 & 84.0 & 80.7 & 92.0 \\
GPT-6-Astra & 85.1 & 83.2 & 89.8 & 83.7 & 80.3 & 92.2 \\
Opus 4.6 & 52.4 & 47.5 & 64.3 & 52.9 & 47.6 & 65.9 \\
Opus 4.8 & 86.3 & 84.7 & 90.4 & 81.0 & 78.9 & 86.0 \\
Opus 5 & 90.0 & 88.7 & 93.0 & 85.3 & 83.1 & 90.8 \\
Sonnet 5 & 84.3 & 82.6 & 88.6 & 79.5 & 77.4 & 84.5 \\
Kimi-K3 & 85.7 & 82.9 & 92.8 & 81.0 & 78.4 & 87.5 \\
DeepSeek-V4-Flash & 38.7 & 50.4 & 10.1 & 44.9 & 58.9 & 10.6 \\
Qwen3.8-27B & 57.3 & 54.0 & 65.2 & 65.1 & 61.0 & 75.1 \\
\bottomrule
\end{tabular}%
}
\par\vspace{2pt}
\begin{minipage}{\textwidth}\footnotesize
\textemdash{} denotes an empty cohort, not zero performance. All = all PDFs; Text = text-sufficient; Vision = vision-required. The labels describe task-relevant PDF text accessibility. Failed/missing outputs remain in the source metric denominator.
\end{minipage}
\end{table*}

\begin{table*}[t]
\centering
\small
\caption{SoA document extraction. Agent results (\%); higher is better. Each metric reports all PDFs, text-sufficient PDFs, and vision-required PDFs.}
\label{tab:pdf-modality-soa-track2}
\setlength{\tabcolsep}{4pt}
\resizebox{\ifdim\width>\textwidth\textwidth\else\width\fi}{!}{%
\begin{tabular}{lrrrrrr}
\toprule
Agent & \multicolumn{3}{c}{Activity--Visit F1} & \multicolumn{3}{c}{Sub/sup F1} \\
\cmidrule(lr){2-4} \cmidrule(lr){5-7}
 & \shortstack{All\\$n=37$} & \shortstack{Text\\$n=29$} & \shortstack{Vision\\$n=8$} & \shortstack{All\\$n=37$} & \shortstack{Text\\$n=29$} & \shortstack{Vision\\$n=8$} \\
\midrule
GPT-4o & 2.1 & 2.7 & 0.0 & 3.1 & 2.3 & 5.9 \\
GPT-5.6-Sol & 83.7 & 84.6 & 80.4 & 81.0 & 79.5 & 86.3 \\
GPT-6-Astra & 86.8 & 88.0 & 82.4 & 84.5 & 83.7 & 87.1 \\
Opus 4.6 & 54.7 & 56.1 & 49.6 & 53.4 & 54.4 & 49.6 \\
Opus 4.8 & 81.4 & 82.0 & 79.2 & 78.0 & 78.1 & 77.7 \\
Opus 5 & 86.0 & 86.4 & 84.9 & 83.9 & 83.1 & 86.7 \\
Sonnet 5 & 74.7 & 77.0 & 66.4 & 71.8 & 72.0 & 71.3 \\
Kimi-K3 & 83.9 & 84.9 & 80.3 & 82.0 & 82.1 & 81.8 \\
DeepSeek-V4-Flash & 47.5 & 56.3 & 15.5 & 54.4 & 61.3 & 29.5 \\
Qwen3.8-27B & 32.5 & 34.0 & 26.9 & 42.5 & 44.5 & 35.3 \\
\bottomrule
\end{tabular}%
}
\par\vspace{2pt}
\begin{minipage}{\textwidth}\footnotesize
\textemdash{} denotes an empty cohort, not zero performance. All = all PDFs; Text = text-sufficient; Vision = vision-required. The labels describe task-relevant PDF text accessibility. Failed/missing outputs remain in the source metric denominator.
\end{minipage}
\end{table*}

\begin{table*}[t]
\centering
\small
\caption{Structured timeline. Agent results (\%); higher is better. Each metric reports all PDFs, text-sufficient PDFs, and vision-required PDFs.}
\label{tab:pdf-modality-timeline}
\setlength{\tabcolsep}{4pt}
\resizebox{\ifdim\width>\textwidth\textwidth\else\width\fi}{!}{%
\begin{tabular}{lrrrrrrrrr}
\toprule
Agent & \multicolumn{3}{c}{Epoch accuracy} & \multicolumn{3}{c}{Activity recall} & \multicolumn{3}{c}{Encounter recall} \\
\cmidrule(lr){2-4} \cmidrule(lr){5-7} \cmidrule(lr){8-10}
 & \shortstack{All\\$n=3$} & \shortstack{Text\\$n=3$} & \shortstack{Vision\\$n=0$} & \shortstack{All\\$n=3$} & \shortstack{Text\\$n=3$} & \shortstack{Vision\\$n=0$} & \shortstack{All\\$n=3$} & \shortstack{Text\\$n=3$} & \shortstack{Vision\\$n=0$} \\
\midrule
GPT-4o & 0.0 & 0.0 & \textemdash{} & 0.0 & 0.0 & \textemdash{} & 0.0 & 0.0 & \textemdash{} \\
GPT-5.6-Sol & 61.1 & 61.1 & \textemdash{} & 88.1 & 88.1 & \textemdash{} & 23.2 & 23.2 & \textemdash{} \\
GPT-6-Astra & 73.3 & 73.3 & \textemdash{} & 91.1 & 91.1 & \textemdash{} & 20.3 & 20.3 & \textemdash{} \\
Opus 4.6 & 61.1 & 61.1 & \textemdash{} & 85.3 & 85.3 & \textemdash{} & 0.0 & 0.0 & \textemdash{} \\
Opus 4.8 & 64.4 & 64.4 & \textemdash{} & 82.5 & 82.5 & \textemdash{} & 0.0 & 0.0 & \textemdash{} \\
Opus 5 & 38.6 & 38.6 & \textemdash{} & 87.2 & 87.2 & \textemdash{} & 0.0 & 0.0 & \textemdash{} \\
Sonnet 5 & 50.6 & 50.6 & \textemdash{} & 59.2 & 59.2 & \textemdash{} & 8.7 & 8.7 & \textemdash{} \\
Kimi-K3 & 61.1 & 61.1 & \textemdash{} & 50.2 & 50.2 & \textemdash{} & 21.7 & 21.7 & \textemdash{} \\
DeepSeek-V4-Flash & 48.5 & 48.5 & \textemdash{} & 72.2 & 72.2 & \textemdash{} & 2.9 & 2.9 & \textemdash{} \\
Qwen3.8-27B & 22.9 & 22.9 & \textemdash{} & 82.5 & 82.5 & \textemdash{} & 5.8 & 5.8 & \textemdash{} \\
\bottomrule
\end{tabular}%
}
\par\vspace{2pt}
\begin{minipage}{\textwidth}\footnotesize
\textemdash{} denotes an empty cohort, not zero performance. All = all PDFs; Text = text-sufficient; Vision = vision-required. The labels describe task-relevant PDF text accessibility. Failed/missing outputs remain in the source metric denominator. Encounter recall counts a matched visit as recovered only when its type and contact modes are also correct. Encounter recall pools gold encounters, while Epoch accuracy and Activity recall are averaged over documents.
\end{minipage}
\end{table*}

\subsection{Full Numerical Results}
\label{app:full-results}
{Tables~\ref{tab:full-text-results} and~\ref{tab:full-soa-page-results} report the exact scores plotted in Figures~\ref{fig:agent-rag-comparison} and~\ref{fig:soa-results}, respectively. Tables~\ref{tab:full-soa-doc-results} and~\ref{tab:full-timeline-results} report the document-level SoA and structured timeline results of coding agents and RAG.}

\begin{table*}[t]
\centering
\small
\setlength{\tabcolsep}{4pt}
\begin{tabular}{lrrrrrrrrrr}
\toprule
 & \multicolumn{2}{c}{Indication F1} & \multicolumn{2}{c}{Criterion F1} & \multicolumn{2}{c}{Objective F1} & \multicolumn{2}{c}{Endpoint F1} & \multicolumn{2}{c}{Linked endpoint F1} \\
\cmidrule(lr){2-3}\cmidrule(lr){4-5}\cmidrule(lr){6-7}\cmidrule(lr){8-9}\cmidrule(lr){10-11}
Model & Agent & RAG & Agent & RAG & Agent & RAG & Agent & RAG & Agent & RAG \\
\midrule
GPT-4o & 41.7 & 44.5 & 6.2 & 61.9 & 11.4 & 48.3 & 5.6 & 36.4 & 2.7 & 20.0 \\
GPT-5.6-Sol & 80.0 & 74.2 & 87.1 & 72.8 & 80.4 & 70.2 & 63.6 & 50.3 & 54.7 & 45.0 \\
GPT-6-Astra & 82.5 & 65.4 & 87.8 & 31.9 & 82.2 & 63.3 & 65.0 & 46.9 & 58.1 & 41.7 \\
Opus 4.6 & 67.5 & 67.5 & 86.8 & 58.4 & 80.6 & 67.7 & 60.6 & 49.5 & 49.6 & 37.3 \\
Opus 4.8 & 72.5 & 57.5 & 87.2 & 56.7 & 82.7 & 62.8 & 66.9 & 42.1 & 58.5 & 33.4 \\
Opus 5 & 85.0 & 54.2 & 87.0 & 49.5 & 82.9 & 57.3 & 67.4 & 38.5 & 58.7 & 29.9 \\
Sonnet 5 & 75.0 & 56.7 & 86.4 & 59.8 & 82.8 & 62.2 & 63.9 & 44.7 & 54.2 & 31.3 \\
Kimi-K3 & 80.0 & 63.3 & 86.9 & 65.3 & 84.9 & 67.4 & 64.6 & 46.7 & 56.4 & 37.2 \\
DeepSeek-V4-Flash & 75.0 & 55.0 & 80.2 & 53.7 & 63.3 & 63.4 & 58.5 & 42.3 & 36.8 & 35.2 \\
Qwen3.8-27B & 68.3 & 52.7 & 67.1 & 45.5 & 56.0 & 54.3 & 46.7 & 42.0 & 32.1 & 27.5 \\
\bottomrule
\end{tabular}
\caption{Coding-agent and RAG results (\%) on disease indication, eligibility criteria, and objectives and endpoints, corresponding to Figure~\ref{fig:agent-rag-comparison}. Linked endpoint F1 is the objective-linked endpoint F1 defined in Appendix~\ref{app:objectives}; it is reported here but not plotted.}
\label{tab:full-text-results}
\end{table*}

\begin{table}[t]
\centering
\small
\setlength{\tabcolsep}{4pt}
\begin{tabular}{lrr}
\toprule
Method & Activity-Visit F1 & Sub/sup F1 \\
\midrule
\multicolumn{3}{l}{\textit{Coding agents}} \\
GPT-4o & 0.5 & 8.1 \\
GPT-5.6-Sol & 88.2 & 84.0 \\
GPT-6-Astra & 85.1 & 83.7 \\
Opus 4.6 & 52.4 & 52.9 \\
Opus 4.8 & 86.3 & 81.0 \\
Opus 5 & 90.0 & 85.3 \\
Sonnet 5 & 84.3 & 79.5 \\
Kimi-K3 & 85.7 & 81.0 \\
DeepSeek-V4-Flash & 38.7 & 44.9 \\
Qwen3.8-27B & 57.3 & 65.1 \\
\midrule
\multicolumn{3}{l}{\textit{OCR and document-understanding baselines}} \\
Mistral OCR & 71.7 & 25.3 \\
LightOnOCR-2-1B & 63.6 & 70.8 \\
LandingAI & 61.1 & 27.8 \\
olmOCR-2-7B & 35.1 & 56.4 \\
PaddleOCR & 7.1 & 9.8 \\
PaddleOCR-VL & 19.3 & 9.3 \\
Qwen3-VL-2B & 9.4 & 23.4 \\
Qwen3-VL-4B & 37.7 & 50.4 \\
\bottomrule
\end{tabular}
\caption{Page-level SoA extraction results (\%) on the 86 single-page inputs, corresponding to Figure~\ref{fig:soa-results}{a-b}.}
\label{tab:full-soa-page-results}
\end{table}

\begin{table}[t]
\centering
\small
\setlength{\tabcolsep}{4pt}
\begin{tabular}{lrr}
\toprule
 & \multicolumn{2}{c}{Activity-Visit F1} \\
\cmidrule(lr){2-3}
Model & Agent & RAG \\
\midrule
GPT-4o & 2.1 & 4.7 \\
GPT-5.6-Sol & 83.7 & 44.0 \\
GPT-6-Astra & 86.8 & 51.5 \\
Opus 4.6 & 54.7 & 33.5 \\
Opus 4.8 & 81.4 & 19.8 \\
Opus 5 & 86.0 & 31.7 \\
Sonnet 5 & 74.7 & 25.0 \\
Kimi-K3 & 83.9 & 27.1 \\
DeepSeek-V4-Flash & 47.5 & 15.8 \\
Qwen3.8-27B & 32.5 & 16.7 \\
\bottomrule
\end{tabular}
\caption{Document-level SoA extraction results (\%) on the 37 protocols. Activity-Visit F1 averages all annotated SoA pages within each protocol, with missing pages scored zero, and then across protocols.}
\label{tab:full-soa-doc-results}
\end{table}

\begin{table*}[t]
\centering
\small
\setlength{\tabcolsep}{4pt}
\begin{tabular}{lrrrrrr}
\toprule
 & \multicolumn{2}{c}{Epoch accuracy} & \multicolumn{2}{c}{Activity recall} & \multicolumn{2}{c}{Encounter recall} \\
\cmidrule(lr){2-3}\cmidrule(lr){4-5}\cmidrule(lr){6-7}
Model & Agent & RAG & Agent & RAG & Agent & RAG \\
\midrule
GPT-4o & 0.0 & 6.7 & 0.0 & 3.7 & 0.0 & 0.0 \\
GPT-5.6-Sol & 61.1 & 61.1 & 88.1 & 54.9 & 23.2 & 40.6 \\
GPT-6-Astra & 73.3 & 67.8 & 91.1 & 46.0 & 20.3 & 0.0 \\
Opus 4.6 & 61.1 & 49.4 & 85.3 & 83.3 & 0.0 & 0.0 \\
Opus 4.8 & 64.4 & 61.1 & 82.5 & 80.3 & 0.0 & 0.0 \\
Opus 5 & 38.6 & 61.1 & 87.2 & 87.2 & 0.0 & 0.0 \\
Sonnet 5 & 50.6 & 45.9 & 59.2 & 25.2 & 8.7 & 0.0 \\
Kimi-K3 & 61.1 & 57.4 & 50.2 & 81.6 & 21.7 & 27.5 \\
DeepSeek-V4-Flash & 48.5 & 61.1 & 72.2 & 82.5 & 2.9 & 13.0 \\
Qwen3.8-27B & 22.9 & 30.0 & 82.5 & 53.1 & 5.8 & 30.4 \\
\bottomrule
\end{tabular}
\caption{Structured timeline extraction results (\%) on the three official example protocols. Encounter recall counts a matched visit as recovered only when its type and contact modes are also correct (Appendix~\ref{app:timeline}).}
\label{tab:full-timeline-results}
\end{table*}

\section{Failure Analysis Details}
\label{app:failure}

\subsection{Failure Taxonomy}
\label{app:failure-taxonomy}
We define a failing run as a coding-agent run with at least one detected defect, regardless of its score. Defects are detected automatically by comparing each prediction with the reference annotations using Opus-5. The five main types of failure are:
\begin{itemize}[leftmargin=*,topsep=2pt,itemsep=1pt,parsep=0pt]
\item \textbf{Transcription fidelity}: a matching item exists in the prediction, but its text differs from the reference, for example through paraphrase, truncation, or changed operators and numbers.
\item \textbf{Long-document localization and coverage}: no similar item exists anywhere in the prediction, indicating that the relevant content was not found.
\item \textbf{Relational structure reconstruction}: the content is present but placed incorrectly, such as SoA marks in the wrong cells, endpoints under the wrong objective, or incorrect epoch order.
\item \textbf{Domain concept grounding}: the prediction selects a different entity, category, or level than the one requested, such as an incorrect encounter attribute.
\item \textbf{Execution reliability}: the run produces no usable output, such as a missing, empty, or unparseable file.
\end{itemize}
Figure~\ref{fig:failure-profile} shows the profile pooled over the nine main-text agents.

\subsection{Per-Model Failure Profiles}
\label{app:failure-per-model}
\begin{figure*}[t]
    \centering
    \includegraphics[width=\textwidth]{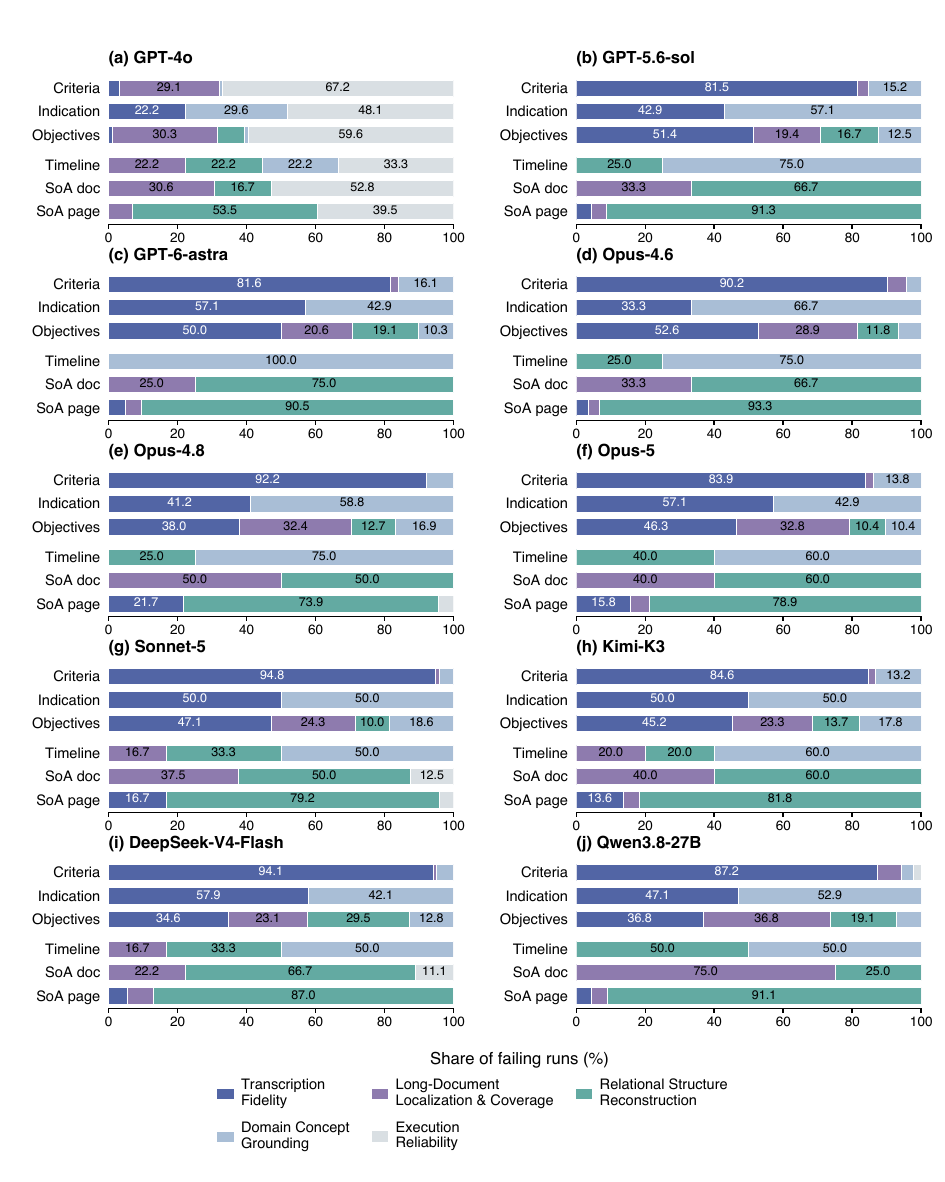}
    \caption{Failure profiles of ten coding agents, the nine main-text agents and GPT-4o, across six extraction tasks. Each bar shows the distribution of dominant failure categories among failing runs for a model and task, normalized to 100\%. These profiles describe failure composition rather than failure rates.}
    \label{fig:failure-profiles-by-agent}
\end{figure*}

Figure~\ref{fig:failure-profiles-by-agent} shows the failure profile of each coding agent. GPT-4o differs from the other agents: execution reliability is its largest failure category in five of the six tasks. The remaining agents show broadly similar profiles within each task and differ more in how often they fail than in which categories dominate.

\subsection{Additional Examples}
\label{app:failure-examples}
Table~\ref{tab:failure-examples} lists additional failures from each category.

\begin{table*}[t]
\centering
\small
\setlength{\tabcolsep}{4pt}
\renewcommand{\arraystretch}{1.15}
\begin{tabular}{@{}>{\raggedright\arraybackslash}p{0.15\textwidth}>{\raggedright\arraybackslash}p{0.10\textwidth}>{\raggedright\arraybackslash}p{0.36\textwidth}>{\raggedright\arraybackslash}p{0.33\textwidth}@{}}
\toprule
Category & Task & Reference & Prediction \\
\midrule
Transcription fidelity & Criteria & Disease duration of more than 10 years in participants with an EDSS$\leq$2.0 at screening & Disease duration of more than 10 years in participants with an EDSS$<$2.0 at screening \\
Transcription fidelity & Endpoints & The cumulative change in EDSS scores measured in area under the concentration-time curve (AUC) by Week 96 & The cumulative change in EDSS scores, measured in AUC by Week 96 \\
Localization and coverage & Objectives & To evaluate progression-free survival (PFS) & Missing; 9 of the 11 reference objectives are omitted \\
Relational structure & Objectives & Endpoint ``The change in low-contrast visual acuity from baseline to Weeks 48 and 96'' under the exploratory objective & The same endpoint attached to a different objective \\
Domain grounding & Indication & Relapsed/Refractory Hematologic Malignancies & relapsed/refractory Multiple Myeloma (MM) \\
Execution reliability & Criteria & -- & No output file written \\
\bottomrule
\end{tabular}
\caption{Additional failure examples.}
\label{tab:failure-examples}
\end{table*}

\section{\ourmethod{} Details}
\label{app:trialscribe}

\subsection{Development Protocols}
\label{app:dev-protocols}
The skills and audit checklists of \ourmethod{} were developed on a small set of protocols per task:
\begin{itemize}[leftmargin=*,topsep=2pt,itemsep=1pt,parsep=0pt]
\item \textbf{Indication} (8): NCT00874614, NCT00942578, NCT02264574, NCT02320149, NCT02354586, NCT02436668, NCT02709512, NCT04573309.
\item \textbf{Eligibility criteria} (8): NCT02320149, NCT03439670, NCT03732820, NCT04770753, NCT04896008, NCT05042362, NCT05523895, NCT05740566.
\item \textbf{Objectives and endpoints} (8): NCT00874614, NCT02320149, NCT02727660, NCT03214380, NCT03439670, NCT03470922, NCT03732820, NCT03962543.
\item \textbf{SoA Tables} (1): NCT04573309.
\item \textbf{Structured timeline} (2): NCT00874614, NCT00942578.
\end{itemize}
These protocols form the development split and are excluded from the test split, on which all results in this paper are reported.

\subsection{Task Skills}
\label{app:skills}
Tables~\ref{tab:skill-indication}-\ref{tab:skill-timeline} are illustrations of the skills.

\subsubsection{Indication Extraction}
The indication skill (Table~\ref{tab:skill-indication}) targets the disease being treated or prevented rather than the enrolled population. It covers four points.
\begin{itemize}[leftmargin=*,topsep=2pt,itemsep=1pt,parsep=0pt]
\item \textbf{Source selection.} The agent collects candidate wording from three locations, the full title, an explicit indication field in the synopsis, and the primary objective, and decides among them by what each states. A dedicated indication field is preferred when it agrees with the trial purpose, and a shorter title may not remove a disease-defining qualifier. Substudy cohorts, amendment history, and background examples are excluded, and for prevention trials the prevented disease is taken.
\item \textbf{Span normalization.} The output is the disease noun phrase: enrolment boilerplate such as ``treatment of'' or ``patients with'' and age wording are removed, while disease-defining stage, severity, histology, resistance, and biomarkers are kept. Eligibility restrictions that follow the disease in a title are not appended.
\item \textbf{Verbatim copying and entry count.} The phrase is copied with its original spelling, case, and abbreviations, without expansion. A combined phrase that names one diagnostic population stays one entry; it is split only when the protocol states separate indications, and alternatives are never listed to hedge.
\end{itemize}

\begin{table*}[p]
\centering
\footnotesize
\begin{tabular}{@{}>{\raggedright\arraybackslash}p{\dimexpr\textwidth-2\tabcolsep\relax}@{}}
\toprule
{\ttfamily \mbox{-{}-{}-{}}\par
\mbox{name:~indication-{}extraction}\par
\mbox{description:~Use~when~extracting~or~auditing~the~treated~or~prevented~indication~of~a~clinical}\par
\mbox{~~trial~protocol,~especially~when~wording,~population~restrictions~or~multiple~diseases~are}\par
\mbox{~~ambiguous.}\par
\mbox{-{}-{}-{}}\par}
\par\smallskip
\textbf{\small Extract the indication, not the population sentence}
\par\smallskip
Write \texttt{/app/output/indications.json} as \texttt{\{"descriptions":["source disease phrase"]\}}. Use only this protocol. The source helper is \texttt{python3 /tools/pdf.py}: \texttt{search 'indication|disease|study population'}, \texttt{pages 1 3}, or \texttt{render 18}. [...]
\par\smallskip
\textbf{Resolve the source phrase}
\par\smallskip
Collect the disease wording from (a) the current full title, (b) an explicit Indication/disease field in the synopsis, and (c) the purpose/primary objective. Do this before deciding. Save the competing source phrases, pages and your selection rationale in \texttt{/app/output/source\_map.json}. [...]
\par\smallskip
Resolve the three source locations by what each states, not by a fixed title-first or longest-phrase rule. A dedicated Indication field containing a disease noun phrase is the preferred literal span when it agrees with the actual trial purpose. A shorter title must not erase the field's resectability, stage, resistance or other disease-defining qualifier. A field written as a treatment/population sentence needs parsing: separate the disease from intervention, symptom endpoints, age and enrolment boilerplate. [...]
\par\smallskip
Return the disease noun phrase. Remove “treatment of”, “patients/subjects with”, age-group wording and other enrolment boilerplate. Retain disease-defining stage, severity, histology, recurrence/resistance and biomarkers when the indication itself states them. [...]
\par\smallskip
Copy the selected phrase verbatim, including its case, hyphenation, spelling and printed abbreviation. Do not expand acronyms or modernize terminology from memory. If OCR appears corrupted, inspect the page and corroborate in another source location. Do not silently change a medical word into another plausible word.
\par\smallskip
Preserve a source's combined disease phrase when it is presented as one indication for a common diagnostic population. The presence of "or", two names or two abbreviations alone does not establish that separate output entries are required. Look for explicit separate indications, diagnostic cohorts or primary anatomical sites before splitting. [...]
\par\smallskip
\textbf{Audit and repair}
\par\smallskip
Independently compare candidate against all three source locations. Check that it preserves the chosen source span, has not dropped disease qualifiers in favor of a shorter title, has not split a combined population without source evidence, is the target disease, is neither an overlong population sentence nor an unsupported short label, retains required qualifiers, and has the right number of distinct conditions. [...]\\
\bottomrule
\end{tabular}
\caption{The \ourmethod{} skill for indication extraction (abridged).}
\label{tab:skill-indication}
\end{table*}

\subsubsection{Eligibility Criteria Extraction}
The criteria skill (Table~\ref{tab:skill-criteria}) aims at a complete, verbatim list. It covers five points.
\begin{itemize}[leftmargin=*,topsep=2pt,itemsep=1pt,parsep=0pt]
\item \textbf{Locating the section.} The agent finds the detailed eligibility section rather than the synopsis or table of contents, identifies where the inclusion and exclusion lists start and end, and follows continuation pages and cohort-specific sections, checking the page after each apparent end of a list.
\item \textbf{Source inventory.} Before extraction, the agent records the section pages and the printed numbering and counts per category and cohort, so that the output can later be checked against counts taken from the source.
\item \textbf{Granularity.} Each top-level numbered criterion becomes one item. Sub-items, qualifications, and tables stay inside their parent criterion, and separately numbered criteria are not merged.
\item \textbf{Verbatim text and encoding.} Criterion text is built from exact source spans in code rather than retyped, with running headers, footers, and top-level numbers removed but units, inequalities, exceptions, and printed dashes preserved. Paragraphs, sub-lists, and tables are encoded in HTML. Pages whose text layer loses characters or symbols are checked visually, and an empty text layer is not taken as evidence that no criteria exist.
\end{itemize}

\begin{table*}[p]
\centering
\footnotesize
\begin{tabular}{@{}>{\raggedright\arraybackslash}p{\dimexpr\textwidth-2\tabcolsep\relax}@{}}
\toprule
{\ttfamily \mbox{-{}-{}-{}}\par
\mbox{name:~criteria-{}extraction}\par
\mbox{description:~Use~when~extracting~or~auditing~inclusion~and~exclusion~criteria~from~a~full~clinical}\par
\mbox{~~trial~protocol,~including~numbered~lists,~cross-{}page~continuations,~cohorts~and~laboratory~tables.}\par
\mbox{-{}-{}-{}}\par}
\par\smallskip
\textbf{\small Complete criteria with source-preserving text}
\par\smallskip
Output \texttt{/app/output/criteria.json} using the original task schema. Do not invent criteria from medical knowledge. \texttt{/app/pages.json} is a one-based physical-page array when addressed as page N -> index N-1. [...]
\par\smallskip
\textbf{Locate, inventory, extract}
\par\smallskip
\begin{enumerate}[leftmargin=*,topsep=1pt,itemsep=0pt,parsep=0pt]
\item Find the authoritative detailed eligibility section, not just the synopsis or table of contents. Identify inclusion start/end, exclusion start/end, continuation pages, and separate cohort/substudy eligibility sections. [...]
\item Save \texttt{/app/output/source\_map.json}: section page ranges; printed top-level numbering and counts by category/cohort; any gaps or intentionally omitted duplicate synopsis. Counts must come from the source, not from your output.
\item One output item per source top-level numbered criterion. Subletters, roman numerals, qualifications, exceptions and tables remain inside their parent. Separate genuinely separate numbered criteria even if topics overlap. Preserve additional cohort criteria without duplicating a repeated summary of the same list.
\item Prefer constructing strings from exact source spans in Python instead of retyping/paraphrasing them. Join wrapped lines only within the same paragraph. Strip running headers/footers and top-level list numbers; preserve actual content, units, inequalities, exceptions, references, punctuation and abbreviation spelling. [...]
\item Encode paragraphs as \texttt{<p>}, sublists as \texttt{<ul>}/\texttt{<ol>} containing \texttt{<li><p>…</p></li>}, and actual tabular criteria as \texttt{<table>}. Escape literal \texttt{<}, \texttt{>} and \texttt{\&} in text. Preserve meaningful printed dash/label text inside subpoints; don't convert a clinical minus sign into a bullet or discard words near numbering. In particular, when the printed subitem begins with a literal dash, retain that dash in the subitem text even inside \texttt{<li><p>...</p></li>}. [...]
\item Use Inclusion/Exclusion exactly, IN01… / EX01… by category, and save valid JSON. Compare output category counts and list boundaries to the source inventory.
\end{enumerate}
\par\smallskip
\textbf{Audit the entire list}
\par\smallskip
Use source numbering as a checklist. Verify every source top-level item has exactly one output counterpart and vice versa; inspect the final item and cross-page joins. Check missing tails, split subcriteria, omitted tables, negation, decimal thresholds, units and category inversions. For a large list, audit it in bounded page batches; do not inspect only the first few examples and declare the whole list correct. [...]
\par\smallskip
An empty list is valid only after locating and checking where eligibility would be stated, including scanned relevant pages. Readability failure means uncertainty, not a supported empty answer. Report concrete missing/incorrect rows with source pages and suggested minimal corrections. [...]\\
\bottomrule
\end{tabular}
\caption{The \ourmethod{} skill for eligibility criteria extraction (abridged).}
\label{tab:skill-criteria}
\end{table*}

\subsubsection{Objectives and Endpoints Extraction}
The objectives skill (Table~\ref{tab:skill-objectives}) centers on attaching each endpoint to the right objective. It covers five points.
\begin{itemize}[leftmargin=*,topsep=2pt,itemsep=1pt,parsep=0pt]
\item \textbf{Two inventories.} Objectives and endpoints are often defined in different sections, so the agent builds an inventory of each, compares the synopsis with the detailed definitions, and records each endpoint with its proposed parent objective and the evidence for that link.
\item \textbf{List boundaries.} A page break does not end a list. For each endpoint subsection, the agent reads until the next heading of the same or a higher level and records the first and last items and the heading that closes the list.
\item \textbf{Field conventions.} Objective and endpoint texts are copied as complete sentences, levels are normalized to primary, secondary, or exploratory, and each endpoint's level is read from its own heading rather than inherited from its objective.
\item \textbf{Attachment.} Endpoints are linked through explicit table rows or cross-references when available, and otherwise by matching the measurement, clinical domain, population, comparison, and time frame to each objective, never by list position. An endpoint may appear under several objectives when the protocol says so, and an objective may have none.
\end{itemize}

\begin{table*}[p]
\centering
\footnotesize
\begin{tabular}{@{}>{\raggedright\arraybackslash}p{\dimexpr\textwidth-2\tabcolsep\relax}@{}}
\toprule
{\ttfamily \mbox{-{}-{}-{}}\par
\mbox{name:~objectives-{}extraction}\par
\mbox{description:~Use~when~extracting~or~auditing~clinical~trial~objectives~and~their~attached}\par
\mbox{~~endpoints,~including~separate~endpoint~sections,~mixed~hierarchy~levels~and~objective-{}endpoint}\par
\mbox{~~tables.}\par
\mbox{-{}-{}-{}}\par}
\par\smallskip
\textbf{\small Recover objectives and endpoint attachments}
\par\smallskip
Output \texttt{/app/output/objectives.json} in the original task schema. Use only the source. Use \texttt{python3 /tools/pdf.py search 'objective|endpoint|efficacy variable|outcome|estimand'} to locate sections, \texttt{pages N M} for bounded text, and \texttt{render N} followed by read for tables or damaged text. Do not confuse physical PDF pages with printed page numbers.
\par\smallskip
\textbf{Build two inventories before linking them}
\par\smallskip
Find the detailed objectives and all endpoint/efficacy/safety variable definitions, which can appear in different sections. Compare the synopsis with the detailed definitions; short summaries may merge separately defined endpoints or omit secondary and exploratory endpoints. [...]
\par\smallskip
Write \texttt{/app/output/source\_map.json} with source pages, each objective's printed level label and text, and each endpoint's own label/text plus proposed parent and attachment evidence. Use one row per explicit objective, not one per clause or drug arm. [...]
\par\smallskip
\textbf{\emph{Close each source list at a verified section boundary}}
\par\smallskip
A page ending is not an endpoint-list ending. For each endpoint subsection, read through every continuation page until the next same-level or higher-level heading. If the last retrieved page contains endpoint bullets, read the following page even when it has no repeated endpoint heading. [...]
\par\smallskip
In source\_map.json, record an \texttt{endpoint\_sections} inventory: subsection heading, physical start/end pages, top-level item count, first/last item, and the following heading that proves the list ended. A clipped tool response or a fixed page window does not establish that boundary. [...]
\par\smallskip
\textbf{Preserve each field independently}
\par\smallskip
\begin{itemize}[leftmargin=*,topsep=1pt,itemsep=0pt,parsep=0pt]
\item \texttt{objectiveText}: copy the complete objective sentence or printed list item; do not shorten a “To assess …” sentence into its topic or combine separate objectives.
\item \texttt{objectiveLevel}: PRIMARY, SECONDARY or EXPLORATORY (including tertiary/other).
\item \texttt{levelVerbatim}: the level label actually used in the source at that objective. Prefer the direct adjacent level cell/label when present (e.g. “Primary”); do not append “Objectives”, numeric section prefixes or invented capitalization. Preserve meaningful phase/cohort distinctions when they are part of the level label. Check this field explicitly; correct objective text with the wrong label is incomplete. For a plain category heading consisting only of Primary, Secondary, Exploratory, Tertiary, Safety or Other followed by Objective(s), use the category word itself as the level-label span, preserving its printed capitalization. [...]
\item \texttt{endpointText}: the detailed printed endpoint sentence/list item, including its timepoint, definition and units. Separate explicitly enumerated/co-primary endpoints instead of replacing them by an earlier summary sentence. [...]
\item \texttt{endpointLevel}: read the endpoint's own heading. Do not inherit the objective level automatically. A primary efficacy objective may have secondary/exploratory efficacy endpoints; they still belong to that primary efficacy objective.
\end{itemize}
\par\smallskip
\textbf{Attach by the question measured, not by the hierarchy label}
\par\smallskip
Prefer explicit table rows or cross-references. Otherwise compare the measurement, clinical domain, population, intervention comparison and time frame to the question posed by each objective. [...]
\par\smallskip
The same endpoint may legitimately occur under more than one objective if the source states this; do not globally deduplicate away valid attachments. An objective may have no endpoint: use [] when none is supplied. [...]
\par\smallskip
\textbf{Audit and minimal fix}
\par\smallskip
Independently check objective count, every verbatim level label, endpoint inventory, endpoint levels and each attachment. Count source endpoints before comparing output. For every unassigned source endpoint, either identify its objective or explain the source ambiguity; for every emitted endpoint, identify source evidence. Inspect cross-page table continuations and detailed definitions, not just the synopsis. Independently open the page after each claimed endpoint-list ending and verify the recorded closing heading. [...]\\
\bottomrule
\end{tabular}
\caption{The \ourmethod{} skill for objectives and endpoints extraction (abridged).}
\label{tab:skill-objectives}
\end{table*}

\subsubsection{SoA Table Extraction}
For SoA table extraction, a skill file (Tables~\ref{tab:skill-soa-a} and~\ref{tab:skill-soa-b}) defines the workflow, and a PDF-processing skill supplies extraction patterns. The file covers three points.
\begin{itemize}[leftmargin=*,topsep=2pt,itemsep=1pt,parsep=0pt]
\item \textbf{Triage.} A triage script first measures the page and selects one of three routes: a usable text layer, a text layer that exists but is unusable, or no text layer.
\item \textbf{Reading routes.} With a usable text layer, the table is parsed with pdfplumber following the PDF skill. With an unusable text layer, the grid may be taken from the text layer, but every character is read from the rendered page. Without a text layer, the table is built from the rendered page alone.
\item \textbf{Fix loop.} The agent runs a structural checker on its HTML until it reports no problems, such as rows that do not fill the grid, and then reads each visit column header back against the rendered page to confirm the number, order, and levels of the visit columns. Header cells that already match the page are not rewritten, because a single dropped footnote marker changes the identity of every column beneath it.
\end{itemize}

\begin{table*}[p]
\centering
\footnotesize
\begin{tabular}{@{}>{\raggedright\arraybackslash}p{\dimexpr\textwidth-2\tabcolsep\relax}@{}}
\toprule
\textbf{\small Reading an SoA page out of a PDF}
\par\smallskip
The job is to reproduce one page of a clinical trial protocol — a Schedule of Activities table — as HTML.
\par\smallskip
There are two ways to read the page, and \textbf{which one works is a property of the file, not a preference.} Text extraction on a page whose text layer is broken produces confident, wrong output. [...]
\par\smallskip
\textbf{Step 1 — Triage (always first, costs one command)}
\par\smallskip
{\ttfamily \mbox{python3~/app/triage\_pdf.py~/app/inputs/protocol.pdf}\par}
\par\smallskip
It prints one of three routes and the numbers behind it. Follow the route it gives you. The thresholds sit in wide empty gaps in the measured distribution, so a borderline page is rare — if you get one, prefer VISION.
\par\smallskip
\begin{tabular}{@{}lll@{}}
\toprule
route & what it means & go to \\
\midrule
\texttt{TEXT} & clean text layer, ruling lines present & Step 2a \\
\texttt{VISION} & text layer exists but is unusable & Step 2b \\
\texttt{VISION\_ONLY} & no text layer at all — a scan or a picture & Step 2c \\
\bottomrule
\end{tabular}
\par\smallskip
Do not skip this and do not argue with it. "The text extraction looked plausible" is exactly how the failures happen: a broken encoding yields well-formed rows full of wrong characters.
\par\smallskip
\textbf{Step 2a — Route TEXT}
\par\smallskip
The text layer is usable. Parse it.
\par\smallskip
{\ttfamily \mbox{python3~-{}c~"import~pdfplumber;}\par
\mbox{~~print(pdfplumber.open('/app/inputs/protocol.pdf').pages[0].extract\_table()~is~not~None)"}\par}
\par\smallskip
\texttt{pdfplumber} and \texttt{PyMuPDF} are already installed — do not spend turns on \texttt{pip install}.
\par\smallskip
Follow \textbf{\texttt{.claude/skills/pdf/SKILL.md}}, section \emph{SoA Table Extraction}, for the patterns that matter here: \texttt{None} vs \texttt{''} semantics, column groups, \texttt{normalize\_row\_v2}, section-header merging, superscript handling, colspan validation, X-cell variants.
\par\smallskip
\textbf{If \texttt{extract\_table()} returns \texttt{None} or an empty table, stop and switch to Step 2b.} That is the one case triage can miss when PyMuPDF is unavailable and the ruling-line check could not run.
\par\smallskip
Then go to Step 3.
\par\smallskip
\textbf{Step 2b — Route VISION (text layer present but unusable)}
\par\smallskip
Two different things are broken under this route, and the triage output tells you which:
\par\smallskip
\textbf{"decodes to junk"} — some fraction of the characters are wrong. The \emph{geometry} is still sound, so you may use \texttt{pdfplumber} to find the grid: how many columns, where the row boundaries are, which cells are spanned. But \textbf{every character you emit must be read off the rendered image}, not copied from the extracted text. [...]
\par\smallskip
\textbf{"no ruling lines"} — \texttt{extract\_table()} will return nothing, because it finds tables by their lines. There is no grid to recover from the text layer. Build the table from the image.
\par\smallskip
Either way, render it:
\par\smallskip
{\ttfamily \mbox{pdftoppm~-{}png~-{}r~200~-{}f~1~-{}l~1~/app/inputs/protocol.pdf~/tmp/page}\par}
\par\smallskip
Re-render a region at higher DPI (300, 400) whenever a character is doubtful. That is cheap, and it is the only way to settle it.
\par\smallskip
If the render shows no schedule table on the page at all — just body text — say so plainly in your final message and write what the page does contain. Do not invent a table to fill the file.
\par\smallskip
Then go to Step 3.
\par\smallskip
\textbf{Step 2c — Route VISION\_ONLY (no text layer)}
\par\smallskip
Same as Step 2b, with one thing to be explicit about:
\par\smallskip
\textbf{Do not install or use \texttt{pytesseract}.} The \texttt{tesseract} binary is not in this image, so OCR cannot run — the import will succeed and the call will fail. Your own vision is the extraction method here, not a fallback for it.
\par\smallskip
{\ttfamily \mbox{pdftoppm~-{}png~-{}r~200~-{}f~1~-{}l~1~/app/inputs/protocol.pdf~/tmp/page}\par}
\par\smallskip
Then go to Step 3.\\
\bottomrule
\end{tabular}
\caption{The \ourmethod{} skill file for SoA table extraction (part 1 of 2): triage and reading routes.}
\label{tab:skill-soa-a}
\end{table*}

\begin{table*}[p]
\centering
\footnotesize
\begin{tabular}{@{}>{\raggedright\arraybackslash}p{\dimexpr\textwidth-2\tabcolsep\relax}@{}}
\toprule
\textbf{Step 3 — The fix loop}
\par\smallskip
Applies to every route. A table that reads plausibly can still attach every mark to the wrong visit, and that scores zero.
\par\smallskip
\textbf{\texttt{/app/verify\_soa.py}} reads only your own HTML — there is no answer key in this container — and reports structural problems.
\par\smallskip
{\ttfamily \mbox{\#~fix~until~the~structural~report~is~clean}\par
\mbox{while~!~python3~/app/verify\_soa.py~/app/output/soa\_table.html;~do}\par
\mbox{~~~~:~read~the~findings,~edit~the~HTML,~run~it~again}\par
\mbox{done}\par}
\par\smallskip
Then read your own column names back against the page:
\par\smallskip
{\ttfamily \mbox{python3~/app/verify\_soa.py~/app/output/soa\_table.html~-{}-{}keys}\par}
\par\smallskip
\texttt{-{}-{}keys} prints, for every visit column, the full header path a reader would follow to name it:
\par\smallskip
{\ttfamily \mbox{col~~~1:~Screening~Assessments~|~Within~4~weeks~(Day~-{}28~to~0)}\par
\mbox{col~~~2:~Screening~Assessments~|~Within~1~week~(Day~-{}7~to~0)}\par
\mbox{col~~~3:~Cycle~1~|~1~(1)}\par}
\par\smallskip
Put that beside the rendered page and walk the columns left to right:
\par\smallskip
\begin{enumerate}[leftmargin=*,topsep=1pt,itemsep=0pt,parsep=0pt]
\item \textbf{The count.} As many printed columns as visit columns on the page.
\item \textbf{The order.} Column \emph{n} names the \emph{n}-th visit column on the page. An off-by-one here is the single most common way to score zero: every mark lands one visit over, so nothing matches even though the table looks right and the checker is happy.
\item \textbf{The levels.} Each key carries every level the page shows for that column — period, visit, visit number — and no level belonging to a neighbour.
\end{enumerate}
\par\smallskip
If column 1's key contains the row-label heading ("Study Procedure", "Visit Number", "Assessment/Procedure"), the label has leaked into the column header and every visit name is shifted by one. That is a shift, not a typo: fix the corner cell, do not patch the names.
\par\smallskip
\textbf{\emph{Do not rewrite a header that is already right}}
\par\smallskip
Rewriting headers that did not need it can cancel out everything the loop gained elsewhere.
\par\smallskip
The reason: a column's identity is its entire header path, including any footnote marker. Drop one \texttt{<sup>a</sup>} from a cell with \texttt{colspan="15"} and all 15 columns under it stop matching.
\par\smallskip
So:
\par\smallskip
\begin{itemize}[leftmargin=*,topsep=1pt,itemsep=0pt,parsep=0pt]
\item If the structural report is clean and the rendered page agrees with your \texttt{-{}-{}keys} output, \textbf{you are done. Submit it.}
\item Change a header cell only when you can point at the specific thing on the page that disagrees with it.
\item When you do change one, copy the cell's text exactly, superscripts included.
\end{itemize}
\par\smallskip
\textbf{\emph{What the checker cannot tell you}}
\par\smallskip
It has no access to the page. It cannot see a table that is internally perfect but uniformly shifted, or a cell whose text is misread. Many ways of failing this task are invisible to it — which is why the read-back against the render is not optional, and also why a clean report is not permission to start editing.
\par\smallskip
\textbf{Output}
\par\smallskip
Write the finished table to \texttt{/app/output/soa\_table.html}.
\par\smallskip
Detailed extraction technique: \texttt{.claude/skills/pdf/SKILL.md}.\\
\bottomrule
\end{tabular}
\caption{The \ourmethod{} skill file for SoA table extraction (part 2 of 2): the fix loop.}
\label{tab:skill-soa-b}
\end{table*}

\subsubsection{Structured Timeline Extraction}
For structured timeline extraction, an integrated timeline skill (Table~\ref{tab:skill-timeline}) is supported by a reference note on USDM v4.0. The timeline skill covers four points.
\begin{itemize}[leftmargin=*,topsep=2pt,itemsep=1pt,parsep=0pt]
\item \textbf{Source inventory.} The agent first inventories stages, contacts and activity rows with their pages and quotations in a sidecar source map, and treats missing evidence as null rather than a guessed default.
\item \textbf{Epochs.} Stages are recorded in study order with a source-supported classification. A visit, an arm or a residence setting is not a stage, and distinct stages are not merged because their types coincide.
\item \textbf{Encounters.} An encounter is a planned participant contact. Daily contacts in an inpatient period are expanded into day-specific records, {and} telephone components tied to a numbered visit keep both contact modes{.}
\item \textbf{Activities and timing.} Activity definitions are reused across visits and source groupings are preserved. Timing keeps its direction, start/end relationship and window, and refers to existing encounter names.
\end{itemize}

\begin{table*}[p]
\centering
\footnotesize
\begin{tabular}{@{}>{\raggedright\arraybackslash}p{\dimexpr\textwidth-2\tabcolsep\relax}@{}}
\toprule
{\ttfamily \mbox{-{}-{}-{}}\par
\mbox{name:~timeline-{}extraction}\par
\mbox{description:~Extract~and~audit~USDM~epochs,~encounters~and~activity~definitions}\par
\mbox{~~from~a~protocol~with~source~inventories,~explicit~attribute~evidence~and}\par
\mbox{~~compatible~timing~output.}\par
\mbox{-{}-{}-{}}\par}
\par\smallskip
\textbf{\small Source-grounded USDM components}
\par\smallskip
Read \texttt{USDM\_REFERENCE.md}, the supplied instruction and schema. The schema is a four-array projection, not the complete USDM graph. Preserve its field names. Never read gold, study-specific reference workbooks or same-study structured examples. A source plan and earlier audit are fallible model outputs.
\par\smallskip
\textbf{\small Extract with a source inventory}
\par\smallskip
\begin{enumerate}[leftmargin=*,topsep=1pt,itemsep=0pt,parsep=0pt]
\item Locate study design, schedule tables, all continuation pages, visit descriptions, activity definitions and relevant footnotes. Inspect page images for ambiguous spans, composite rows and damaged text. Track physical PDF pages.
\item Inventory source stages, contacts and activity rows/groups before deciding output counts. Retain source\_id, page, quotation, scope and output row indices in source\_map.json. Explain exclusions. Prioritize a usable timeline.json before elaborating the sidecar; update both when fixing output.
\item For every row, keep a concise source-readable name and label. Put qualifications in description/notes, not synthetic ID-like names or long identity labels.
\item Explicitly review the fields below; missing evidence means null, not a guessed default. An omitted field must not pass unnoticed merely because schema allows it.
\end{enumerate}
\par\smallskip
\textbf{\small Epoch: stage, classification and sequence}
\par\smallskip
Record source-supported sequential stages in study order. Classification belongs in type, e.g. SCREENING, RUN-IN, TREATMENT, WASHOUT, FOLLOW-UP when appropriate. These examples are not an exhaustive ontology. Keep period numbers in name/label; use the same stage class for distinct treatment periods when the source defines them. Do not emit \texttt{run\_in}, generic \texttt{Study phase}, an arm, or a residence setting as a stage classification. Never automatically split a stage at every dose or site change. Never merge genuinely distinct source stages solely because their types coincide.
\par\smallskip
[...]
\par\smallskip
\textbf{\small Encounter: identity, kind, and mode}
\par\smallskip
\begin{enumerate}[leftmargin=*,topsep=1pt,itemsep=0pt,parsep=0pt]
\item Locate the schedule of activities AND its footnotes. Read the relevant visit/contact prose. Render a schedule page when merged headers make text ambiguous. Use a compact working inventory of visit/day columns, not a long planning document.
\item An encounter is a planned participant contact, not an epoch or an individual procedure. Preserve distinct visit numbers and period/day identities. A continuous inpatient period contains daily encounters when daily clinical assessments, blood sampling, or staff-administered treatment are scheduled: expand an explicit daily range into day-specific records. [...] Distinguish conditional/unscheduled contacts from routine contacts.
\item Read footnotes for telephone contacts tied to specific visits. If a numbered visit comprises an onsite component and a later telephone assessment, retain its identity and both contact modes, describe their different timing. [...]
\item Use type "Visit" for clinical study visits; put screening, inpatient, unscheduled, early termination, and follow-up qualifiers in name/label/description, not invented type values such as "Visit (unscheduled)".
\item contactModes is an array of display terms, e.g. ["In Person"], ["Telephone Call"], or both. [...] Use null only when the source and scheduled procedures do not establish a mode.
\item Use short identifiable names/labels containing period and day or visit number. [...]
\end{enumerate}
\par\smallskip
Required final consistency check: if your source evidence states a telephone component is attached to a numbered visit, that visit must include Telephone Call as well as its onsite mode. [...]
\par\smallskip
Keep encounter evidence in the shared source\_map.json, alongside epoch, activity and timing evidence. [...]
\par\smallskip
\textbf{\small Activity: definition, group and scheduled occurrence}
\par\smallskip
Inventory exact schedule row designations and explicit presentation group headings. Read prose/footnotes for additional source-stated procedures. Preserve a composite row as a composite unless the source defines independent activities. Reuse an activity definition across visits; occurrence scheduling does not create new copies.
\par\smallskip
USDM permits grouping activities. [...]
\par\smallskip
\textbf{\small Compact audit and repair}
\par\smallskip
Review the actual source, not only candidate counts. Report specific errors with short literal quotes and physical page numbers. Uncertainty is a warning, not an automatic repair or deletion. [...]
\par\smallskip
Repair only explicitly identified errors after an independent source check; preserve correct values. [...]
\par\smallskip
For timing, FROM occurs BEFORE/AFTER TO, and start/end relationships and windows retain their original meaning. Use readable source event names and period context rather than anonymous TP IDs. A zero offset can legitimately relate distinct coincident events. Do not infer elapsed durations by subtracting nominal day labels. Reference existing encounter names when the endpoint is that encounter; a within-visit procedure endpoint must identify its own source event and period.\\
\bottomrule
\end{tabular}
\caption{The \ourmethod{} skill for structured timeline extraction (abridged; omitted passages are marked [...]).}
\label{tab:skill-timeline}
\end{table*}

\subsection{Verification Loop}
\label{app:verification}
The harness runs the verification loop after every draft. Each phase runs in a new agent session: the draft phase writes the candidate output, and the audit phase receives the protocol and the candidate with read-only access and follows the audit checklist in the task skill. The auditor writes a structured report containing a pass or fail verdict, concrete content issues with the affected output field, the problem, a source-supported correction, and the relevant pages, the source sections it checked, short quotes from the source as evidence, and any unresolved uncertainties.

A gold-free gate then checks the report for schema validity and internal consistency: the report must be well formed, a failing verdict must list concrete issues, a passing verdict must list none, the checked pages must exist in the document, and the report must refer to the candidate. The gate also checks that the candidate follows the task schema. Evidence quotes that cannot be found in the PDF text layer raise a warning without blocking the output. If errors remain, the agent applies minimal source-supported fixes in a new session: it reads the audit feedback, rechecks the source pages, corrects only the reported content or schema errors, and leaves supported text unchanged. The revised output is then audited again, for at most two rounds. An incomplete report triggers one repeated audit. For SoA table extraction, verification instead runs within the agent's current session: the agent repeatedly checks its HTML table against the rendered page and revises it until the checks pass (Figure~\ref{fig:soa-case-iterations}).

\clearpage
\subsection{Ablation of Skills and Verification}
\label{app:ablation}
We ablate the two components of \ourmethod{}, the task skills and the verification loop, on 40 eligibility-criteria protocols randomly sampled from the test split, with Qwen3.8-27B. We compare four configurations: the base coding agent, skills only, verification only, and the full \ourmethod{}. As Table~\ref{tab:ablation-criteria} shows, each component improves criterion F1 on its own, by 2.6 points for skills and 7.7 points for verification, and combining them gives the largest gain, 14.5 points over the base agent. Verification adds more when the skills are present (11.9 vs.\ 7.7 points).

\begin{center}
\small
\begin{tabular}{@{}lccc@{}}
\toprule
Configuration & Skills & Verification & Criterion F1 (\%) \\
\midrule
Base agent & \ding{55} & \ding{55} & 65.4 \\
Skills only & \ding{51} & \ding{55} & 68.0 \\
Verification only & \ding{55} & \ding{51} & 73.1 \\
\ourmethod{} & \ding{51} & \ding{51} & \textbf{79.9} \\
\bottomrule
\end{tabular}
\captionof{table}{Ablation of the task skills and the verification loop on eligibility-criteria protocols, with Qwen3.8-27B.}
\label{tab:ablation-criteria}
\end{center}

\subsection{SoA Table Extraction Case Study}
\label{app:soa-case}
\begin{figure*}[p]
    \centering
    \begin{subfigure}[t]{0.49\textwidth}
        \centering
        \includegraphics[width=\linewidth]{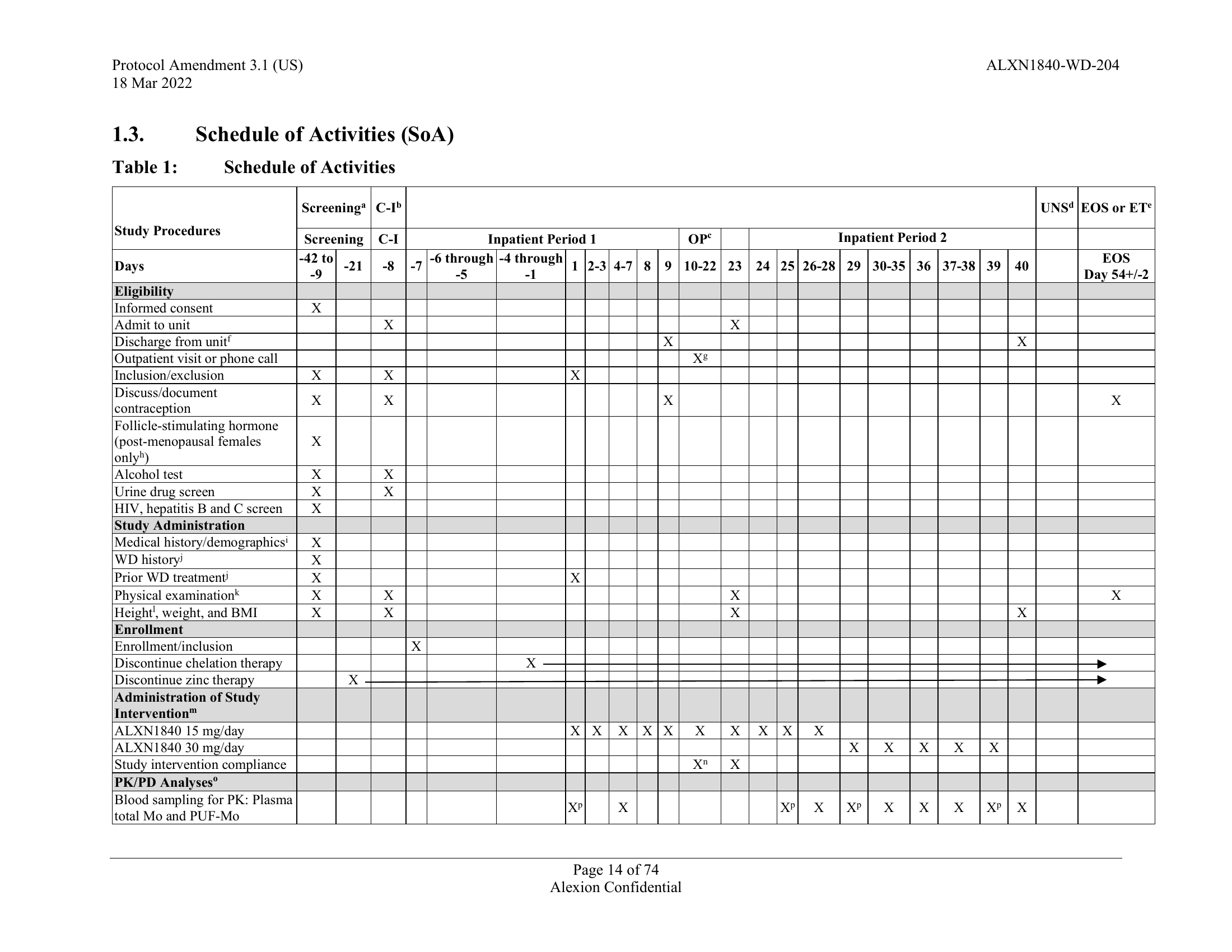}
        \caption{Source page 14.}
    \end{subfigure}\hfill
    \begin{subfigure}[t]{0.49\textwidth}
        \centering
        \includegraphics[width=\linewidth]{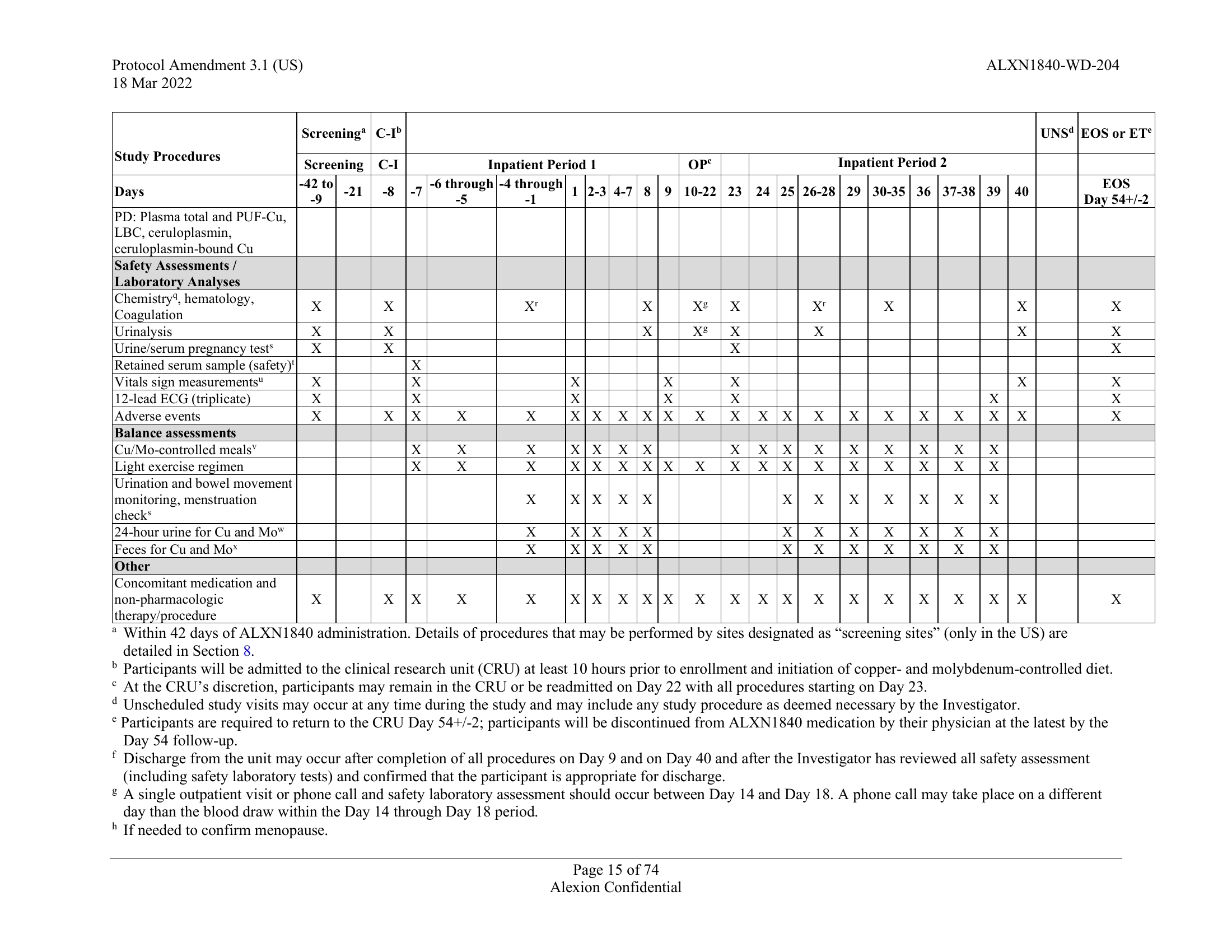}
        \caption{Source page 15.}
    \end{subfigure}
    \par\medskip
    \begin{subfigure}[t]{\textwidth}
        \centering
        \includegraphics[width=0.9\linewidth]{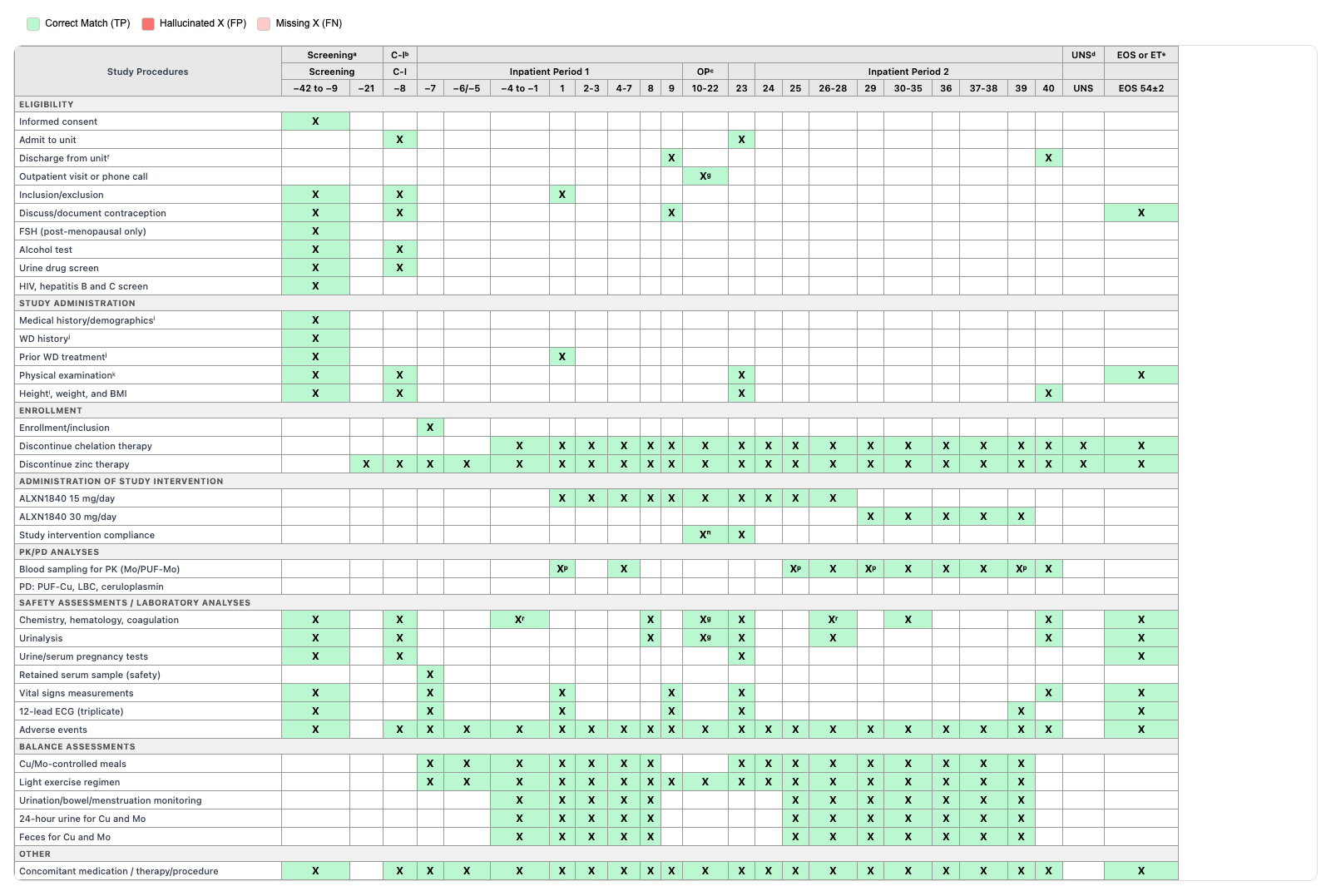}
        \caption{\ourmethod{} output for both pages, compared with the reference.}
    \end{subfigure}
    \caption{SoA table extraction case study: source pages and the \ourmethod{} output.}
    \label{fig:soa-case-output}
\end{figure*}

\begin{figure*}[p]
    \centering
    \begin{subfigure}[t]{0.49\textwidth}\centering\includegraphics[width=\linewidth]{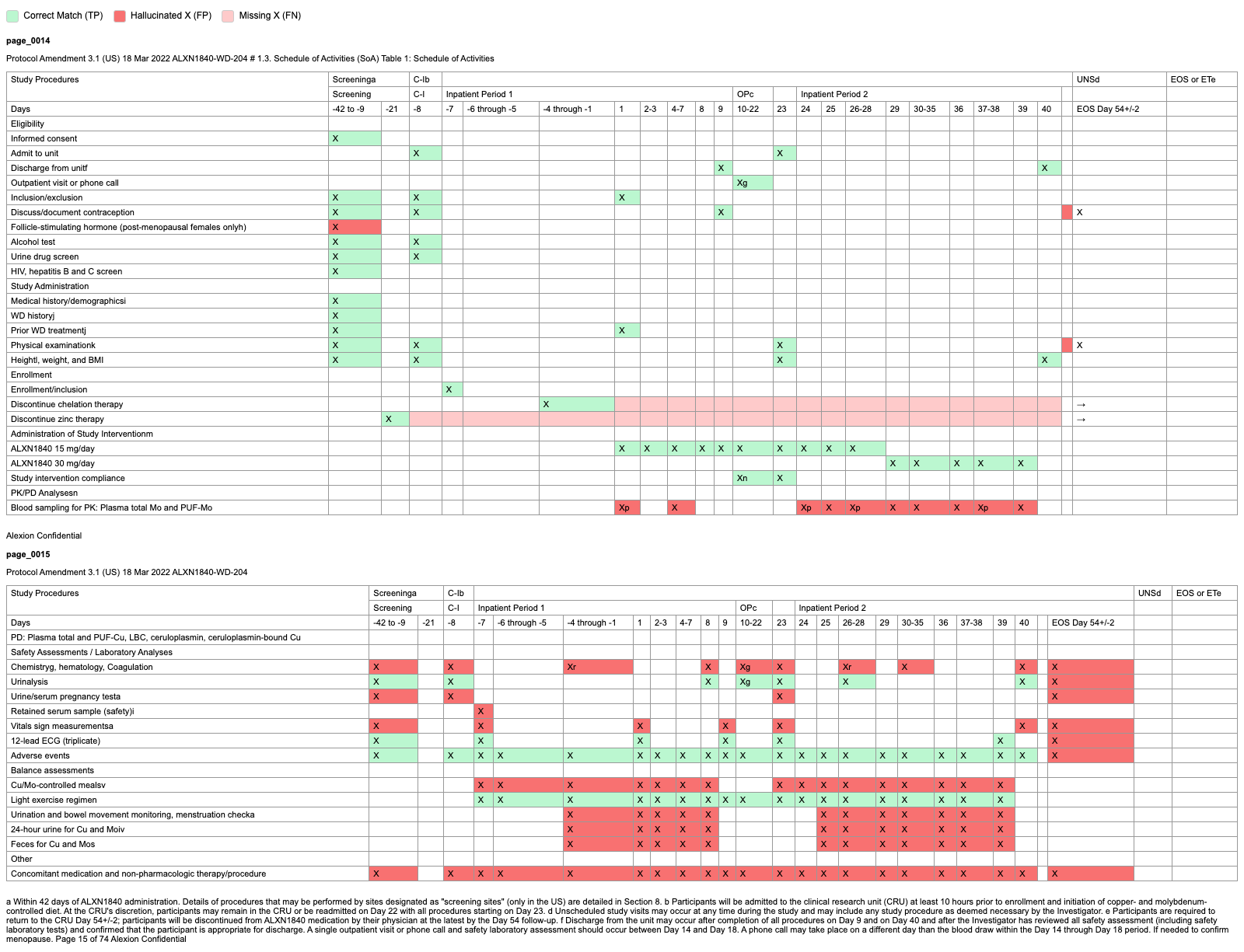}\caption{Mistral OCR.}\end{subfigure}\hfill
    \begin{subfigure}[t]{0.49\textwidth}\centering\includegraphics[width=\linewidth]{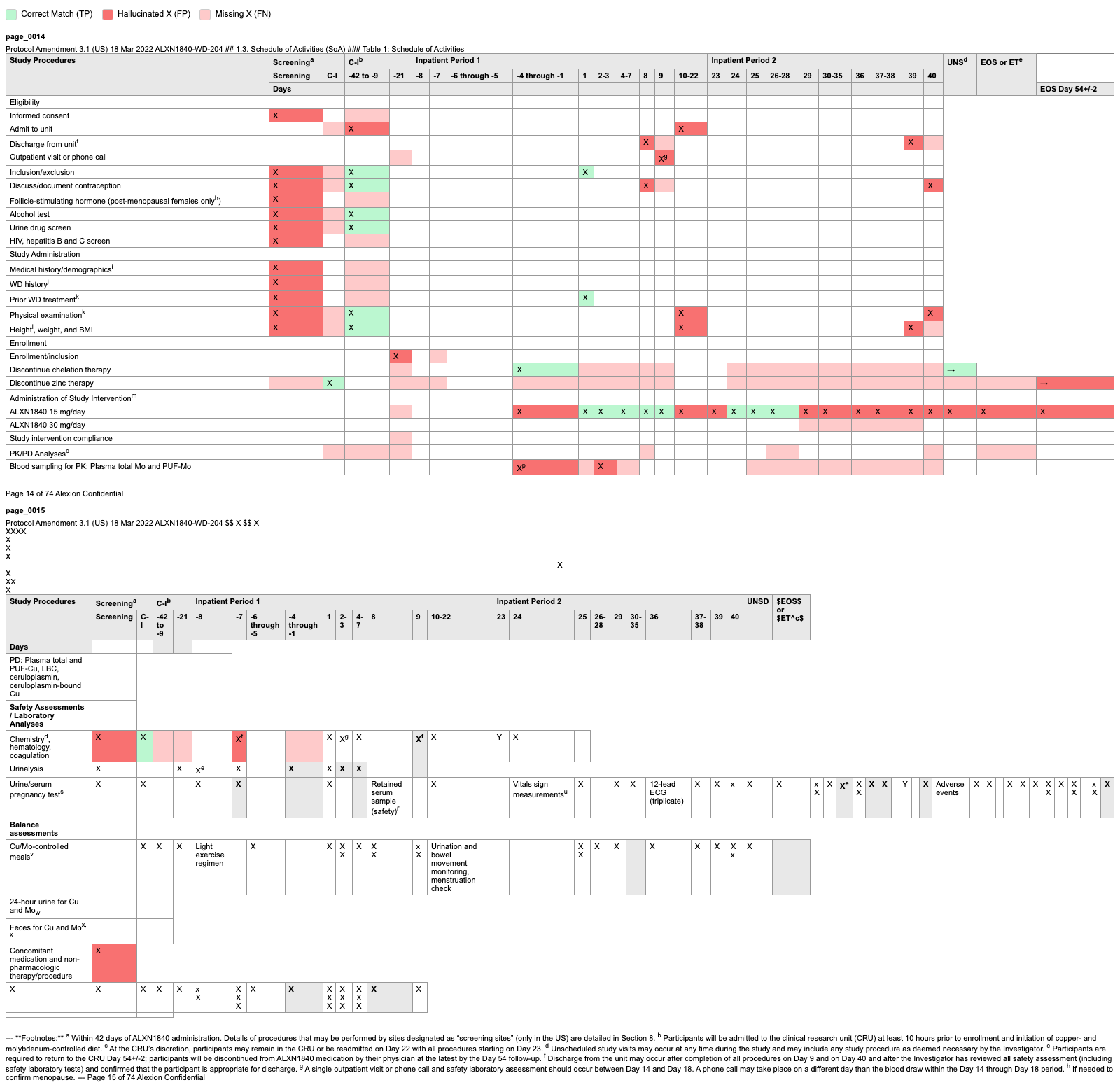}\caption{LightOnOCR.}\end{subfigure}
    \par\medskip
    \begin{subfigure}[t]{0.49\textwidth}\centering\includegraphics[width=\linewidth]{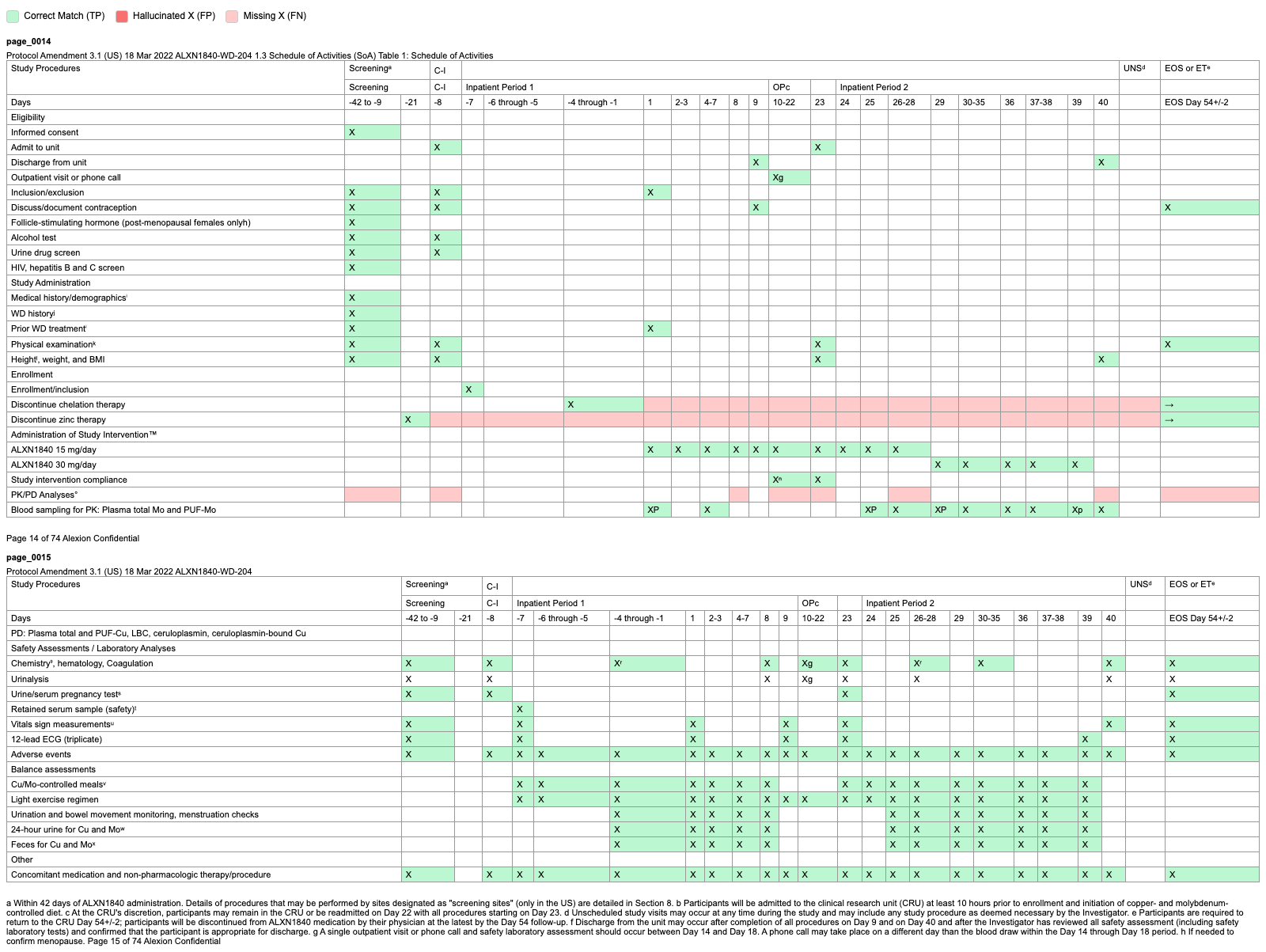}\caption{LandingAI.}\end{subfigure}\hfill
    \begin{subfigure}[t]{0.49\textwidth}\centering\includegraphics[width=\linewidth]{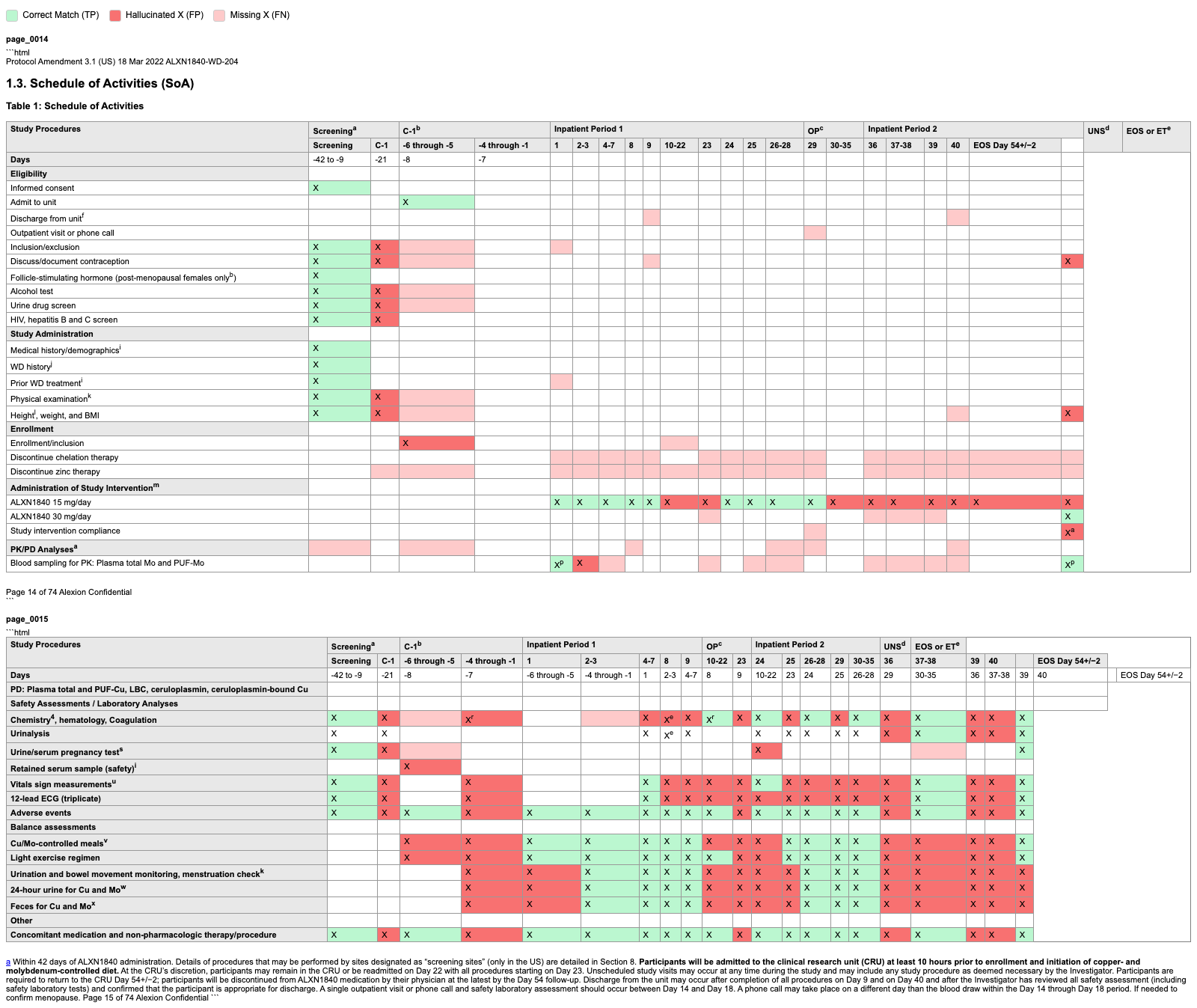}\caption{Qwen3-VL-4B.}\end{subfigure}
    \par\medskip
    \begin{subfigure}[t]{0.49\textwidth}\centering\includegraphics[width=\linewidth]{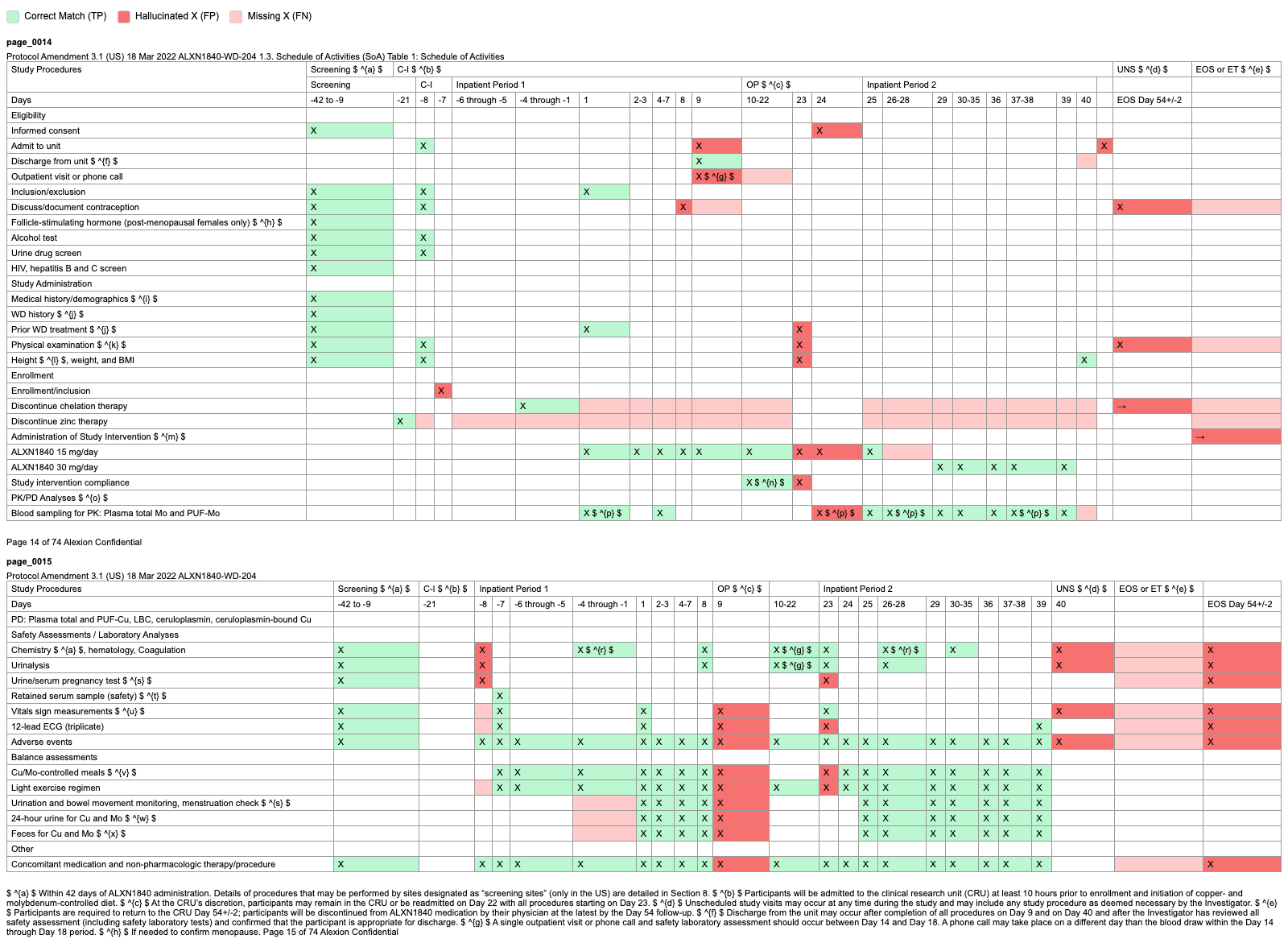}\caption{PaddleOCR-VL.}\end{subfigure}\hfill
    \begin{subfigure}[t]{0.49\textwidth}\centering\includegraphics[width=\linewidth]{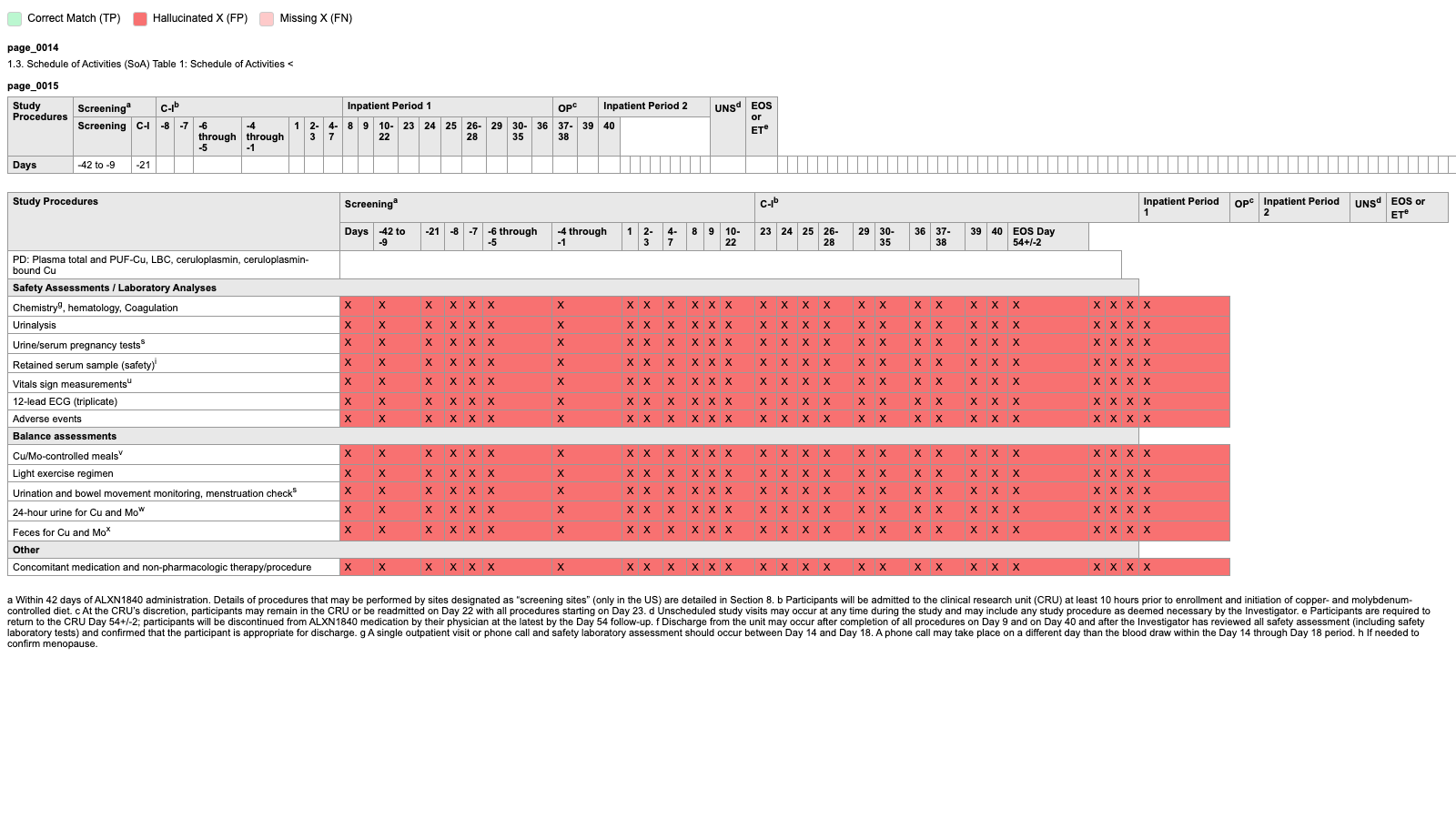}\caption{olmOCR.}\end{subfigure}
    \caption{SoA table extraction case study: OCR and document-understanding baselines on the same pages, annotated against the reference.}
    \label{fig:soa-case-ocr}
\end{figure*}

\begin{figure*}[p]
    \centering
    \begin{subfigure}[t]{\textwidth}\centering\includegraphics[width=0.8\linewidth]{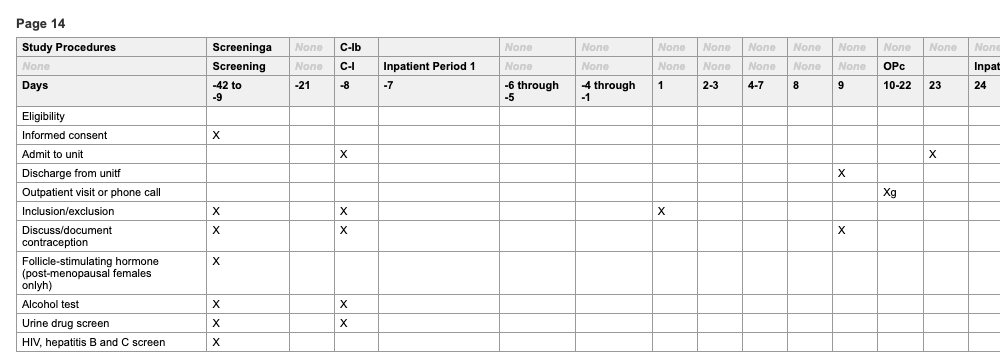}\caption{Pass 1: direct text-layer extraction. Merged cells appear as empty placeholders (``None''), footnote markers are fused into the text (``Screening$^{a}$''), and multi-row labels are broken.}\end{subfigure}
    \par\medskip
    \begin{subfigure}[t]{\textwidth}\centering\includegraphics[width=0.8\linewidth]{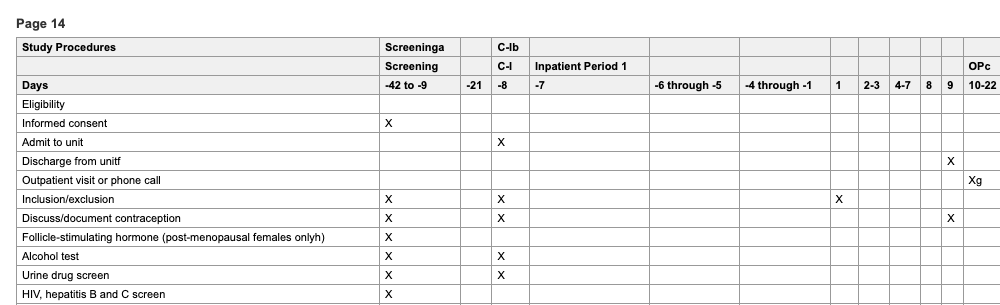}\caption{Pass 2: structural correction. Merged cells are resolved, broken labels are joined, and empty continuation rows are removed; footnote markers are still inline.}\end{subfigure}
    \par\medskip
    \begin{subfigure}[t]{\textwidth}\centering\includegraphics[width=0.8\linewidth]{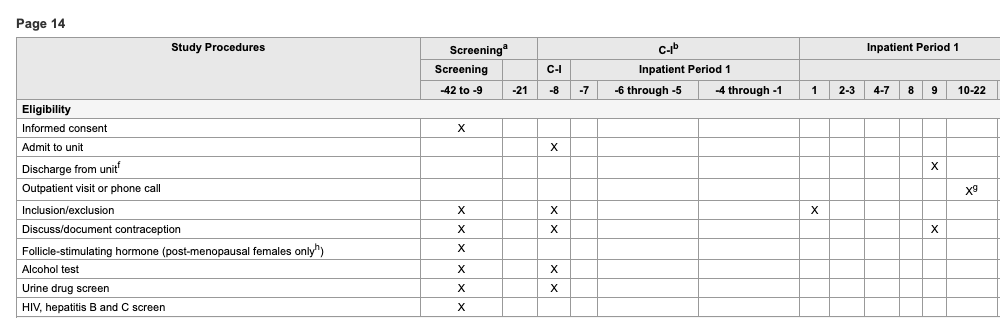}\caption{Pass 3: footnote markers are separated as superscripts, header spans are resolved, and section headers are identified.}\end{subfigure}
    \par\medskip
    \begin{subfigure}[t]{\textwidth}\centering\includegraphics[width=0.8\linewidth]{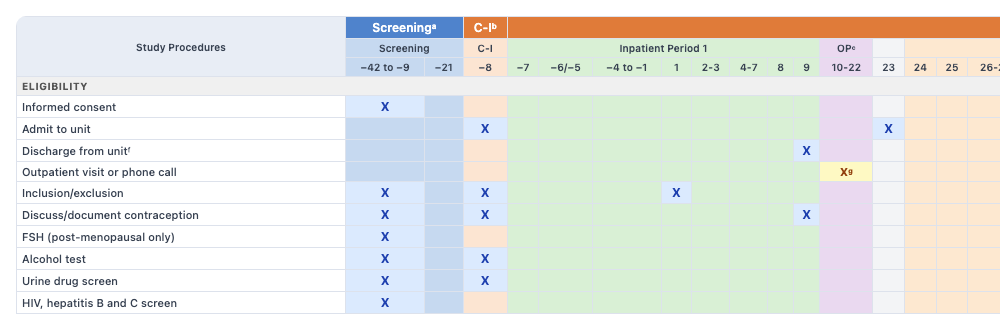}\caption{Pass 4: final output after visual verification against the source page; colors distinguish visit periods and mark variants.}\end{subfigure}
    \caption{SoA table extraction case study of \ourmethod{}.}
    \label{fig:soa-case-iterations}
\end{figure*}

We present a case study on the two-page Schedule of Activities. Figure~\ref{fig:soa-case-output} shows the source pages and the \ourmethod{} output; Figure~\ref{fig:soa-case-ocr} shows the output of six OCR and document-understanding baselines on the same pages, and Figure~\ref{fig:soa-case-iterations} shows how the output changes over successive verification passes. In the annotated outputs, green cells are correctly placed scheduling marks, red cells are marks absent from the reference, and pink cells are reference marks that are missing.

\section{Future Work}
\label{app:future-work}
Beyond SoA tables, protocols contain many other types of tables that downstream clinical systems rely on. Table~\ref{tab:table-types} lists common examples. Extending \ourbench{} and \ourmethod{} to these table types is a natural direction for future work.

\begin{table}[h]
\centering
\small
\setlength{\tabcolsep}{4pt}
\begin{tabular}{@{}>{\raggedright\arraybackslash}p{0.45\columnwidth}>{\raggedright\arraybackslash}p{0.5\columnwidth}@{}}
\toprule
Table type & What it describes \\
\midrule
PK sampling & PK samples relative to dosing \\
PD sampling/assessment & Pharmacodynamic measurements \\
Laboratory assessment & Hematology, chemistry, urinalysis, etc. \\
Safety assessment & AEs, vitals, ECG, physical exam, etc. \\
Efficacy assessment & Tumor assessments, disease scores, etc. \\
Study treatment/dosing & Dose, schedule, route, cycles \\
Randomization/stratification & Arms, cohorts, strata \\
Visit/assessment window & Target dates and allowable windows \\
Sample/specimen & Blood, urine, tissue, storage, processing \\
Imaging/scan schedule & CT, MRI, PET, or other imaging \\
Dose modification & Rules for interruption, reduction, or discontinuation \\
Stopping/discontinuation rules & Treatment or study discontinuation criteria \\
\bottomrule
\end{tabular}
\caption{Other common protocol table types beyond those covered in \ourbench{}.}
\label{tab:table-types}
\end{table}

\section{Use of AI Assistants}
\label{app:ai-use}
We used AI assistants to polish the writing of this paper.

\section{Data Sources and Licenses}
\label{app:licenses}
The protocol documents in \ourbench{} are publicly posted on ClinicalTrials.gov \citep{clinicaltrials_gov}, the trial registry of the U.S. National Library of Medicine (NLM), where sponsors upload protocols as part of results reporting. We do not redistribute the protocol PDFs. Instead, we release the NCT identifier and the ClinicalTrials.gov URL of each document, so that users obtain the documents directly from the registry. The structured timeline references are taken from the USDM example studies published by CDISC \citep{cdisc_usdm}, which we use as released and credit to CDISC. Our reference annotations and code are released under the Creative Commons Attribution 4.0 (CC BY 4.0) license.

\end{document}